\documentclass[10pt,twocolumn,letterpaper]{article}

\usepackage[pagenumbers]{cvpr} 

\usepackage[accsupp]{axessibility}  

\usepackage{graphicx}
\usepackage{amsmath}
\usepackage{amssymb}
\usepackage{booktabs}
\usepackage{arydshln}
\usepackage{float}
\usepackage[pagebackref,breaklinks,colorlinks]{hyperref}
\usepackage[capitalize]{cleveref}

\crefname{section}{Sec.}{Secs.}
\Crefname{section}{Section}{Sections}
\Crefname{table}{Table}{Tables}
\crefname{table}{Tab.}{Tabs.}

\newcommand{\mypar}[1]{\vspace{1mm}\noindent\textbf{#1.}}

\begin{document}

\title{StyleMesh: Style Transfer for Indoor 3D Scene Reconstructions}

\author{
Lukas H{\"o}llein$^1$ \quad Justin Johnson$^2$ \quad Matthias Nie{\ss}ner$^1$ \\
$^1$Technical University of Munich \quad $^2$University of Michigan
}

\begin{figure}

\twocolumn[{
\renewcommand\twocolumn[1][]{#1}
\maketitle
\centering
\includegraphics[width=\linewidth]{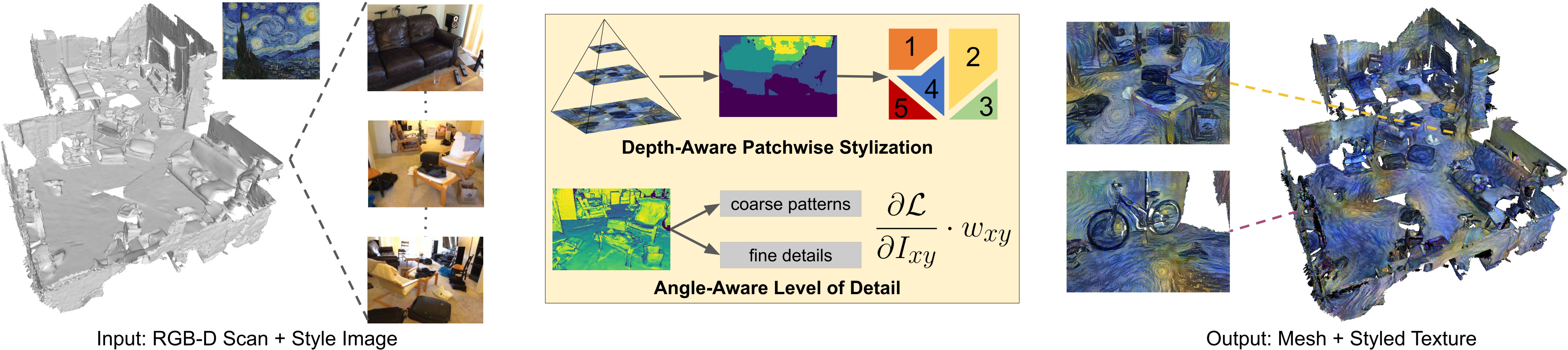}
\caption{
We perform style transfer on reconstructed 3D meshes by synthesizing stylized textures.
We compute style transfer losses on views of the scene and backpropagate gradients to the texture.
Depth and surface normal data from the mesh enable 3D-aware stylization, preventing artifacts that arise from standard 2D losses.
Our stylized meshes can be rendered using the traditional graphics pipeline.
}
\vspace{3mm}
\label{problem}
}]
\end{figure}

\begin{abstract}
\vspace{-2mm}
We apply style transfer on mesh reconstructions of indoor scenes.
This enables VR applications like experiencing 3D environments painted in the style of a favorite artist.
Style transfer typically operates on 2D images, making stylization of a mesh challenging.
When optimized over a variety of poses, stylization patterns become stretched out and inconsistent in size.
On the other hand, model-based 3D style transfer methods exist that allow stylization from a sparse set of images, but they require a network at inference time.
To this end, we optimize an explicit texture for the reconstructed mesh of a scene and stylize it jointly from all available input images.
Our depth- and angle-aware optimization leverages surface normal and depth data of the underlying mesh to create a uniform and consistent stylization for the whole scene.
Our experiments show that our method creates sharp and detailed results for the complete scene without view-dependent artifacts.
Through extensive ablation studies, we show that the proposed 3D awareness enables style transfer to be applied to the 3D domain of a mesh.
Our method \footnote[1]{\url{https://lukashoel.github.io/stylemesh/}} can be used to render a stylized mesh in real-time with traditional rendering pipelines.
\vspace{-6mm}
\end{abstract}

\section{Introduction}
\label{sec:intro}

Creating 3D content from RGB-D scans is a popular topic in computer vision~\cite{dai2017bundle, izadi2011kinectfusion, newcombe2011kinectfusion, niessner2013real, avetisyan2019scan2cad}.
We tackle a novel use case in this area: stylization of a reconstructed mesh with an explicit RGB texture.
Neural Style Transfer (NST) shows great results for stylization of images or videos,
but stylization of 3D content like meshes has been underexplored.
We synthesize a texture for the mesh which is a combination of observed RGB colors and a painting's artistic style.
After stylization, one could explore the space in VR and see it painted in the style of Van Gogh.

Our use case is similar to prior texture mapping methods~\cite{zhou2014color, bi2017patch, huang20173dlite, huang2020adversarial, izadi2011kinectfusion, dessein2014seamless, waechter2014let}
which construct a texture from a set of posed RGB images, but we produce a stylized texture rather than directly matching input images.
This is difficult since style transfer losses are typically defined on 2D image features~\cite{Gatys_2016_CVPR}, 
so NST does not immediately generalize to 3D meshes.
Recently, style transfer has been combined with novel view synthesis to stylize arbitrary scenes with a neural renderer from a sparse set of input images~\cite{huang_2021_3d_scene_stylization, kopanas2021point, chiang2021stylizing}.
These model-based methods require a forward pass during inference and cannot directly be applied to meshes.
Kato \etal~\cite{kato2018neural} and Mordvintsev \etal~\cite{mordvintsev2018differentiable} use differentiable rendering to bridge the gap between image style transfer and texture mapping:
backpropagating image losses to a texture representation enables consistent mesh stylization.

However, applying these methods to room-scale geometry is challenging as the resulting stylization patterns are noisy and can contain view-dependent stretch and size artifacts.
For example, optimizing a surface from a small grazing angle creates patterns in the image plane for that pose.
Viewing the same surface from an orthogonal angle then shows stretched-out patterns due to the perspective distortion.
Similarly, seeing an object from close and far-away viewpoints mixes small and large patterns on the same surface.
Perceiving the depth thus becomes harder, due to inconsistent stylization sizes.
These issues arise because 2D style transfer losses do not incorporate 3D data like surface normals and depth.
Instead, textures are separately stylized in each pose's image plane.

To this end, we formulate an energy minimization problem over the texture that combines texture mapping with style transfer (similar to~\cite{mordvintsev2018differentiable}) and minimizes style transfer losses for each pose in a 3D-aware manner that avoids view-dependent artifacts.
First, we utilize depth to render image patches at increasingly larger screen-space resolutions. 
By splitting the style loss calculation over these patches, we create larger stylization patterns in the foreground than the background.
As a result, patterns have the same size in world-space and are optimized in a view-independent way.
Second, we use the angle between the surface normal and view direction to determine the degree of stylization for each pixel.
By calculating Gram matrices from different style image resolutions (similar to~\cite{liu2020geometric}) areas seen from small grazing angles are stylized with coarse details,
which are later refined if they are seen from better angles.
Third, we avoid discretization artifacts by scaling gradients with per-pixel angle and depth weights during backpropagation.

Compared to state-of-the-art 3D style transfer methods, our experiments show an improvement in terms of 3D consistent stylization both qualitatively and quantitatively.
Additionally, our explicit texture representation allows for direct usage with traditional rendering pipelines.

To summarize, our contributions are:
\begin{itemize}
  \item Style transfer for room-scale indoor scene meshes with a new texture optimization, which results in 3D consistent textures and mitigates view-dependent artifacts.
  \item A depth-aware optimization at different screen-space resolutions, that creates equally-sized stylization patterns in the world-space of the mesh.
  \item An angle-aware optimization at different stylization details, that creates unstretched stylization patterns in the world-space of the mesh.
\end{itemize}

\section{Related Work}
\label{sec:related_work}

Our approach is a NST method operating on the texture parametrization of a mesh. It is related to recent work on style transfer for videos and 3D objects, as well as texture generation from RGB-D images.

\mypar{Texture Mapping}
Many methods texture a reconstructed mesh from multiple RGB images, i.e., they map a texture onto the geometry that combines the color information of all images~\cite{zhou2014color, bi2017patch, huang20173dlite, huang2020adversarial, izadi2011kinectfusion, dessein2014seamless, waechter2014let}.
These methods must handle inaccuracies in pose, geometry, color and distortions to find the best texture for the scene.
In contrast, we aim to create a texture that is also styled to a specific image
and avoid view-dependent stylization artifacts by introducing depth- and angle-awareness into the optimization.

\mypar{Image Style Transfer}
NST, first introduced in Gatys \etal~\cite{Gatys_2016_CVPR}, can be optimization-based~\cite{Gatys_2016_CVPR, gatys2017controlling, chiu2020iterative} or model-based~\cite{ulyanov2016texture, johnson2016perceptual, huang2017arbitrary, deng2020arbitrary}.
It is inherently defined in the image domain by matching CNN features either globally or in a local, patch-based manner ~\cite{Gatys_2016_CVPR, li2016combining, kolkin2019style, mechrez2018contextual, kalischek2021light}.
Thus, it cannot directly utilize 3D data like depth or surface normals of a mesh.
This can lead to view-dependent stylization artifacts when optimizing a texture through multiple poses.
We induce 3D-awareness into optimization-based NST by splitting the loss calculation across different image segments.

\mypar{Video Style Transfer}
Video style transfer (VST) methods consistently stylize RGB video frames with a given style.
These methods are optimization-based~\cite{ruder2018artistic, ruder2016artistic} or model-based~\cite{gupta2017characterizing, chen2017coherent, xia2021real, chen2020optical, gao2018reconet, gao2020fast, wang2020consistent} and employ temporal consistency or optical flow constraints.
Other methods combine features in a temporally consistent way, without using optical flow or depth constraints directly~\cite{li2019learning, deng2020arbitrary}.
VST methods can be combined with texture mapping to achieve consistent stylization of indoor scenes.
However, since VST optimizations are unaware of the underlying 3D structures, the resulting textures are often blurry or low-detail.

\mypar{3D Style Transfer}
Lifting style transfer into 3D has been explored for texturing individual objects~\cite{kato2018neural, mordvintsev2018differentiable, yin20213dstylenet} or faces~\cite{han2021exemplar}.
However, they focus on isolated objects (not room-scale scenes) and do not utilize 3D data.
In contrast, our method stylizes complete indoor scenes in a 3D-aware way.
Another line of work applies exemplar-based NST to 3D models~\cite{Hauptfleisch20-PG, Sykora19-EG}, guiding the stylization process explicitly from (hand-crafted) examples.
In contrast, we follow original NST by stylizing 3D scene models from artistic paintings and camera images.
Cao \etal~\cite{cao2020psnet} stylize indoor scenes using a point cloud that cannot be directly used to texture a mesh.
Other methods combine novel view synthesis and NST for consistent stylization from only a few input images~\cite{kopanas2021point, chiang2021stylizing, huang_2021_3d_scene_stylization}.
In contrast, we do not require a network during inference to produce stylization results;
our results can be rendered by a standard graphics pipeline.

\begin{figure*}[ht]
\centering
\includegraphics[width=\linewidth]{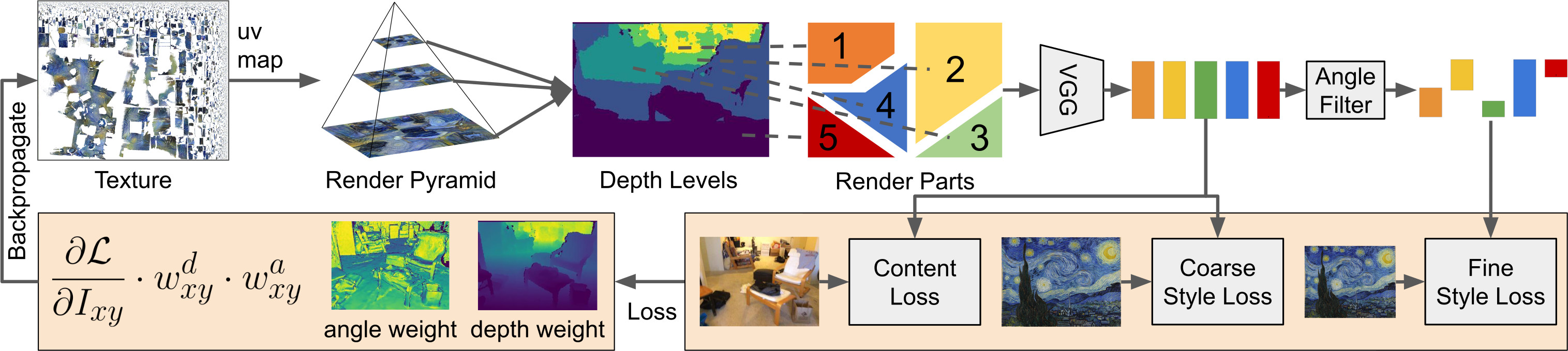}
\caption{We optimize a texture for the mesh reconstruction of a scene using multiple RGB images and a style image. We sample the texture with $uv$ maps at different resolutions, yielding a render pyramid for each pose. Using depth, we divide the screen space into parts, each corresponding to a different pyramid level. Each is encoded separately and compared with its RGB image part in the content loss. Using normals, the style loss is further split into fine and coarse branches. We discard features seen from small grazing angles for fine stylization. Finally, we scale the resulting image gradients with continuous angle and depth weights before backpropagating to the texture.}
\label{fig:pipeline}
\end{figure*}

\section{Method}
\label{sec:method}

Our goal is to stylize the mesh of an indoor scene: we want to create a texture that is a mixture of original RGB colors and a style image.
To avoid view-dependent artifacts, we formulate a depth- and angle-aware optimization problem over all images.
We require a set of $N$ images $\{I_k\}_{k=1}^N$ captured at different poses. 
We also need a mesh reconstruction of the scene for which we create a texture parametrization, i.e., we need a $uv$ coordinate per vertex. 
For each pose, we sample the texture with the corresponding $uv$ map at multiple resolutions, yielding a render pyramid.
Depending on the depth of each pixel, we split the image into multiple render parts, each belonging to one pyramid resolution.
Each part is used in content and style losses, where we only stylize pixels with fine details, that are seen from good angles.
Finally, we smooth the per-pixel gradients before backpropagating to the texture.
The complete method is visualized in~\cref{fig:pipeline}.

\subsection{Texture Optimization}\label{subsec:Texture Optimization}
We optimize a stylized RGB texture $\mathcal{T}^*$ from all RGB images $\{I_k\}_{k=1}^N$ and a separate style image $I_s$.
Similar to~\cite{mordvintsev2018differentiable}, we formulate a minimization problem with content and style losses $\mathcal{L}_c, \mathcal{L}_s$ and add a regularization term $\mathcal{L}_r$:
\begin{equation}
\label{eq:step-1}
    \mathcal{T}^* = \arg\min_{\mathcal{T}} \sum_{i=0}^{N}(
        \lambda_c \mathcal{L}_c(I_i, \hat{P}_i) + \lambda_s \mathcal{L}_s(I_s, \hat{P}_i) + \lambda_{r} \mathcal{L}_{r}(\mathcal{T})
    )
\end{equation}
where $\hat{P}_i$ is the render pyramid for the current pose, sampled from the texture with the corresponding $uv$ maps and $\lambda_c, \lambda_s, \lambda_r$ are loss weights.
The sampling operation is identical to traditional graphics and differentiable, i.e., we bilinearly interpolate each pixel from four neighboring texels.
Similar to Thies \etal~\cite{thies2019deferred}, we define our texture using a Laplacian Pyramid to regularize the texels in each layer with $\mathcal{L}_{r}$.
This helps avoid magnification and minification artifacts and reduces visible noise in the texture.
For each pose, we optimize the subset of observed texels.
Thus, we require a pose set covering most of the scene to optimize the texture completely.
In contrast, stylizing in texture space directly~\cite{yin20213dstylenet} is problematic for a room-scale texture parametrization, which may contain many seams.

\subsection{Depth Level Render Parts}\label{subsec:Depth Scaling}
\begin{figure}
\centering
\setlength\tabcolsep{1pt}
\begin{tabular}{cc}
\includegraphics[width=0.48\linewidth]{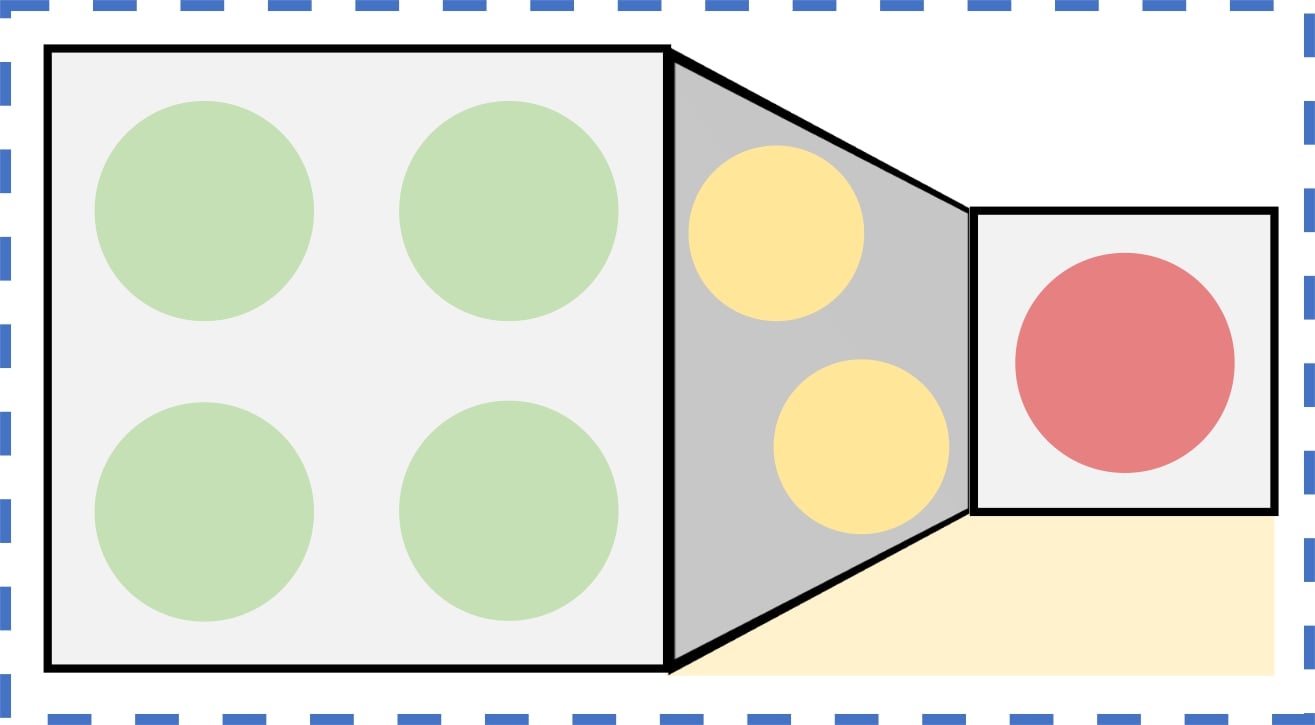} & \includegraphics[width=0.48\linewidth]{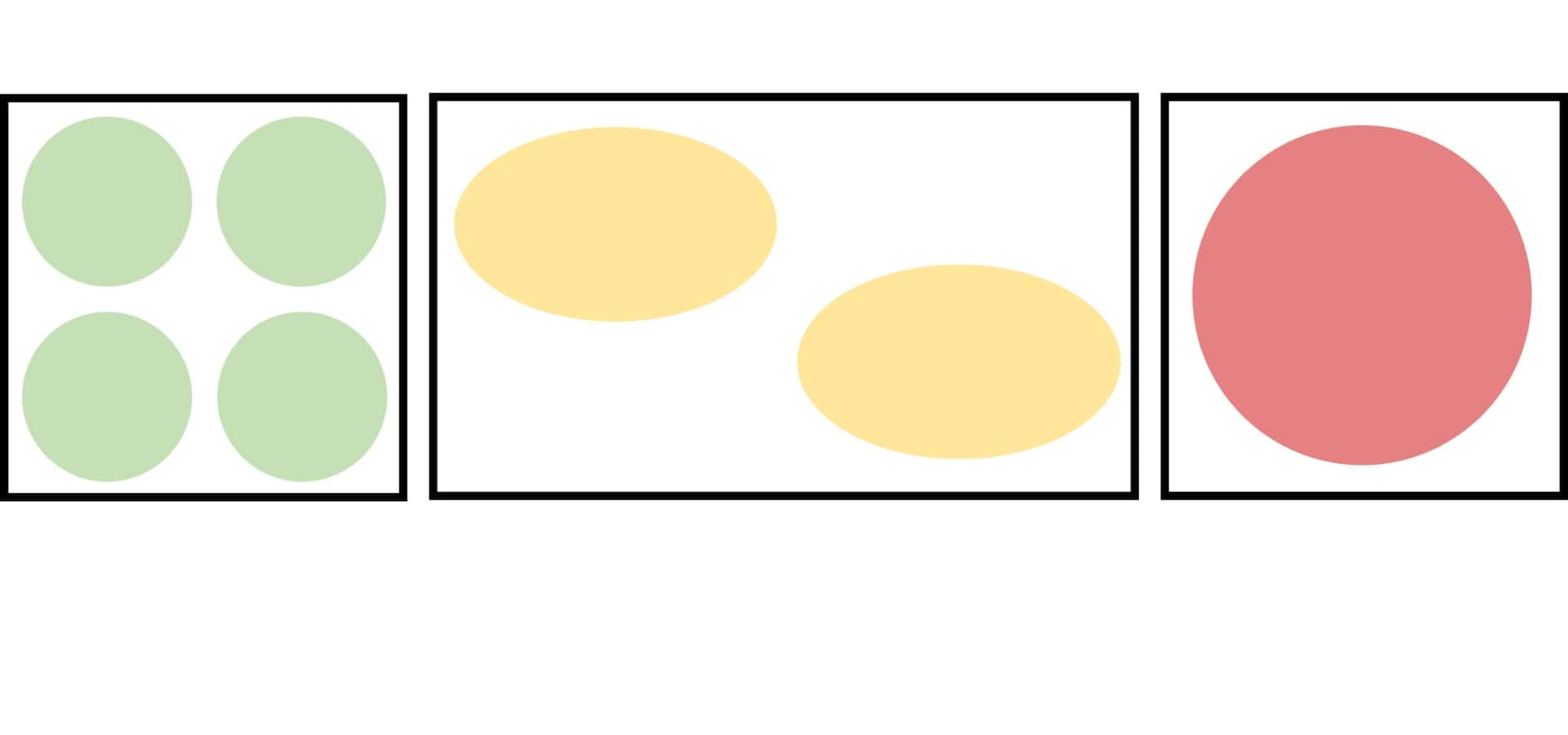} \\
(a) Projected Geometry & (b) Geometry in World-Space
\end{tabular}
\caption{Stylizing mesh faces in 2D (screen-space) is dependent on angle and size of the projected geometry. Optimizing the faces of a wall from a small grazing angle leads to stretched-out stylization patterns in world-space (yellow). Regions of similar size can receive more or less patterns, depending on their projected size (green and red). We denote stylization patterns as circles that are optimized from screen-space into texture space.}
\label{fig:depth-scaling-angle-weighting}
\end{figure}

Style transfer operates on the CNN features of an image~\cite{Gatys_2016_CVPR}.
This leads to a limited sense of depth when optimizing over multiple poses.
Stylization patterns can appear equally large in the foreground and background, e.g., when parts of a surface are seen far-away and close-up (see~\cref{fig:depth-scaling-angle-weighting}).
Observing the same surface from multiple poses thus mixes small and large patterns next to each other.
As a result, renderings using the optimized texture do not convincingly capture depth.
Liu \etal~\cite{liu2017depth} make style transfer depth-aware in the image plane with a depth-loss network.
In contrast, we incorporate depth-awareness by optimizing at multiple screen-space resolutions.
Larger patterns appear in the foreground than the background of an image, ultimately leading to equally large style in world space.

We make use of the relation that area in screen-space is inversely proportional to depth, i.e., when depth increases by a factor of $p$, a given projected area decreases by $p^2$.
On the other hand, style transfer is agnostic to the image resolution that it is applied on, i.e., when resolution increases by $p^2$, stylization patterns appear proportionally smaller (because the receptive field becomes smaller relative to the resolution)~\cite{gatys2017controlling}.
We combine both relations to optimize stylization patterns having the same size in world space: when depth increases by $p$, we increase image resolution by $p^2$.

We apply the relation to divide the image into parts, sampled from the render pyramid at increasingly larger resolutions.
The content and style losses are then calculated independently for each part.
To discretize into parts, we define a minimum depth value $\theta_d$, making the relation absolute.
We calculate the optimal image height per-pixel as 
\begin{equation}
    R_{xy} = \theta_{min} \cdot \frac{d_{xy}}{\theta_d}
\end{equation}
where $d_{xy}$ is the depth at pixel $(x,y)$ and $\theta_{min}$ is the minimum resolution.
We express resolution $R_{xy}$ as height in pixels and scale the width accordingly.
We then map $R_{xy}$ to the nearest neighbor in the render pyramid, yielding its index as depth level per-pixel.
Finally, we apply a $3{\times}3$ erosion kernel to smooth the depth level map over all pixels.

\subsection{Angle Filter}\label{subsec:Angle Weighting}
Style transfer in screen-space can create stretched-out stylization patterns (see~\cref{fig:depth-scaling-angle-weighting}).
Patterns might look circular from one view, but are stretched-out ellipses in world space (e.g., when optimizing from a small grazing angle).
To prevent this, we combine coarse and fine style losses and optimize fine details only for areas seen from good angles.
Similar to previous work~\cite{liu2020geometric, gatys2017controlling}, we utilize the fact that the receptive field of high-resolution images is still small~\cite{luo2016understanding}.
As a result, stylization patterns appear coarser and less detailed, when optimized from a larger style image.
We find that coarse patterns are less prone to stretch artifacts.

For each pixel, we calculate its normal-to-view angle $\alpha_{xy} = \measuredangle(\Vec{n}_{xy}, \Vec{v})$ where $\Vec{n}_{xy}$ is the interpolated surface normal at pixel $(x,y)$ and $\Vec{v}$ is the viewing direction.
Only the pixels where $\alpha_{xy} \leq \theta_a$ are used for the style loss with a low-resolution style image that produces fine stylizations.
We always use all pixels and a high-resolution style image to optimize coarse stylization patterns.
This creates a combination of coarse and fine patterns without stretch artifacts.

\subsection{Multi-Resolution Part-based Losses}
Multiple content and style losses combine depth levels (\cref{subsec:Depth Scaling}) and angle filtering (\cref{subsec:Angle Weighting}) to optimize the texture without view-dependent artifacts.
We encode the render pyramid $\hat{P}$ with a pretrained VGG network~\cite{simonyan2014very} into the feature pyramid $\hat{F}$.
Using the depth level map, we only keep corresponding features in each layer of $\hat{F}$.
We compute a coarse Gram matrix $\hat{G}_{c}$ and an angle-filtered fine one $\hat{G}_{f}$ from the features in every layer.
Similarly, $G_c$ and $G_f$ correspond to the high- and low-resolution style images.
We define the style loss as
\begin{equation}
\mathcal{L}_s(I_s, \hat{P}) = \sum_{l}^{\theta_l} \hat{w}_l \cdot ( ||G_{c} - \hat{G}_{c}^l||^2_F + ||G_{f} - \hat{G}_{f}^l||^2_F)
\end{equation}
which sums over all depth levels independently (part-based) and combines coarse and fine stylization (multi-resolution).
We calculate the normalized weighting factor $\hat{w}_l$ as 
\begin{equation}
w_l = \frac{v_l}{t_l} \quad\mathrm{and}\quad \hat{w}_l = \frac{w_l}{\sum_{i}^{\theta_l} w_i}
\end{equation}
where $v_l$ is the visible and $t_l$ the total number of pixels in depth level $l$.
Similarly, the content loss is defined as
\begin{equation}
L_c(I, \hat{P}) = \sum_{l}^{\theta_l} \hat{w}_l \cdot ||F^l - \hat{F}^l||^2_F
\end{equation}
where $F^l$ are the features of the content image $I$, split in a similar way. 
For brevity, we omit different VGG layers and image indices from the notation.
As proposed in Gatys \etal~\cite{Gatys_2016_CVPR}, we use the layers \textit{relu\_\{1-5\}\_1} for the style loss and \textit{relu\_4\_2} for the content loss.
We calculate the losses independently for every VGG layer and sum them accordingly.

\subsection{Per-Pixel Gradient Scaling}
Depth levels (\cref{subsec:Depth Scaling}) and angle filtering (\cref{subsec:Angle Weighting}) impose hard thresholds on the image of each pose.
To avoid discretization artifacts at decision boundaries, we scale the per-pixel gradients before backpropagating them to the texture.
First, we calculate a weighting factor $w_{xy}^a = cos(\alpha_{xy})$ from the normal-to-view angle $\alpha_{xy}$.
This controls the influence of a pose on each pixel by preferring orthogonal over small grazing viewing angles.
Scaling features similar to Gatys \etal~\cite{gatys2017controlling} instead results in oversaturation artifacts.

Second, we adapt the idea of trilinear Mipmap interpolation~\cite{foley1996computer}.
Each pixel contributes to the render parts of its nearest two pyramid layers, resulting in two per-pixel gradients.
We calculate the distance to the nearest layer as
\begin{equation}
w_{xy}^d = |\frac{R_{xy} - L_{xy}^1}{L_{xy}^1 - L_{xy}^2}|
\end{equation}
where $R_{xy}$ is the optimal resolution for pixel $(x,y)$ and $L_{xy}^1$, $L_{xy}^2$ are the resolutions of the nearest and second nearest pyramid layers.
Finally, we linearly interpolate between the per-pixel gradients as
\begin{equation}
\frac{\partial \mathcal{L}}{\partial I_{xy}} = \frac{\partial \mathcal{L}_1}{\partial I_{xy}} \cdot w_{xy}^d \cdot w_{xy}^a + \frac{\partial \mathcal{L}_2}{\partial I_{xy}} \cdot (1 - w_{xy}^d) \cdot w_{xy}^a
\end{equation}
where $\mathcal{L}_1$ is the loss term for the nearest pyramid layer of pixel $(x,y)$ and $\mathcal{L}_2$ for the second nearest, respectively.

\subsection{Data Preprocessing}\label{subsec:Data Preprocessing}
We use the ScanNet~\cite{dai2017scannet} and Matterport3D~\cite{chang2017matterport3d} datasets, which provide RGB-D images and reconstructed meshes (we use per-region meshes for Matterport3D~\cite{chang2017matterport3d}).
We use the RGB images for optimization, but filter them with a Laplacian kernel to remove blurry images.
We reduce each mesh's complexity by merging vertices until $\leq500$K faces remain.
Then, we generate a texture parametrization with Blender's smart $uv$ project~\cite{blender} with an angle limit of $70^{\circ}$.
We precompute the $uv$ maps for each estimated pose.

\section{Results}
\label{sec:results}

\mypar{Implementation Details} 
We optimize textures at a resolution of $4096{\times}4096$ as a Laplacian Pyramid~\cite{thies2019deferred} with 4 layers and regularization strength $\lambda_{r}{=}5000$.
We use $\lambda_c{=}70$ and $\lambda_s{=}0.0001$ for content and style loss weights. 
We optimize for 7 epochs and repeat each frame 10 times.
We set $\theta_{min}{=}32$ and use $\theta_l{=}4$ render pyramid layers at heights of $\{256, 432, 608, 784\}$ pixels.
We set $\theta_a{=}30^{\circ}$, $\theta_d{=}0.25$ meters for ScanNet~\cite{dai2017scannet} and $\theta_a{=}40^{\circ}$, $\theta_d{=}0.2$ meters for Matterport3D~\cite{chang2017matterport3d}.
We incrementally halve the original style image resolution until either the width or height reaches a size of 256 pixels.
We use the resulting image for the stylization of fine details and a two steps larger image for coarse details.
We use Adam~\cite{kingma2014adam} with batch size 1 and initial learning rate $1$ which decays multiplicatively by $0.1$ every 3 epochs.
We tried L-BFGS~\cite{zhu1997algorithm} which gave similar results.
After optimization, we export the Laplacian Pyramid to a single texture image and use a standard rasterizer with Mipmaps and shading~\cite{foley1996computer} for rendering.

\mypar{Evaluation Metrics} 
We conduct a user study to show the advantages of depth- and angle-awareness (\cref{fig:user-study-ablation}). 
Additionally, we quantify them by stylizing with a ``circle'' image (\cref{fig:circles}).
We calculate the correlation between circle size and depth in screen-space (Corr. 2D) and world-space (Corr. 3D), as well as circle stretch as the ratio of horizontal and vertical radius in world-space (\cref{tab:ablation}).
For quantifying 3D consistency, we calculate the $L_1$ distance between source and reprojected target frames (\cref{tab:reprojection}).
Please refer to the supplemental material for more details about metrics.

\subsection{Style Transfer on Scenes}\label{subsec:Style Transfer on Scenes}

\begin{figure*}
\centering
\setlength\tabcolsep{1pt}
\begin{tabular}{ccccc}
\includegraphics[height=24mm, width=32mm]{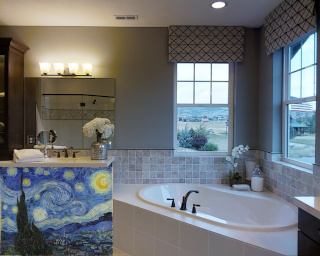} & 
\includegraphics[height=24mm, width=32mm]{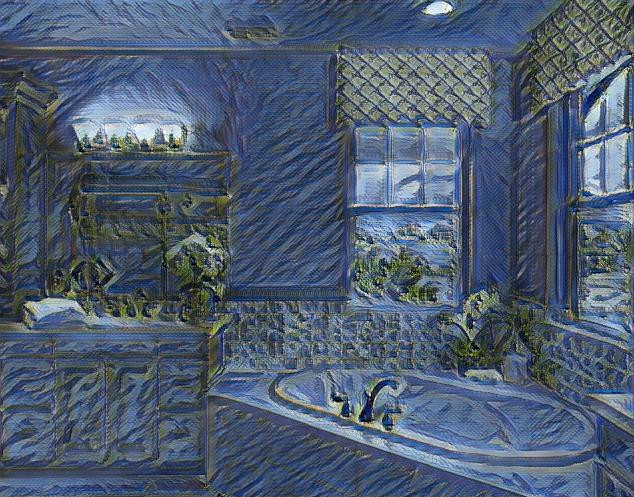} & 
\includegraphics[height=24mm, width=32mm]{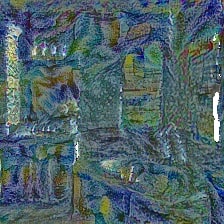} & 
\includegraphics[height=24mm, width=32mm]{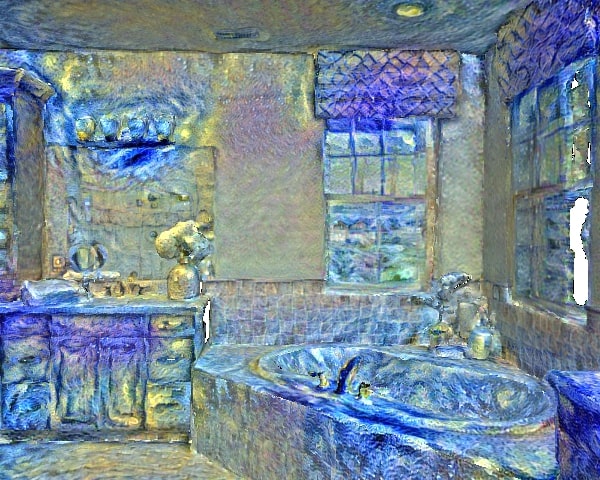} & 
\includegraphics[height=24mm, width=32mm]{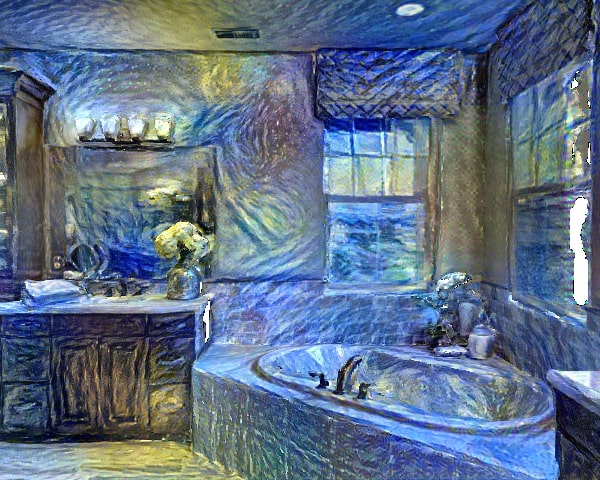} \\
\includegraphics[height=24mm, width=32mm]{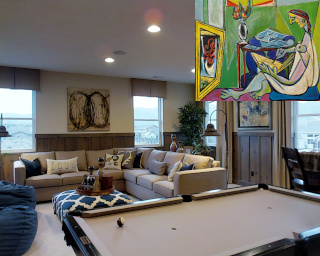} & 
\includegraphics[height=24mm, width=32mm]{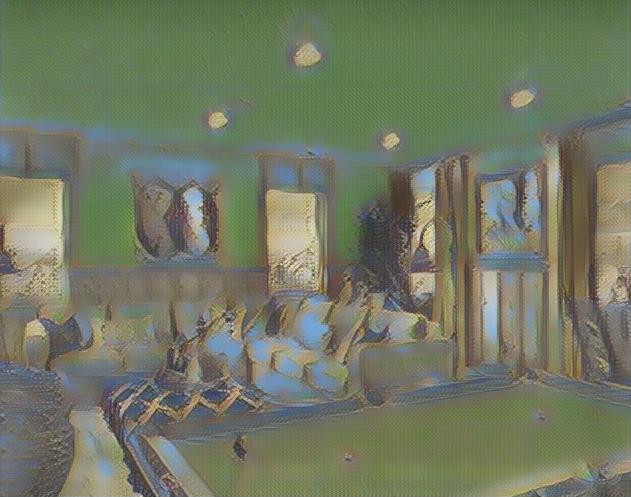} & 
\includegraphics[height=24mm, width=32mm]{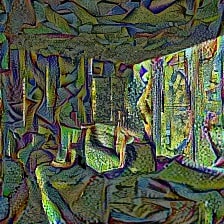} & 
\includegraphics[height=24mm, width=32mm]{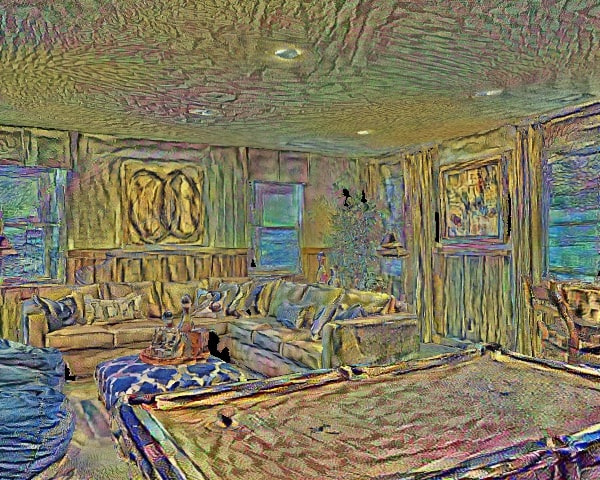} & 
\includegraphics[height=24mm, width=32mm]{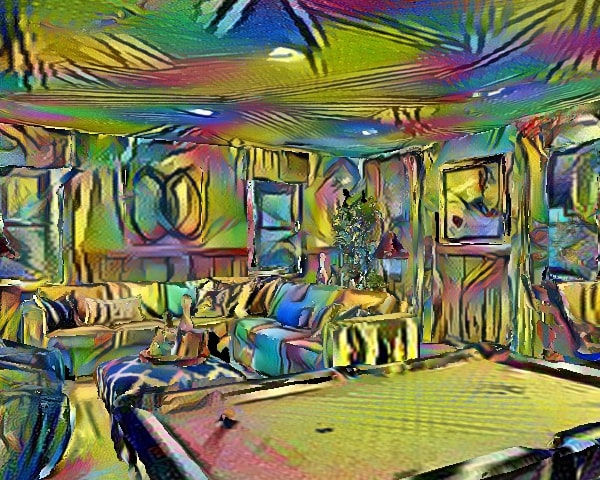} \\
\includegraphics[height=24mm, width=32mm]{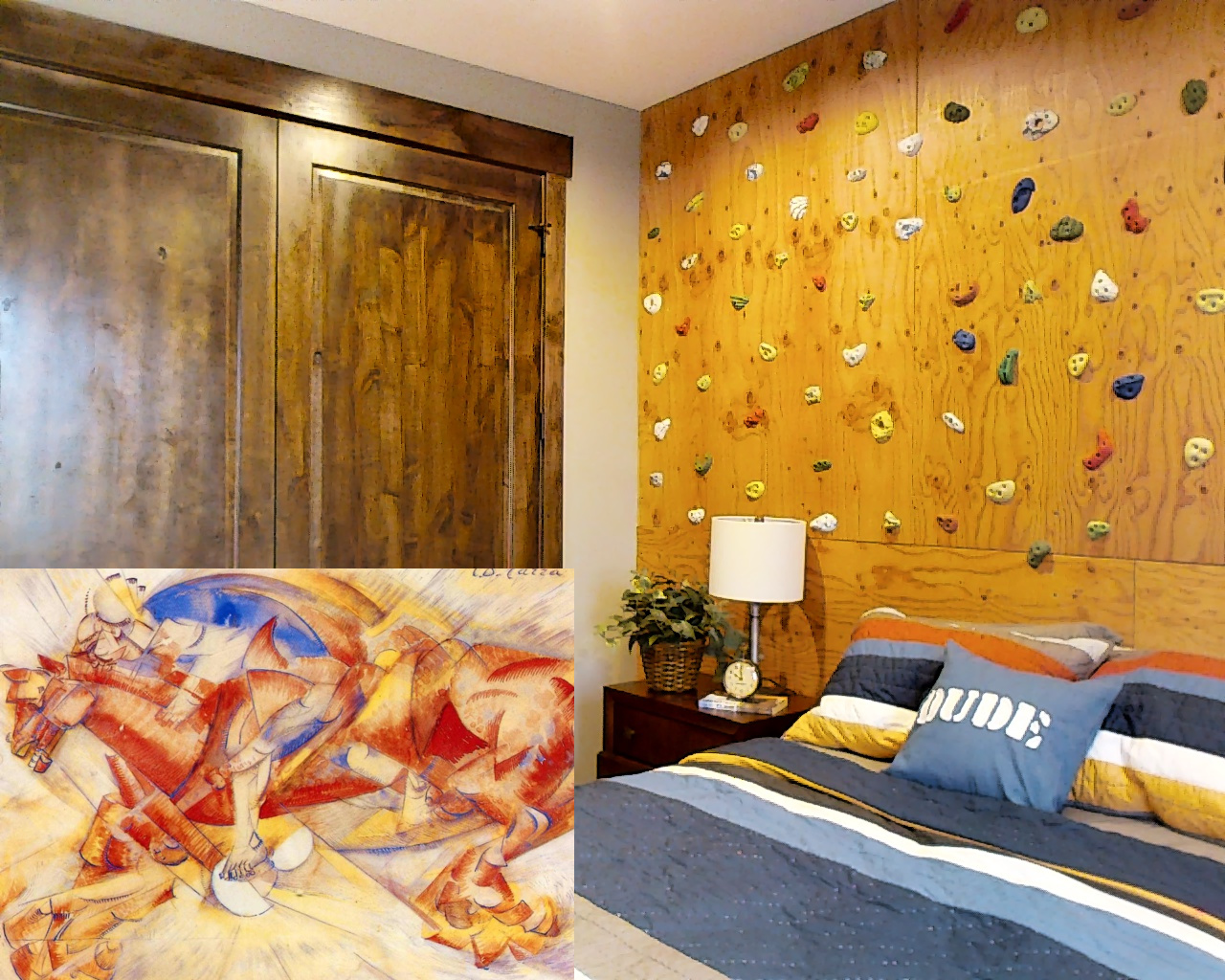} & 
\includegraphics[height=24mm, width=32mm]{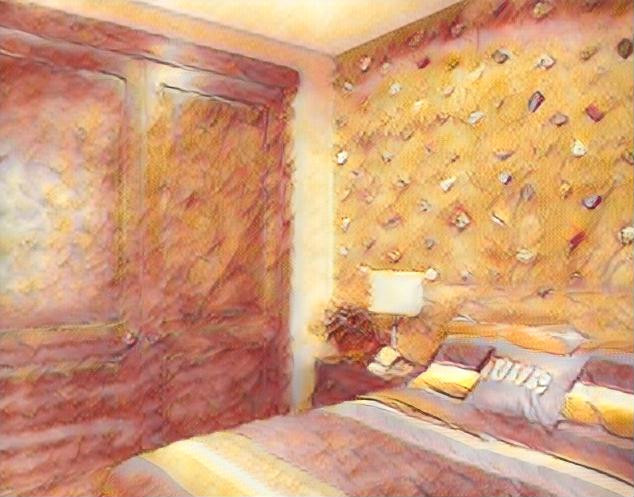} & 
\includegraphics[height=24mm, width=32mm]{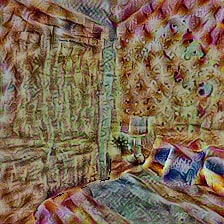} & 
\includegraphics[height=24mm, width=32mm]{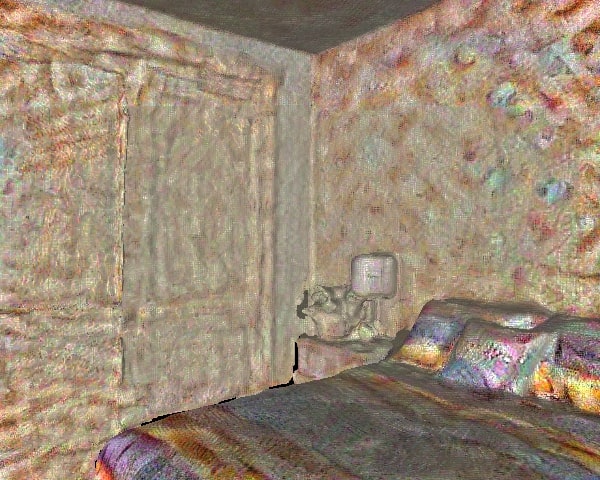} & 
\includegraphics[height=24mm, width=32mm]{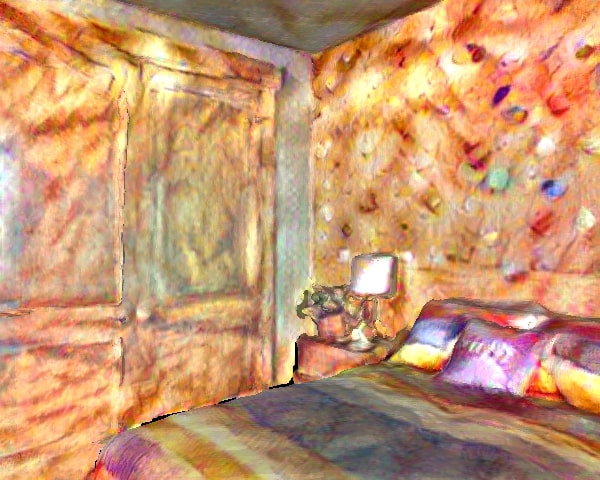} \\
RGB and Style & LSNV~\cite{huang_2021_3d_scene_stylization} & NMR~\cite{kato2018neural} & DIP~\cite{mordvintsev2018differentiable} & \textbf{Ours}
\end{tabular}
\caption{Comparison of stylization results for our method and related work on the Matterport3D~\cite{chang2017matterport3d} dataset. We texture the mesh with each method (point cloud for Huang \etal~\cite{huang_2021_3d_scene_stylization} respectively) and render a single pose that is also captured in the RGB images.}
\label{fig:comp-baseline-matterport}
\end{figure*}

\begin{figure*}
\centering
\setlength\tabcolsep{1pt}
\begin{tabular}{ccccc}
\includegraphics[height=24mm, width=32mm]{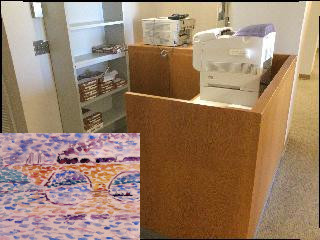} & 
\includegraphics[height=24mm, width=32mm]{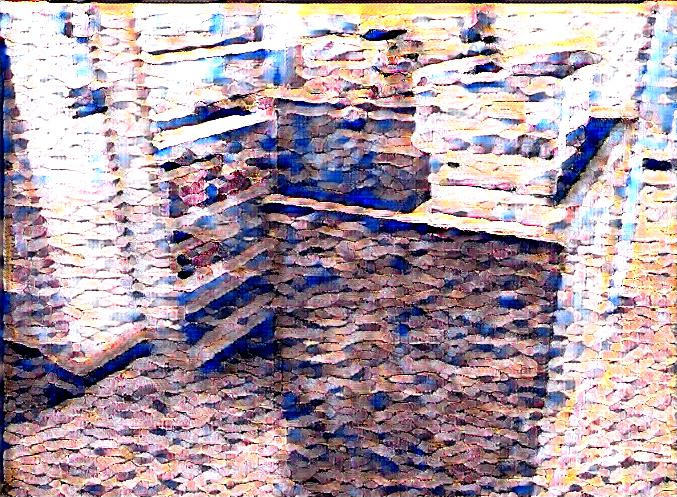} & 
\includegraphics[height=24mm, width=32mm]{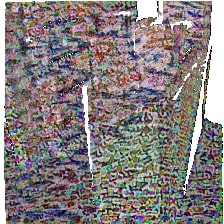} & 
\includegraphics[height=24mm, width=32mm]{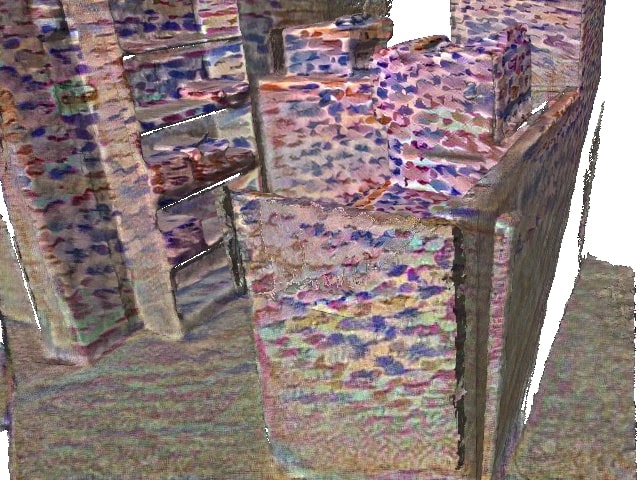} & 
\includegraphics[height=24mm, width=32mm]{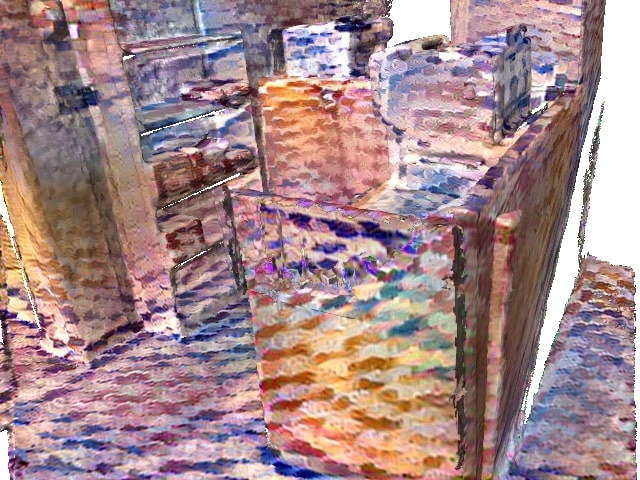} \\
\includegraphics[height=24mm, width=32mm]{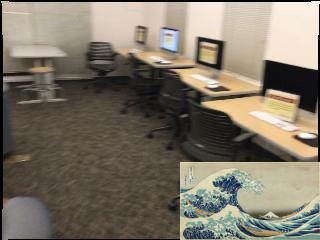} & 
\includegraphics[height=24mm, width=32mm]{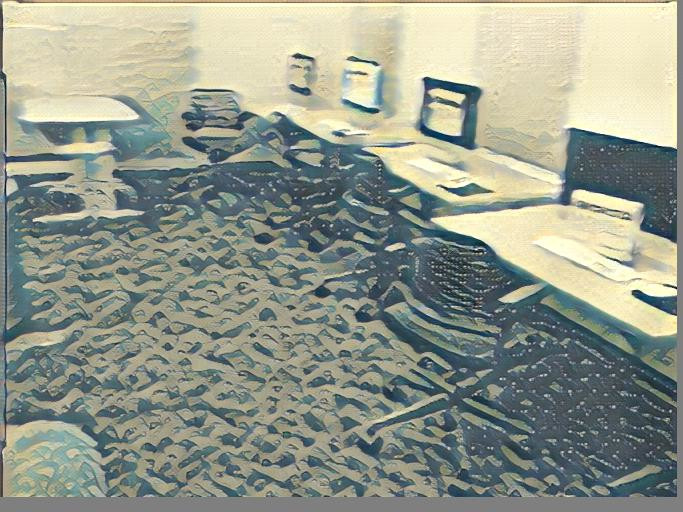} & 
\includegraphics[height=24mm, width=32mm]{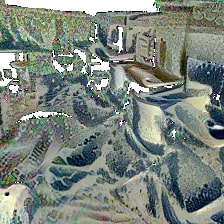} & 
\includegraphics[height=24mm, width=32mm]{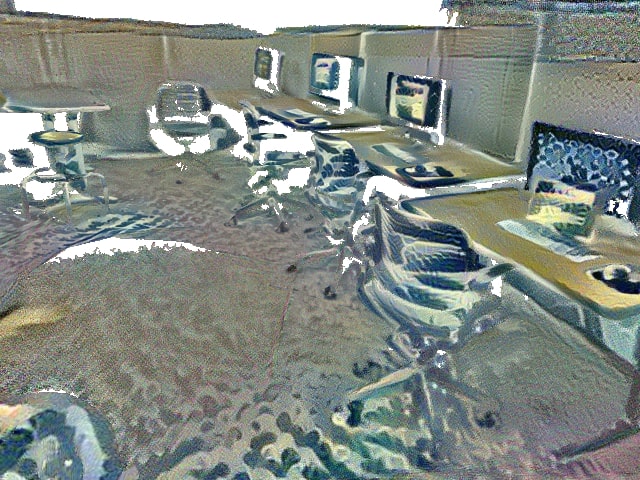} & 
\includegraphics[height=24mm, width=32mm]{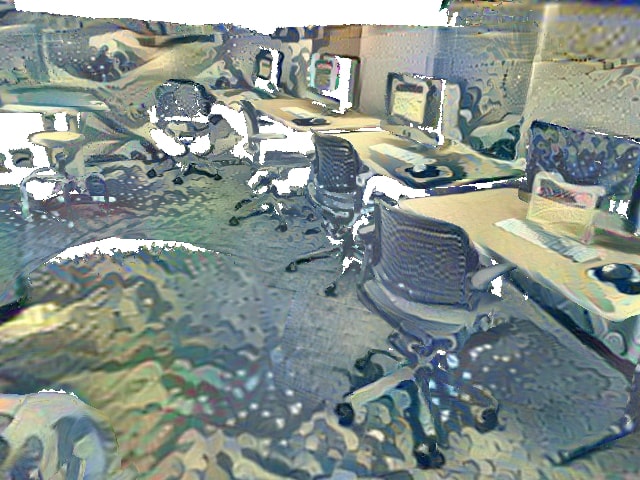} \\
RGB and Style & LSNV~\cite{huang_2021_3d_scene_stylization} & NMR~\cite{kato2018neural} & DIP~\cite{mordvintsev2018differentiable} & \textbf{Ours}
\end{tabular}
\caption{Comparison of stylization results for our method and related work on the ScanNet~\cite{dai2017scannet} dataset. We texture the mesh with each method (point cloud for Huang \etal~\cite{huang_2021_3d_scene_stylization} respectively) and render a single pose that is also captured in the RGB images.}
\label{fig:comp-baseline-scannet-3D}
\end{figure*}

Our method competes with 3D style transfer methods that stylize a scene through an explicit or implicit representation.
Specifically, we compare our method with DIP of Mordvintsev \etal~\cite{mordvintsev2018differentiable} and NMR of Kato \etal~\cite{kato2018neural}: like us they also optimize a texture, but they do not utilize angle or depth data.
Additionally, we compare with LSNV of Huang \etal~\cite{huang_2021_3d_scene_stylization}, which uses a neural renderer to stylize point clouds.
We show results on the Matterport3D~\cite{chang2017matterport3d} dataset in~\cref{fig:comp-baseline-matterport} and on the ScanNet~\cite{dai2017scannet} dataset in~\cref{fig:comp-baseline-scannet-3D}.
A visualization of textured meshes is given in~\cref{fig:comp-baseline-3D}.
Please see the supplemental material for more examples.

Our results show that we are able to stylize scenes without view-dependent size or stretch artifacts.
In contrast to the other methods, our approach creates sharp and detailed effects for the complete scene.
Optimizing the complete texture is especially difficult for DIP~\cite{mordvintsev2018differentiable} and NMR~\cite{kato2018neural}, which both contain noisy texels.
LSNV~\cite{huang_2021_3d_scene_stylization} stylizes complete images, but their results are less detailed.
To quantitatively evaluate our method and the related approaches, we compute the mean $L_1$ distance between source frame and a reprojected target frame.
The results are listed in~\cref{tab:reprojection}.

\begin{figure}
\centering
\setlength\tabcolsep{1pt}
\begin{tabular}{cccc}
\includegraphics[width=0.24\linewidth]{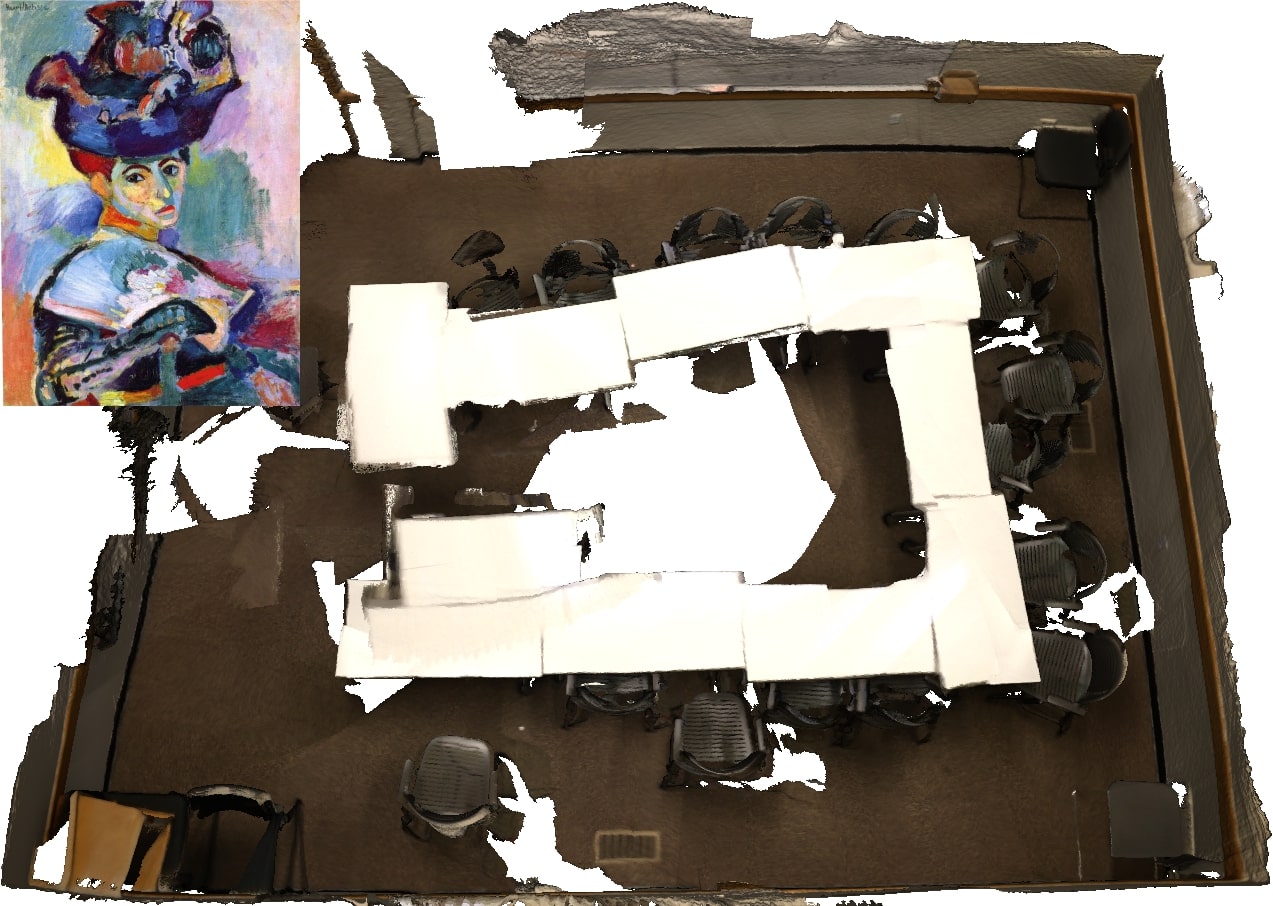} & 
\includegraphics[width=0.24\linewidth]{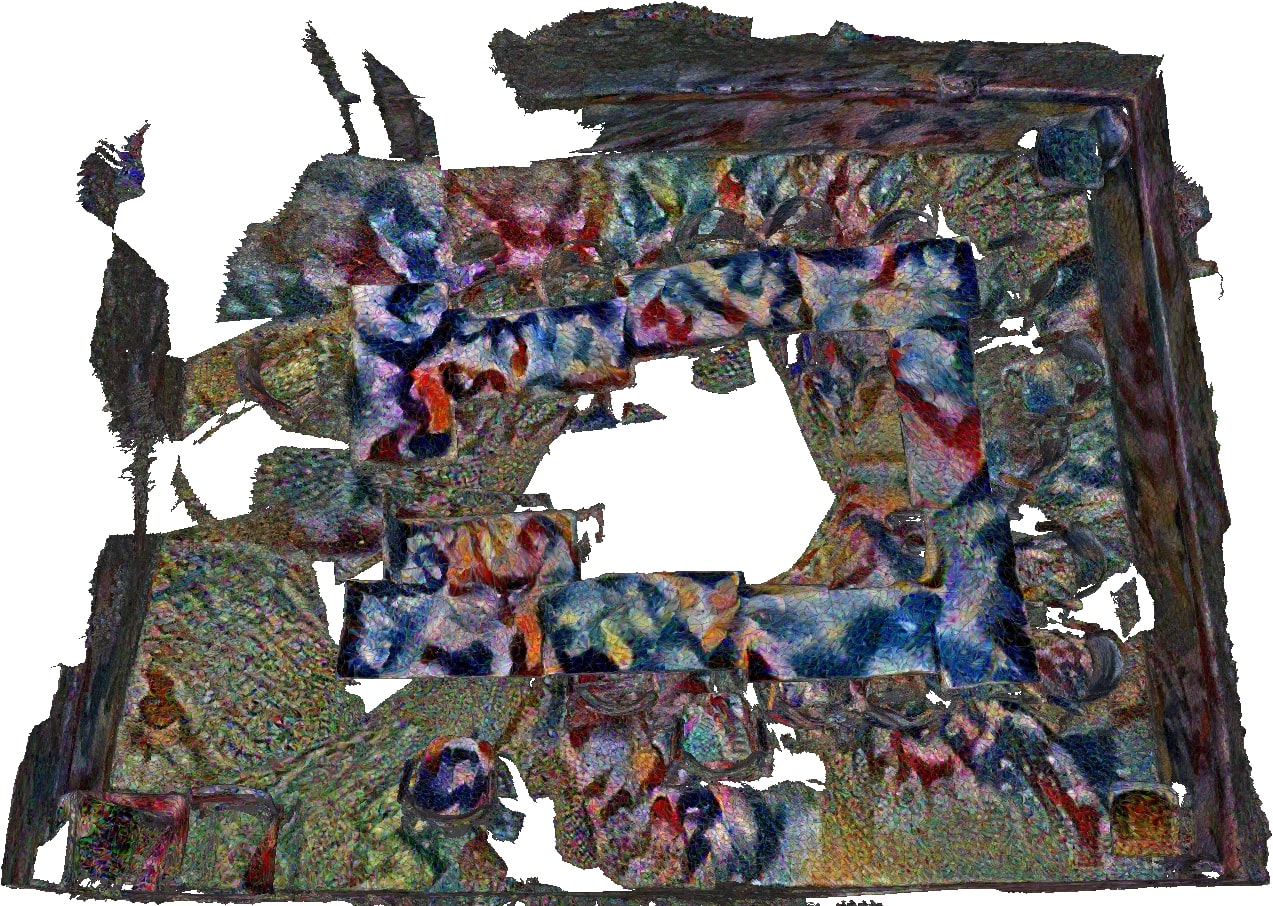} & 
\includegraphics[width=0.24\linewidth]{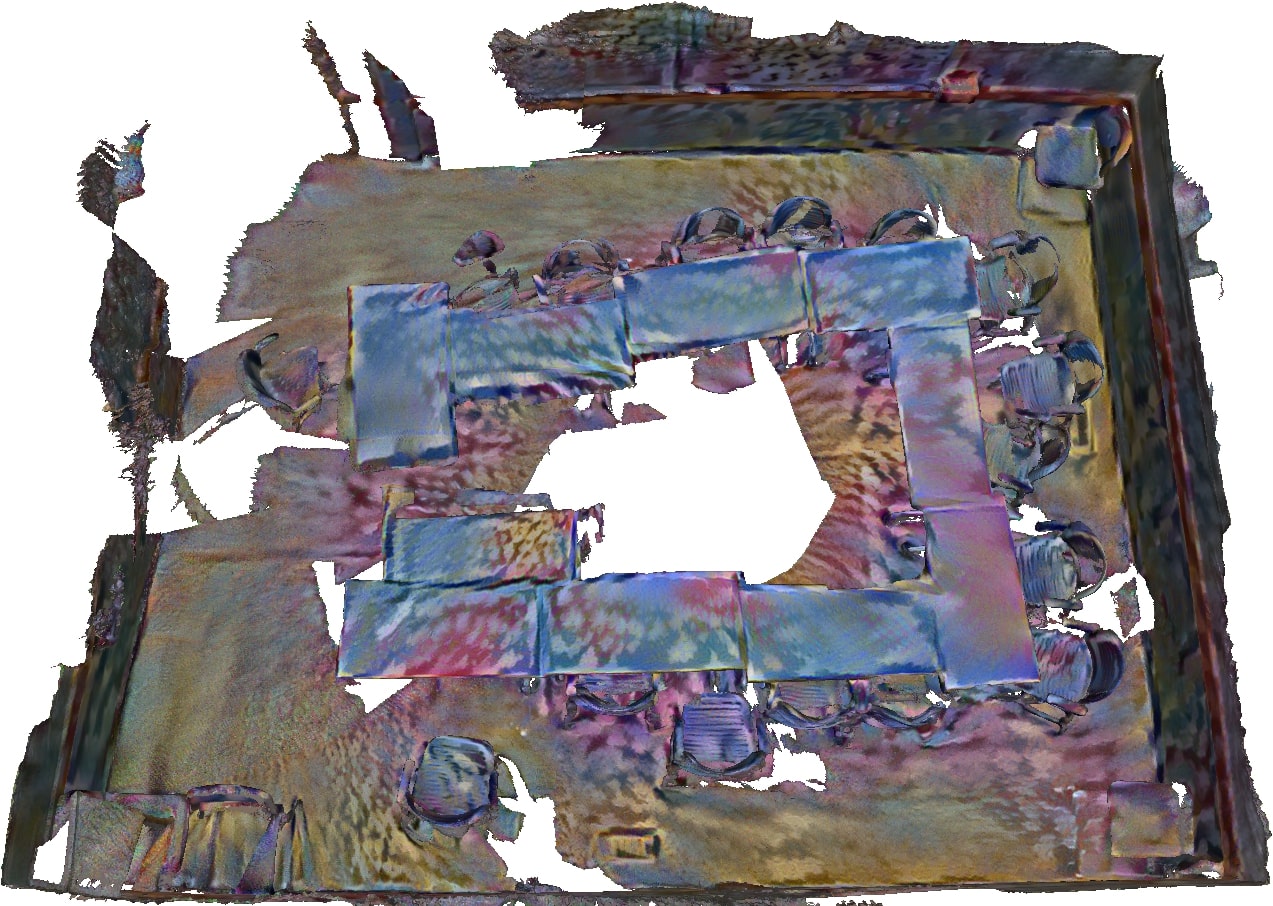} & 
\includegraphics[width=0.24\linewidth]{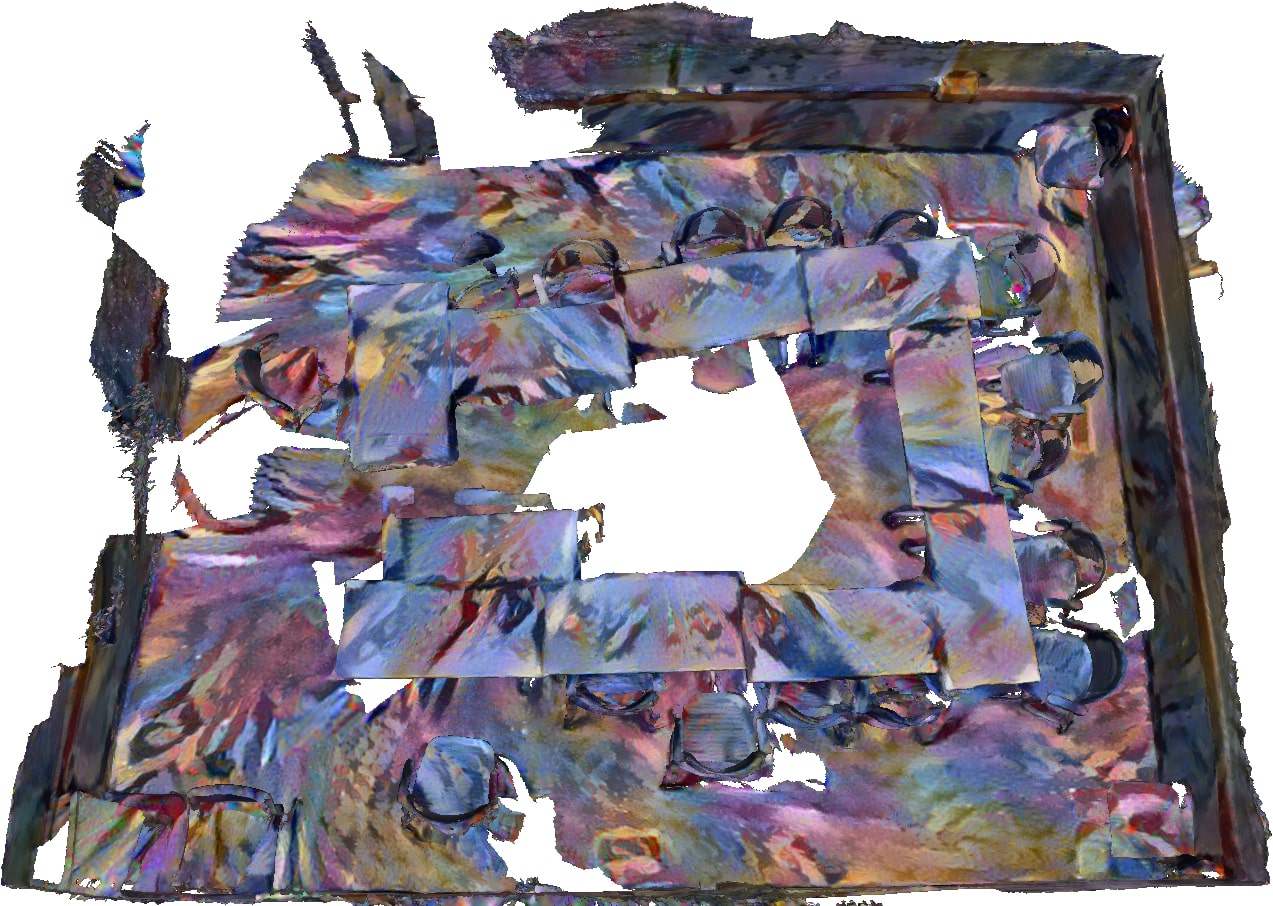} \\
\includegraphics[width=0.24\linewidth]{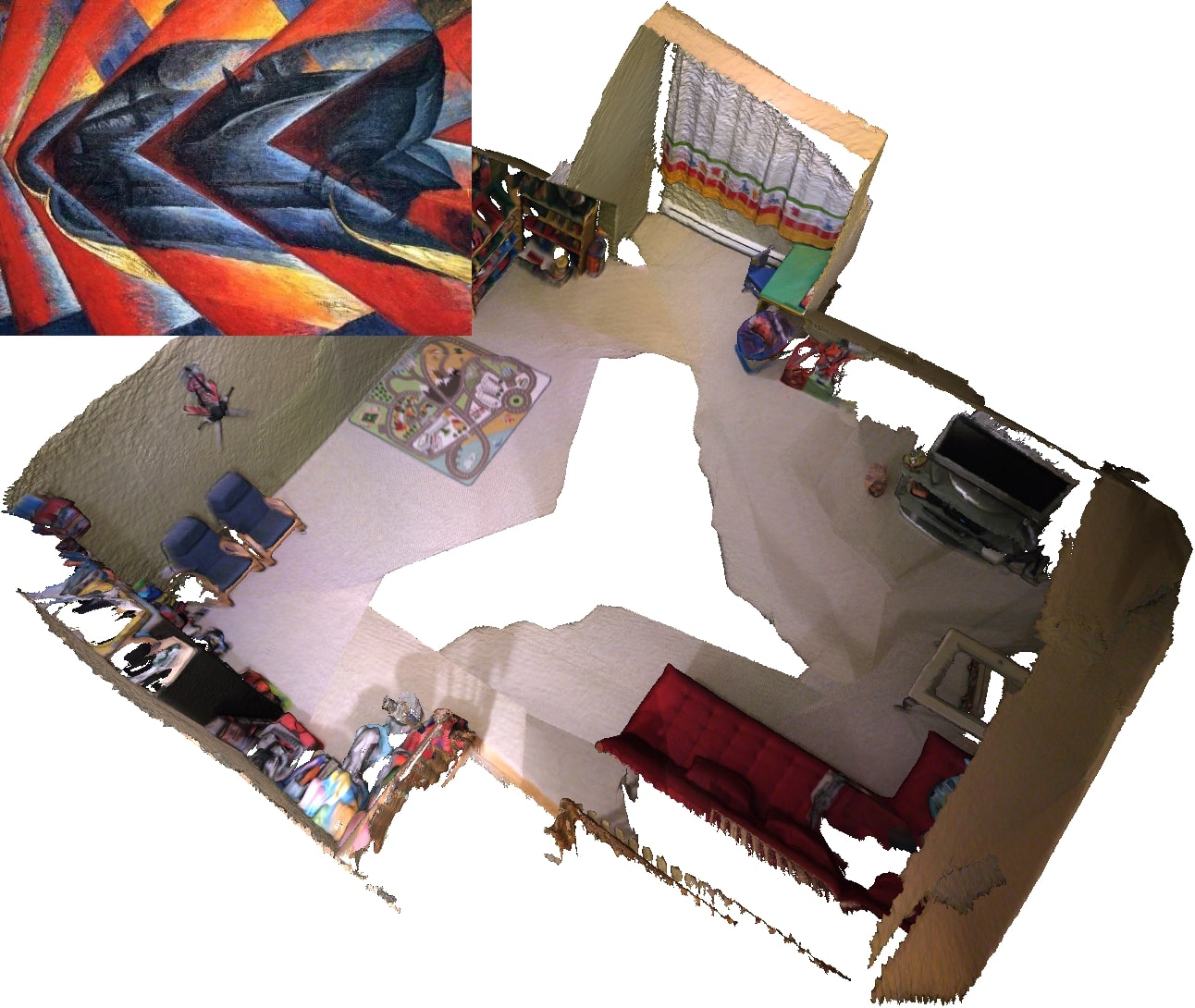} & 
\includegraphics[width=0.24\linewidth]{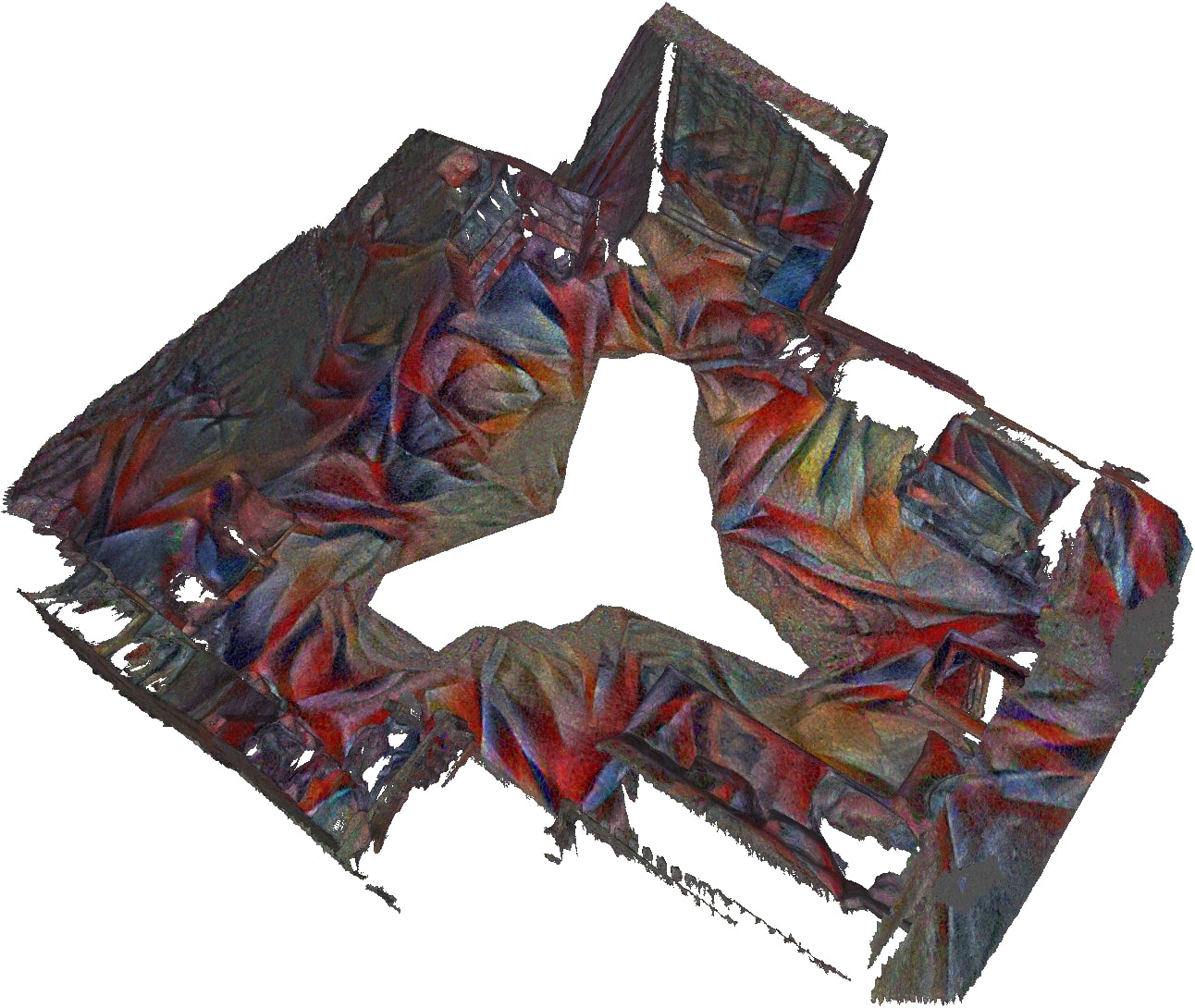} & 
\includegraphics[width=0.24\linewidth]{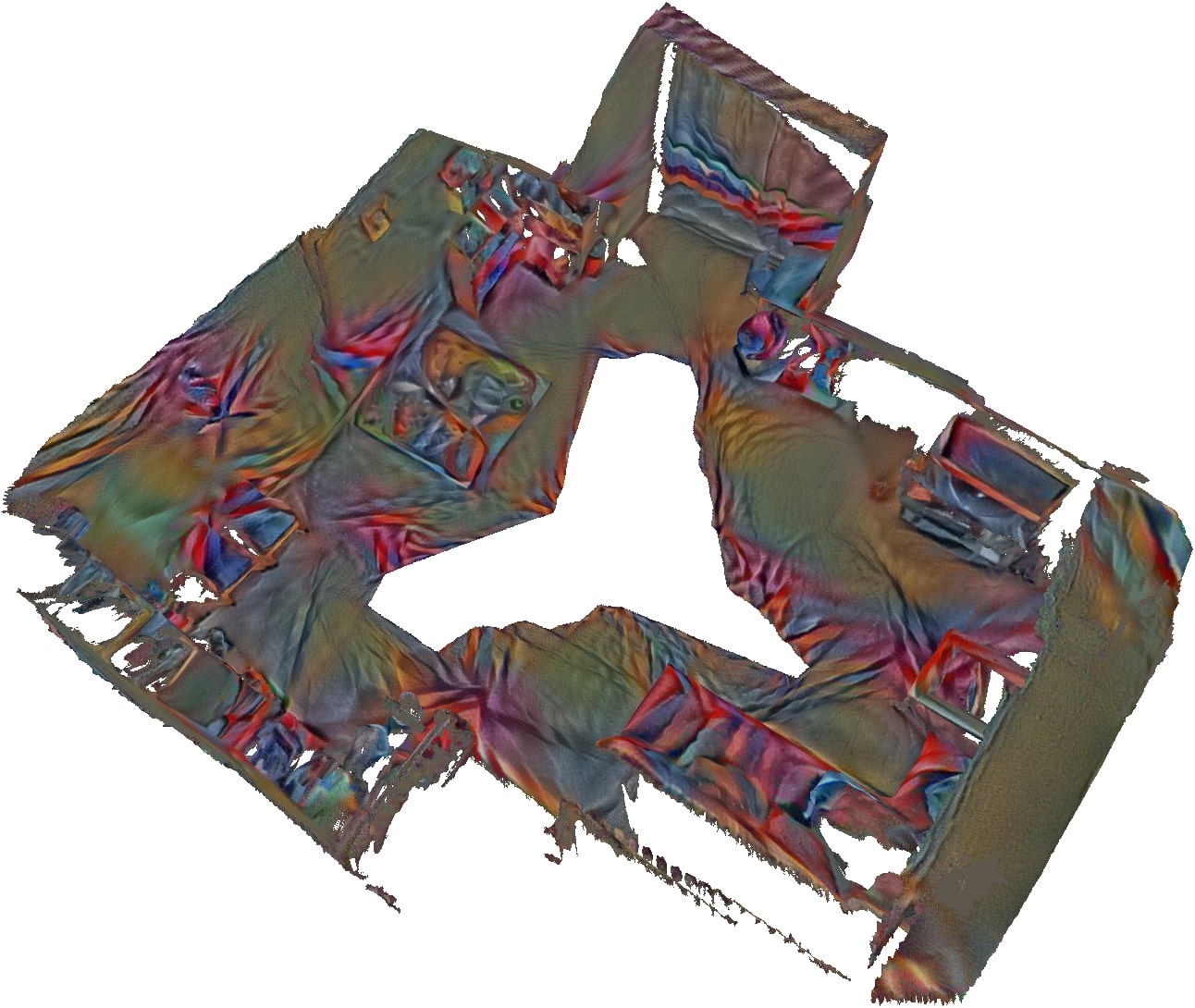} & 
\includegraphics[width=0.24\linewidth]{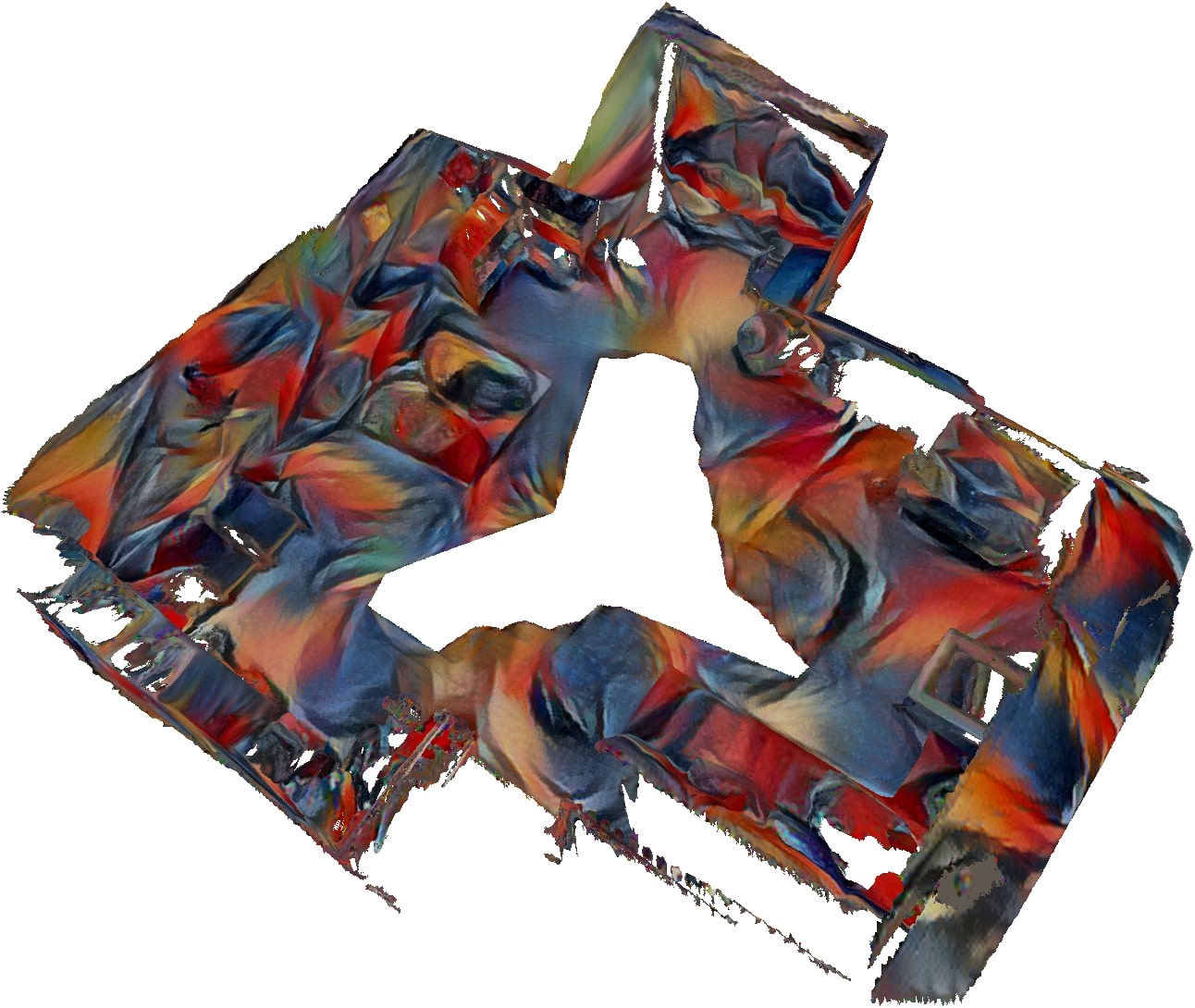} \\
RGB Mesh & NMR~\cite{kato2018neural} & DIP~\cite{mordvintsev2018differentiable} & \textbf{Ours}
\end{tabular}
\caption{Top-down view on stylized meshes in comparison to previous work.}
\label{fig:comp-baseline-3D}
\end{figure}

\begin{table}
  \centering
  \begin{tabular}{|l|c|c|}
    \hline
    Method & Short-Range $\downarrow$ & Long-Range $\downarrow$ \\
    \hline
    \hline
    LSNV~\cite{huang_2021_3d_scene_stylization} & 4.873 & 7.207 \\
    NMR~\cite{kato2018neural} & 1.565 & 2.165 \\
    DIP~\cite{mordvintsev2018differentiable} & 1.396 & 1.723 \\
    \hline
    Ours & \textbf{1.225} & \textbf{1.566} \\
    \hline
  \end{tabular}
  \caption{$L_1$ distance between source frame and a reprojected target frame. We report mean values for short-range (2nd next frame) and long-range (20th next frame) in 10 different ScanNet~\cite{dai2017scannet} scenes.}\label{tab:reprojection}
\end{table}

\subsection{Ablation Studies}\label{subsec:Ablation Study}

\begin{figure*}
\centering
\setlength\tabcolsep{1pt}
\begin{tabular}{cccc}
\includegraphics[width=0.24\linewidth]{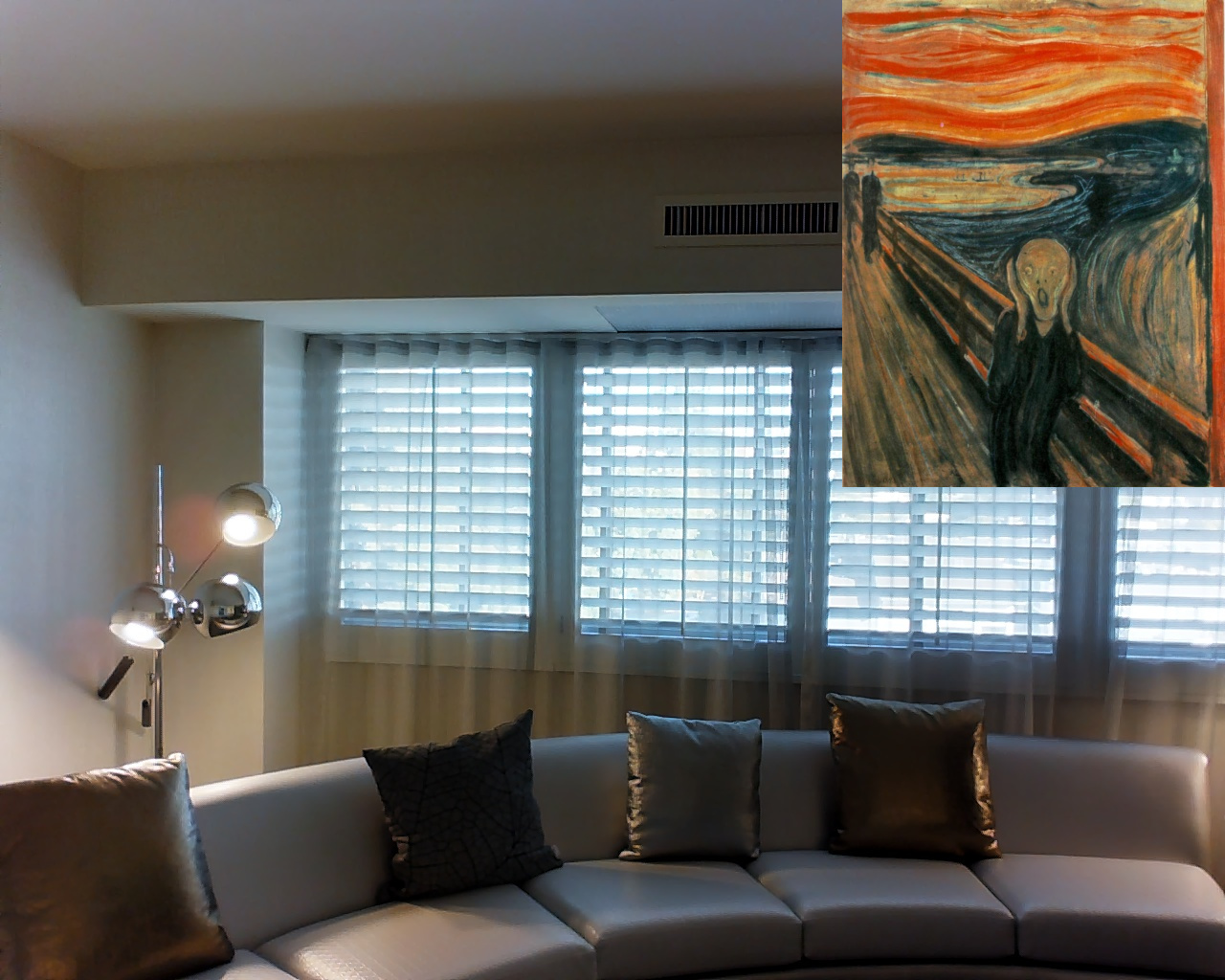} & 
\includegraphics[width=0.24\linewidth]{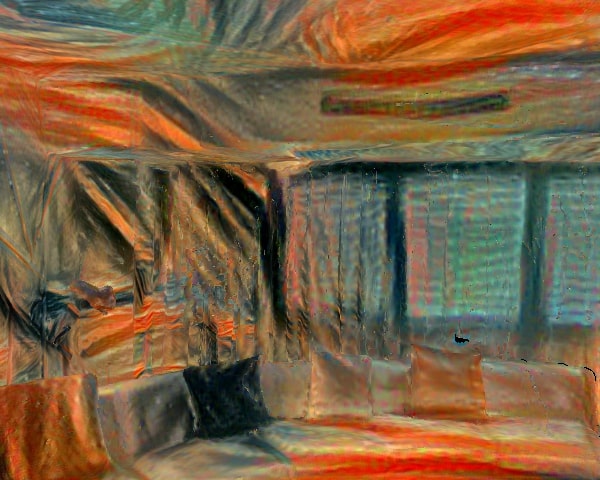} & 
\includegraphics[width=0.24\linewidth]{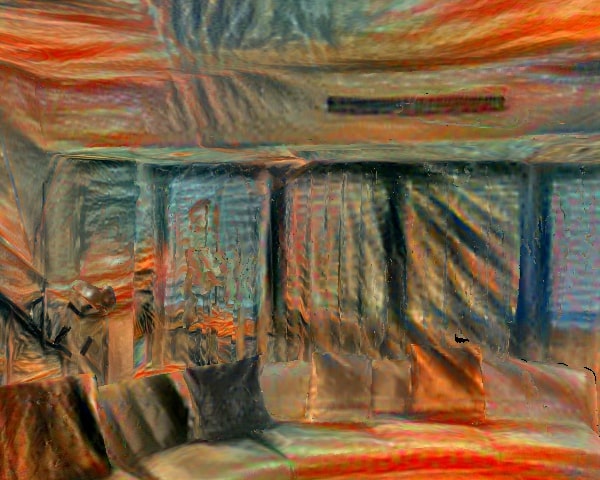} & 
\includegraphics[width=0.24\linewidth]{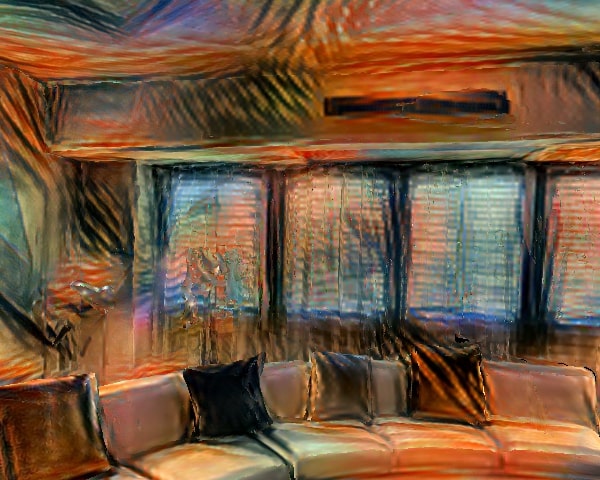} \\
\includegraphics[width=0.24\linewidth]{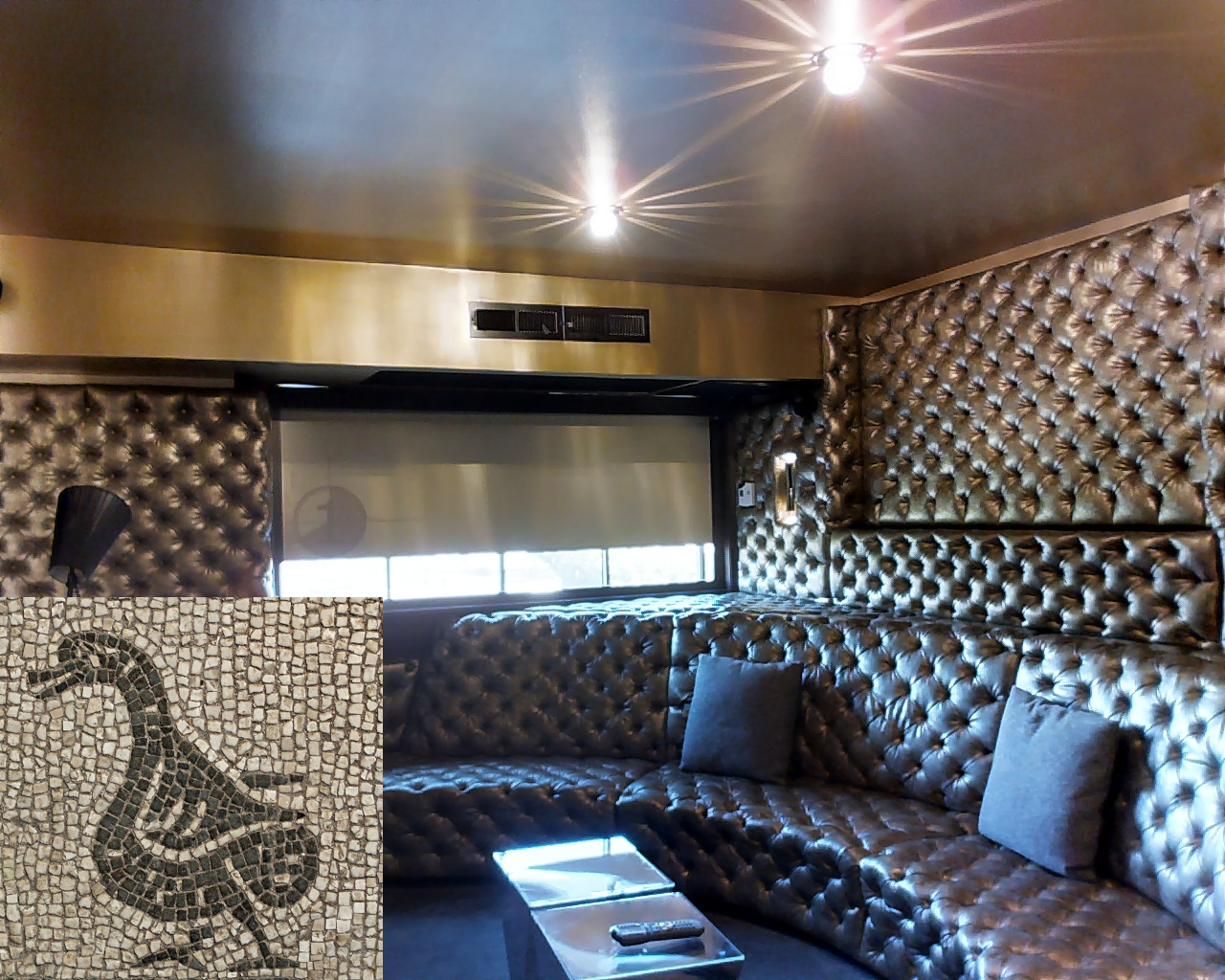} & 
\includegraphics[width=0.24\linewidth]{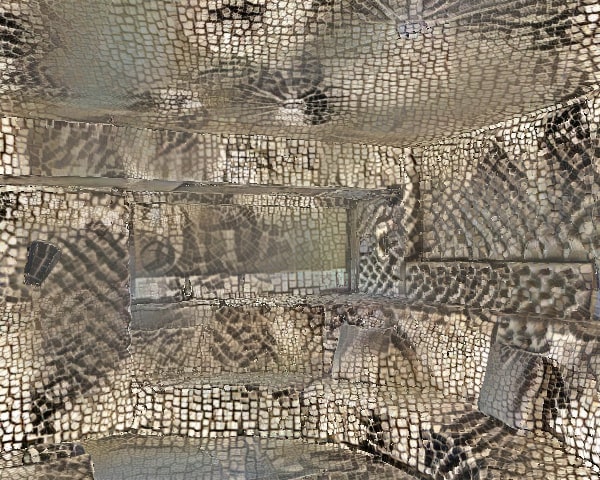} & 
\includegraphics[width=0.24\linewidth]{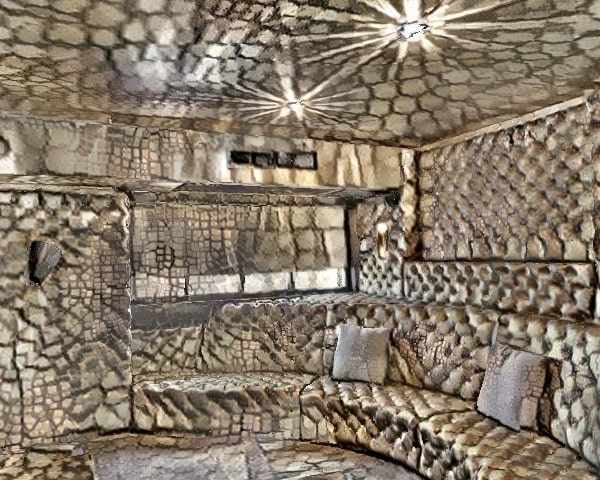} & 
\includegraphics[width=0.24\linewidth]{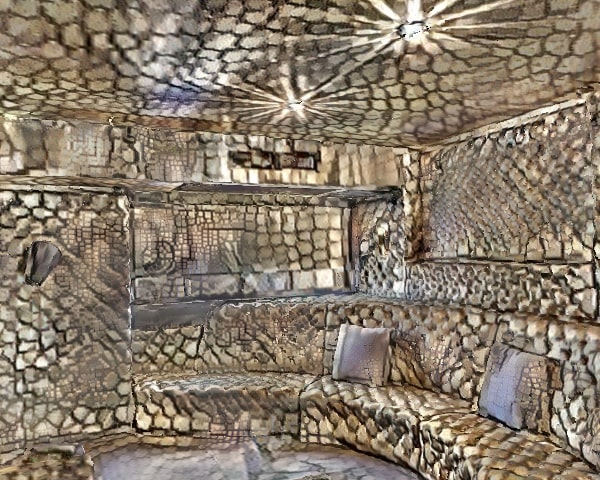} \\
\includegraphics[width=0.24\linewidth]{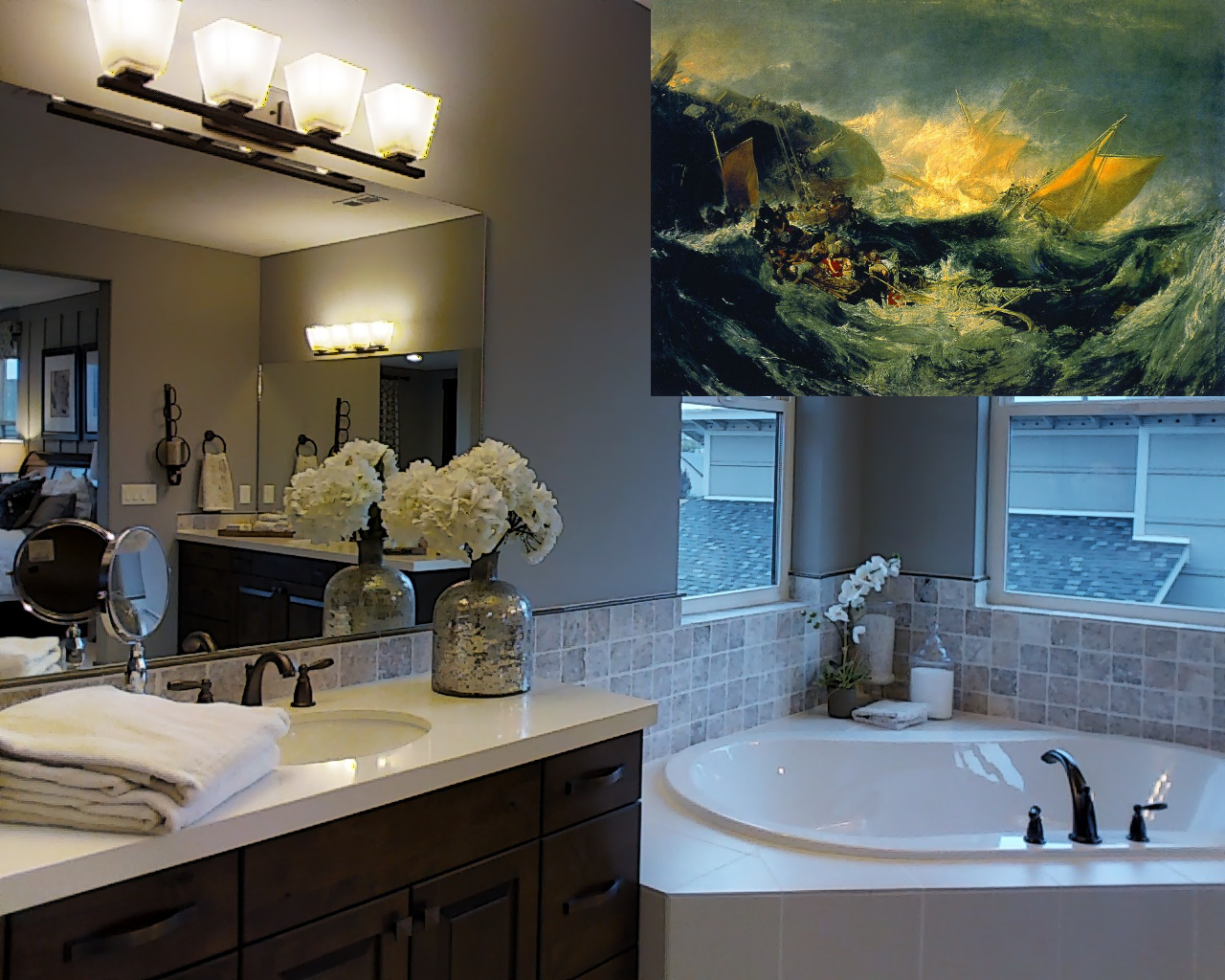} & 
\includegraphics[width=0.24\linewidth]{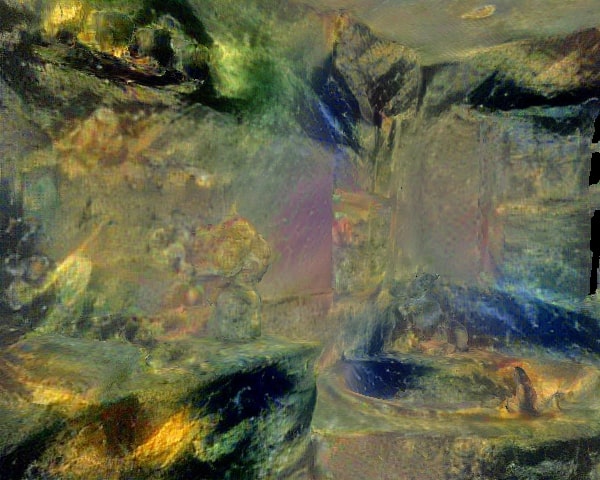} & 
\includegraphics[width=0.24\linewidth]{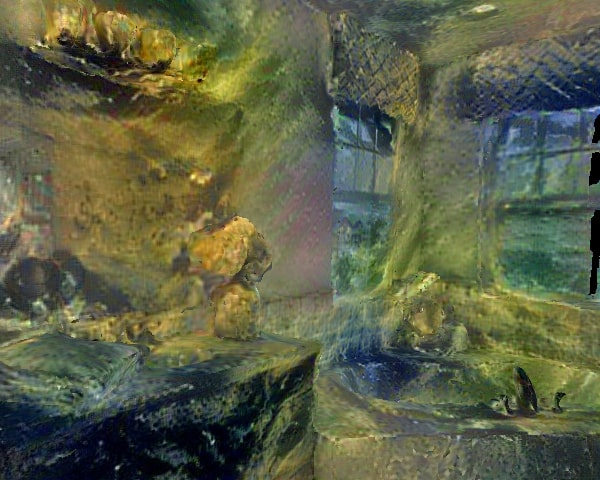} & 
\includegraphics[width=0.24\linewidth]{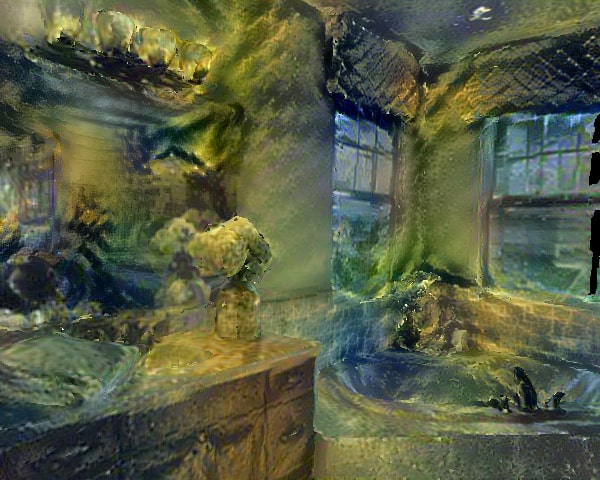} \\
(a) RGB and Style & (b) Only 2D & (c) With Angle & (d) With Angle and Depth
\end{tabular}
\caption{Qualitative ablation study of our method. We compare ours (d) against only using angle (c) and not using angle and depth (b). Using angle better distinguishes surfaces and using depth creates smaller/detailed stylization in the background.}
\label{fig:ablation}
\end{figure*}

\mypar{Qualitative Comparison}
Our method uses per-pixel angle and depth as input to optimize the texture in a 3D-aware manner.
This helps avoid view-dependent stretch and size artifacts being optimized into the texture from different poses.
We compare only using angle input (no render pyramid) and not using angle/depth (only 2D texture optimization with Laplacian Pyramid representation).
In~\cref{fig:ablation} we can see that using angle makes it easier to distinguish between surfaces like the wall and sofa in row 2.
Adding depth creates smaller and detailed patterns in the background (e.g., the strokes in the background of row 1).
Please see the supplemental material for more examples.

We optimize all ablation modes such that stylization patterns are equally strong, i.e., style should be similar for a fair comparison.
A too low degree of stylization would reduce view-dependent artifacts because original RGB colors get more dominant.
Similarly, a too high degree discards content features too much, which increases artifacts.

\begin{figure}
\centering
\setlength\tabcolsep{1pt}
\begin{tabular}{ccc}
\includegraphics[width=0.33\linewidth]{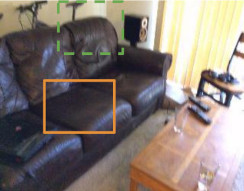} & 
\includegraphics[width=0.33\linewidth]{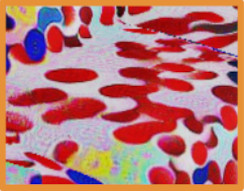} & 
\includegraphics[width=0.33\linewidth]{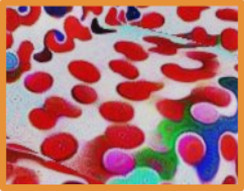} \\
RGB & (a) Only 2D & (b) Ours \\
\includegraphics[width=0.33\linewidth]{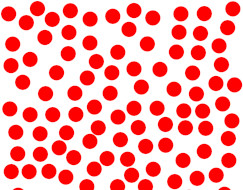} & 
\includegraphics[width=0.33\linewidth]{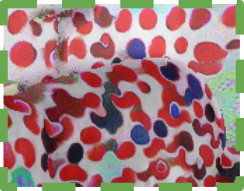} & 
\includegraphics[width=0.33\linewidth]{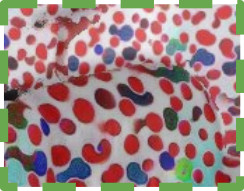} \\
Style & (c) Only 2D & (d) Ours \\
\end{tabular}
\caption{Stylizing with a ``circle'' image quantifies 3D-awareness. Circles become ellipsoidal on the couch without angle-awareness (a); using angle, they are less distorted (b). Using depth, circles are smaller in the back (d); they are equally large without (c).}
\label{fig:circles}
\end{figure}

\mypar{Quantitative Comparison}
We measure the effects of angle- and depth-awareness as follows.
We stylize a scene with a ``circle'' image using only the style loss (see~\cref{fig:circles}).
We then detect ellipses in the resulting images and measure their horizontal and vertical axis lengths.
Naturally, NST creates ellipses of different shapes, but their overall distribution reveals the degree of 3D awareness for the complete scene.
Inverse correlation between per-pixel depth and ellipse size in screen-space (Corr. 2D) indicates that stylized features are smaller in the background.
A weak correlation in world-space (Corr. 3D) indicates that absolute size is independent of the observed poses.
Both metrics together classify the depth-awareness.
View-dependent stretch is larger if ellipse's horizontal and vertical axes are of different lengths.
The stylization is angle-aware if the stretch is reduced.
We do not measure coarse and fine stylization this way, because the ``circle'' image contains too few high-resolution features.
Please see the supplemental material for more details about metric computation.
As can be seen in~\cref{tab:ablation}, using angle and depth improves our method.

\begin{table}
  \centering
  \begin{tabular}{|l|cc|c|}
    \hline
    Method & Corr. 2D $\uparrow$ & Corr. 3D $\downarrow$ & Stretch $\downarrow$ \\
    \hline
    \hline
    Only 2D & 0.172 & 0.126 & 3.512 \\
    Angle & 0.126 & \textbf{0.110} & 3.396 \\
    Angle/Depth & \textbf{0.538} & 0.125 & \textbf{3.391} \\
    \hline
  \end{tabular}
  \caption{Quantitative results of stylizing a scene with a ``circle'' image.
  With depth-awareness (Angle/Depth), depth and size strongly correlate in screen-space (2D), but not in world-space (3D). 
  That is, the stylized circle size is smaller in the background of an image, but uniformly distributed in 3D. Adding angle-awareness reduces the circle stretch (Angle).
  In contrast, circles show more view-dependent artifacts without depth/angle (Only 2D).
  }\label{tab:ablation}
\end{table}

\begin{figure}
\centering
\setlength\tabcolsep{1pt}
\begin{tabular}{cc}
\includegraphics[width=0.49\linewidth]{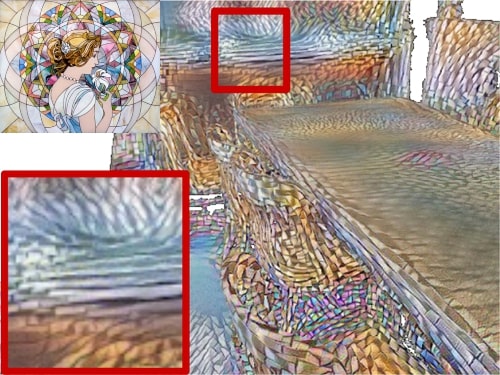} & 
\includegraphics[width=0.49\linewidth]{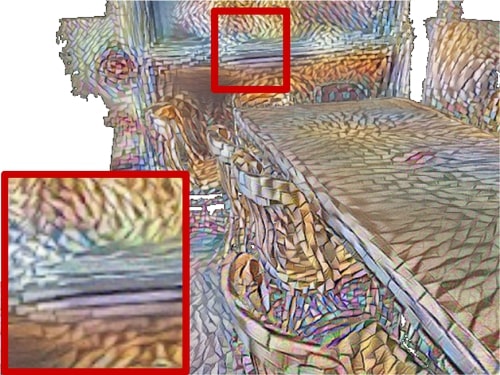} \\
$\theta_d = 0.05$ & $\theta_d = 0.1$ \\
\includegraphics[width=0.49\linewidth]{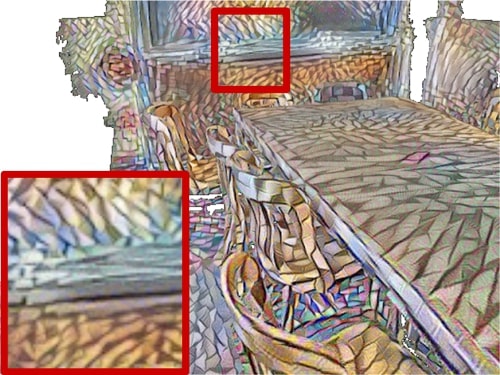} & 
\includegraphics[width=0.49\linewidth]{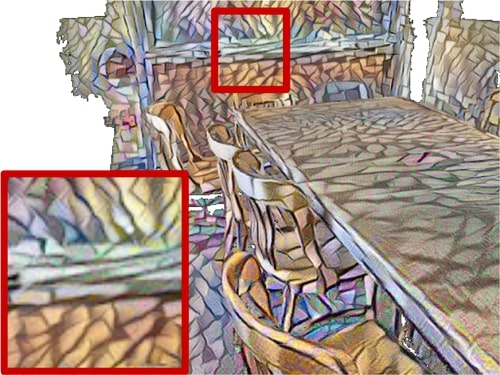} \\
$\theta_d = 0.15$ & $\theta_d = 0.2$ \\
\end{tabular}
\caption{Variation of the minimum depth $\theta_d$ from $0.05$ to $0.2$ meters. Increasing its value leads to overall larger stylizations, while still retaining relative size differences within the scene. This can be used to fine-tune the stylization to a satisfying look.}
\label{fig:depth-slider}
\end{figure}

\mypar{Depth Scaling}
A key piece of our method is the render pyramid of different image resolutions.
By tuning the value of $\theta_d$, we change the threshold of when to sample from the next higher resolution.
This increases (higher $\theta_d$) or decreases (lower $\theta_d$) the absolute stylization size, while still retaining relative change in size (see~\cref{fig:depth-slider}).
This allows to fine-tune the complete scene until a desired look is obtained.

\begin{figure}
\centering
\setlength\tabcolsep{1pt}
\begin{tabular}{cc}
\includegraphics[width=0.49\linewidth]{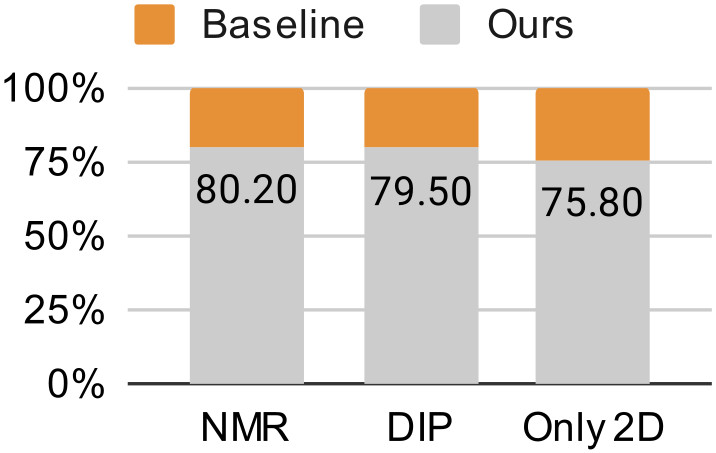} & 
\includegraphics[width=0.49\linewidth]{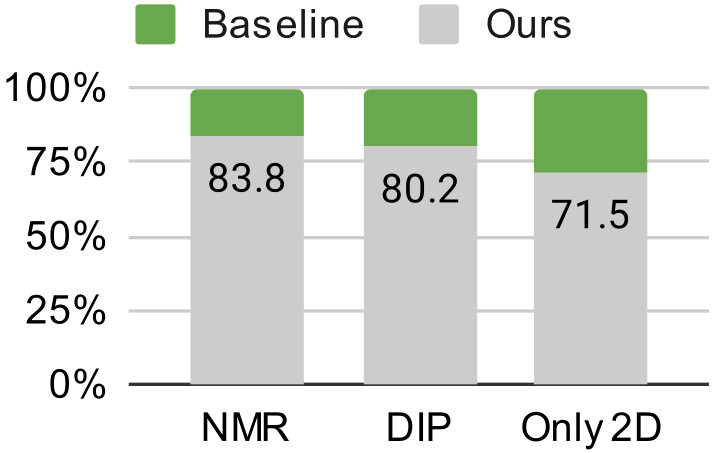} \\
(a) Stretch & (b) Size \\
\end{tabular}
\caption{We conduct a user study and ask subjects to select the results where stylization patterns (a) have less visible stretch and (b) are smaller in the background. The numbers indicate the percentage of preference for our method.}
\label{fig:user-study-ablation}
\end{figure}

\mypar{User Study}
We conduct a user study on the effectiveness of our proposed depth- and angle-awareness.
Users compared our method against each baseline separately by preferring one of two images.
They judged in which image stylization patterns (a) have less visible stretch and (b) are smaller in the background.
In total, 20 users each answered 70 questions, comparing against NMR~\cite{kato2018neural}, DIP~\cite{mordvintsev2018differentiable} and ours without angle- and depth-awareness (Only 2D).
As can be seen in~\cref{fig:user-study-ablation}, our method is preferred in both categories.

\subsection{Comparison to Video Style Transfer}\label{subsec:comp-video}

\begin{figure}
\centering
\setlength\tabcolsep{1pt}
\begin{tabular}{cccc}
\includegraphics[width=0.24\linewidth]{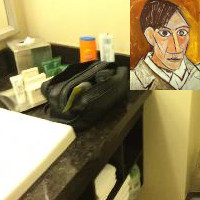} & 
\includegraphics[width=0.24\linewidth]{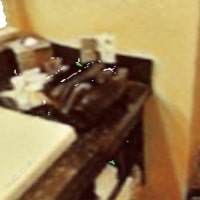} & 
\includegraphics[width=0.24\linewidth]{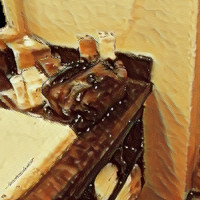} & 
\includegraphics[width=0.24\linewidth]{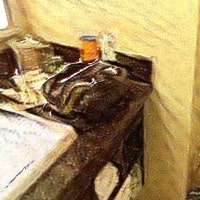} \\
\includegraphics[width=0.24\linewidth]{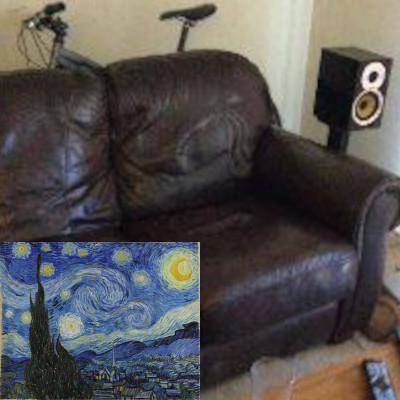} & 
\includegraphics[width=0.24\linewidth]{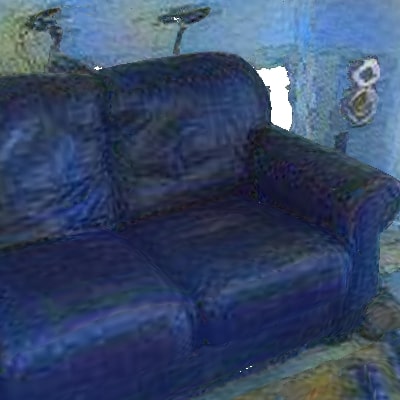} & 
\includegraphics[width=0.24\linewidth]{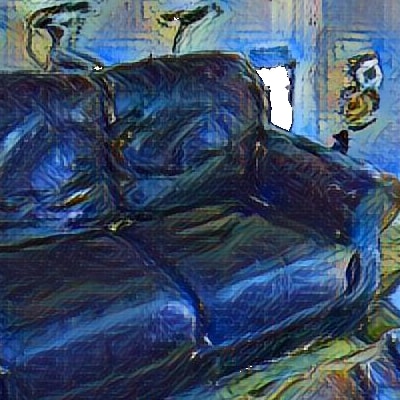} & 
\includegraphics[width=0.24\linewidth]{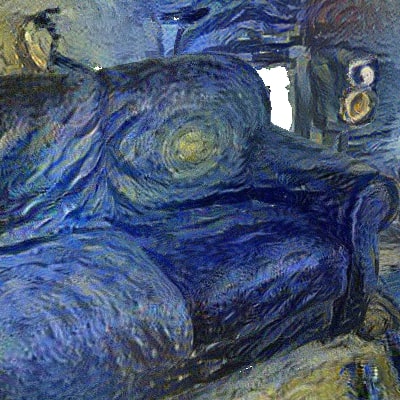} \\
RGB & (a) \footnotesize{VST$\rightarrow$Tex} & (b) \footnotesize{Tex$\rightarrow$VST} & Ours
\end{tabular}
\caption{Video style transfer (VST) can be combined with RGB texture mapping to stylize an indoor scene. Either, we optimize a texture from the output of VST (a), or we render trajectories from an RGB texture and then apply VST (b). The first row uses VST of Wang \etal~\cite{wang2020consistent} and the second Deng \etal~\cite{deng2020arbitrary}, respectively. In comparison, we produce sharper details and less noise.}
\label{fig:blurry-texture}
\end{figure}

As an alternative way to optimizing a stylized texture, one could combine video style transfer (VST) methods and RGB texture mapping to produce a stylized scene in two steps (see~\cref{fig:blurry-texture}).
We can obtain an RGB texture from all images of the scene and render arbitrary trajectories, that we stylize with a VST method (Tex$\rightarrow$VST).
However, we never obtain a stylized texture this way and thus need the VST method during inference for each novel pose.
Stylization details are also much lower, due to missing details in the RGB texture and reconstructed geometry.
By optimizing directly from camera images, we obtain sharper details.

Alternatively, we can stylize a trajectory of camera images with a VST method and optimize an RGB texture from these images (VST$\rightarrow$Tex).
However, we might only have access to a sparse set of images in some scenarios.
Due to inconsistencies between stylized frames (e.g., caused by illumination changes), the optimized texture is blurrier, as well.
Our method is 3D-consistent by combining stylization and texture optimization over all available images directly.

\subsection{Runtime Comparison}\label{subsec:Runtime Comparison}
We propose an optimization-based NST method, that converges in roughly 3 hours on a single RTX 3090 GPU.
After optimization, we can use the texture in traditional graphics pipelines and achieve real-time rendering, similar to~\cite{kato2018neural, mordvintsev2018differentiable}.
In contrast, model-based NST~\cite{huang_2021_3d_scene_stylization, kato2018neural, kopanas2021point} might take days to train and needs a forward-pass at inference.
However, these methods can generalize across scenes, whereas we need to optimize a separate texture per-scene.

\subsection{Limitations}
\label{subsec:limitations}
By design, our method is a per-scene/per-style NST algorithm, i.e., we optimize each explicit texture image separately.
Recent work in implicit texture representations~\cite{oechsle2019texture} could enable training generative models for our task.
We do not disentangle lighting and albedo, i.e., view-dependent effects in camera images can be visible in the stylized texture.
One could leverage neural rendering techniques to train a relightable stylization model~\cite{tewari2021advances}.
Incomplete mesh reconstructions lead to holes in rendered poses, which can be reduced by employing mesh completion techniques first~\cite{dai2021spsg, dai2020sg, dai2018scancomplete}.
Similarly, an insufficient number of poses may lead to unobserved surfaces during optimization, i.e., we do not hallucinate texture.
Inpainting techniques~\cite{tewari2021advances} could be utilized to complete those texels.

\section{Conclusion}
\label{sec:conclusion}
We have shown a method to stylize the mesh of room-scale indoor scene reconstructions.
We lift style transfer to the 3D domain by optimizing a texture only through 2D images.
Our method makes use of depth and surface normals of the mesh to achieve uniform world space stylization without view-dependent artifacts.
For that, we split the loss calculation into image parts and stylize coarse and fine details separately.
The explicit texture representation allows for real-time rendering of the scene after optimization.

\section*{Acknowledgements}
\label{sec:acknowledgements}
This project is funded by a TUM-IAS Rudolf Mößbauer Fellowship, the ERC Starting Grant Scan2CAD (804724), and the German Research Foundation (DFG) Grant Making Machine Learning on Static and Dynamic 3D Data Practical.
We also thank Angela Dai for the video voice-over.

{\small
\bibliographystyle{ieee_fullname}
\bibliography{Paper}

\begin{thebibliography}{10}\itemsep=-1pt

\bibitem{avetisyan2019scan2cad}
Armen Avetisyan, Manuel Dahnert, Angela Dai, Manolis Savva, Angel~X Chang, and
  Matthias Nie{\ss}ner.
\newblock Scan2cad: Learning cad model alignment in rgb-d scans.
\newblock In {\em Proceedings of the IEEE/CVF Conference on Computer Vision and
  Pattern Recognition}, pages 2614--2623, 2019.

\bibitem{bi2017patch}
Sai Bi, Nima~Khademi Kalantari, and Ravi Ramamoorthi.
\newblock Patch-based optimization for image-based texture mapping.
\newblock {\em ACM Trans. Graph.}, 36(4):106--1, 2017.

\bibitem{opencv_library}
G. Bradski.
\newblock {The OpenCV Library}.
\newblock {\em Dr. Dobb's Journal of Software Tools}, 2000.

\bibitem{cao2020psnet}
Xu Cao, Weimin Wang, Katashi Nagao, and Ryosuke Nakamura.
\newblock Psnet: A style transfer network for point cloud stylization on
  geometry and color.
\newblock In {\em Proceedings of the IEEE/CVF Winter Conference on Applications
  of Computer Vision}, pages 3337--3345, 2020.

\bibitem{chang2017matterport3d}
Angel Chang, Angela Dai, Thomas Funkhouser, Maciej Halber, Matthias Niessner,
  Manolis Savva, Shuran Song, Andy Zeng, and Yinda Zhang.
\newblock Matterport3d: Learning from rgb-d data in indoor environments.
\newblock {\em arXiv preprint arXiv:1709.06158}, 2017.

\bibitem{chen2017coherent}
Dongdong Chen, Jing Liao, Lu Yuan, Nenghai Yu, and Gang Hua.
\newblock Coherent online video style transfer.
\newblock In {\em Proceedings of the IEEE International Conference on Computer
  Vision}, pages 1105--1114, 2017.

\bibitem{chen2020optical}
Xinghao Chen, Yiman Zhang, Yunhe Wang, Han Shu, Chunjing Xu, and Chang Xu.
\newblock Optical flow distillation: Towards efficient and stable video style
  transfer.
\newblock In {\em European Conference on Computer Vision}, pages 614--630.
  Springer, 2020.

\bibitem{chiang2021stylizing}
Pei-Ze Chiang, Meng-Shiun Tsai, Hung-Yu Tseng, Wei sheng Lai, and Wei-Chen
  Chiu.
\newblock Stylizing 3d scene via implicit representation and hypernetwork,
  2021.

\bibitem{chiu2020iterative}
Tai-Yin Chiu and Danna Gurari.
\newblock Iterative feature transformation for fast and versatile universal
  style transfer.
\newblock In {\em European Conference on Computer Vision}, pages 169--184.
  Springer, 2020.

\bibitem{blender}
Blender~Online Community.
\newblock {\em Blender - a 3D modelling and rendering package}.
\newblock Blender Foundation, Stichting Blender Foundation, Amsterdam, 2018.

\bibitem{dai2017scannet}
Angela Dai, Angel~X Chang, Manolis Savva, Maciej Halber, Thomas Funkhouser, and
  Matthias Nie{\ss}ner.
\newblock Scannet: Richly-annotated 3d reconstructions of indoor scenes.
\newblock In {\em Proceedings of the IEEE Conference on Computer Vision and
  Pattern Recognition}, pages 5828--5839, 2017.

\bibitem{dai2020sg}
Angela Dai, Christian Diller, and Matthias Nie{\ss}ner.
\newblock Sg-nn: Sparse generative neural networks for self-supervised scene
  completion of rgb-d scans.
\newblock In {\em Proceedings of the IEEE/CVF Conference on Computer Vision and
  Pattern Recognition}, pages 849--858, 2020.

\bibitem{dai2017bundle}
Angela Dai, Matthias Nie{\ss}ner, Michael Zoll{\"o}fer, Shahram Izadi, and
  Christian Theobalt.
\newblock Bundlefusion: Real-time globally consistent 3d reconstruction using
  on-the-fly surface re-integration.
\newblock {\em ACM Transactions on Graphics 2017 (TOG)}, 2017.

\bibitem{dai2018scancomplete}
Angela Dai, Daniel Ritchie, Martin Bokeloh, Scott Reed, J{\"u}rgen Sturm, and
  Matthias Nie{\ss}ner.
\newblock Scancomplete: Large-scale scene completion and semantic segmentation
  for 3d scans.
\newblock In {\em Proceedings of the IEEE Conference on Computer Vision and
  Pattern Recognition}, pages 4578--4587, 2018.

\bibitem{dai2021spsg}
Angela Dai, Yawar Siddiqui, Justus Thies, Julien Valentin, and Matthias
  Nie{\ss}ner.
\newblock Spsg: Self-supervised photometric scene generation from rgb-d scans.
\newblock In {\em Proceedings of the IEEE/CVF Conference on Computer Vision and
  Pattern Recognition}, pages 1747--1756, 2021.

\bibitem{deng2020arbitrary}
Yingying Deng, Fan Tang, Weiming Dong, Haibin Huang, Chongyang Ma, and
  Changsheng Xu.
\newblock Arbitrary video style transfer via multi-channel correlation.
\newblock {\em arXiv preprint arXiv:2009.08003}, 2020.

\bibitem{dessein2014seamless}
Arnaud Dessein, William~AP Smith, Richard~C Wilson, and Edwin~R Hancock.
\newblock Seamless texture stitching on a 3d mesh by poisson blending in
  patches.
\newblock In {\em 2014 IEEE International Conference on Image Processing
  (ICIP)}, pages 2031--2035. IEEE, 2014.

\bibitem{foley1996computer}
James~D Foley, Foley~Dan Van, Andries Van~Dam, Steven~K Feiner, John~F Hughes,
  and J Hughes.
\newblock {\em Computer graphics: principles and practice}, volume 12110.
\newblock Addison-Wesley Professional, 1996.

\bibitem{gao2018reconet}
Chang Gao, Derun Gu, Fangjun Zhang, and Yizhou Yu.
\newblock Reconet: Real-time coherent video style transfer network.
\newblock In {\em Asian Conference on Computer Vision}, pages 637--653.
  Springer, 2018.

\bibitem{gao2020fast}
Wei Gao, Yijun Li, Yihang Yin, and Ming-Hsuan Yang.
\newblock Fast video multi-style transfer.
\newblock In {\em Proceedings of the IEEE/CVF Winter Conference on Applications
  of Computer Vision}, pages 3222--3230, 2020.

\bibitem{Gatys_2016_CVPR}
Leon~A. Gatys, Alexander~S. Ecker, and Matthias Bethge.
\newblock Image style transfer using convolutional neural networks.
\newblock In {\em Proceedings of the IEEE Conference on Computer Vision and
  Pattern Recognition (CVPR)}, June 2016.

\bibitem{gatys2017controlling}
Leon~A Gatys, Alexander~S Ecker, Matthias Bethge, Aaron Hertzmann, and Eli
  Shechtman.
\newblock Controlling perceptual factors in neural style transfer.
\newblock In {\em Proceedings of the IEEE Conference on Computer Vision and
  Pattern Recognition}, pages 3985--3993, 2017.

\bibitem{gupta2017characterizing}
Agrim Gupta, Justin Johnson, Alexandre Alahi, and Li Fei-Fei.
\newblock Characterizing and improving stability in neural style transfer.
\newblock In {\em Proceedings of the IEEE International Conference on Computer
  Vision}, 2017.

\bibitem{han2021exemplar}
Fangzhou Han, Shuquan Ye, Mingming He, Menglei Chai, and Jing Liao.
\newblock Exemplar-based 3d portrait stylization.
\newblock {\em arXiv preprint arXiv:2104.14559}, 2021.

\bibitem{Hauptfleisch20-PG}
Filip Hauptfleisch, Ond\v{r}ej Texler, Aneta Texler, Jaroslav K\v{r}iv\'{a}nek,
  and Daniel S\'{y}kora.
\newblock {StyleProp}: {Real}-time example-based stylization of 3d models.
\newblock {\em Computer Graphics Forum}, 39(7):575--586, 2020.

\bibitem{huang_2021_3d_scene_stylization}
Hsin-Ping Huang, Hung-Yu Tseng, Saurabh Saini, Maneesh Singh, and Ming-Hsuan
  Yang.
\newblock Learning to stylize novel views.
\newblock {\em arXiv preprint arXiv:2105.13509}, 2021.

\bibitem{huang20173dlite}
Jingwei Huang, Angela Dai, Leonidas~J Guibas, and Matthias Nie{\ss}ner.
\newblock 3dlite: towards commodity 3d scanning for content creation.
\newblock {\em ACM Trans. Graph.}, 36(6):203--1, 2017.

\bibitem{huang2020adversarial}
Jingwei Huang, Justus Thies, Angela Dai, Abhijit Kundu, Chiyu Jiang, Leonidas~J
  Guibas, Matthias Nie{\ss}ner, Thomas Funkhouser, et~al.
\newblock Adversarial texture optimization from rgb-d scans.
\newblock In {\em Proceedings of the IEEE/CVF Conference on Computer Vision and
  Pattern Recognition}, pages 1559--1568, 2020.

\bibitem{huang2017arbitrary}
Xun Huang and Serge Belongie.
\newblock Arbitrary style transfer in real-time with adaptive instance
  normalization.
\newblock In {\em Proceedings of the IEEE International Conference on Computer
  Vision}, pages 1501--1510, 2017.

\bibitem{izadi2011kinectfusion}
Shahram Izadi, David Kim, Otmar Hilliges, David Molyneaux, Richard Newcombe,
  Pushmeet Kohli, Jamie Shotton, Steve Hodges, Dustin Freeman, Andrew Davison,
  et~al.
\newblock Kinectfusion: real-time 3d reconstruction and interaction using a
  moving depth camera.
\newblock In {\em Proceedings of the 24th annual ACM symposium on User
  interface software and technology}, pages 559--568, 2011.

\bibitem{johnson2016perceptual}
Justin Johnson, Alexandre Alahi, and Li Fei-Fei.
\newblock Perceptual losses for real-time style transfer and super-resolution.
\newblock In {\em European conference on computer vision}, pages 694--711.
  Springer, 2016.

\bibitem{kalischek2021light}
Nikolai Kalischek, Jan~D Wegner, and Konrad Schindler.
\newblock In the light of feature distributions: moment matching for neural
  style transfer.
\newblock In {\em Proceedings of the IEEE/CVF Conference on Computer Vision and
  Pattern Recognition}, pages 9382--9391, 2021.

\bibitem{kato2018neural}
Hiroharu Kato, Yoshitaka Ushiku, and Tatsuya Harada.
\newblock Neural 3d mesh renderer.
\newblock In {\em Proceedings of the IEEE conference on computer vision and
  pattern recognition}, pages 3907--3916, 2018.

\bibitem{kingma2014adam}
Diederik~P Kingma and Jimmy Ba.
\newblock Adam: A method for stochastic optimization.
\newblock {\em arXiv preprint arXiv:1412.6980}, 2014.

\bibitem{kolkin2019style}
Nicholas Kolkin, Jason Salavon, and Gregory Shakhnarovich.
\newblock Style transfer by relaxed optimal transport and self-similarity.
\newblock In {\em Proceedings of the IEEE/CVF Conference on Computer Vision and
  Pattern Recognition}, pages 10051--10060, 2019.

\bibitem{kopanas2021point}
Georgios Kopanas, Julien Philip, Thomas Leimk{\"u}hler, and George Drettakis.
\newblock Point-based neural rendering with per-view optimization.
\newblock In {\em Computer Graphics Forum}, volume~40, 2021.

\bibitem{li2016combining}
Chuan Li and Michael Wand.
\newblock Combining markov random fields and convolutional neural networks for
  image synthesis.
\newblock In {\em Proceedings of the IEEE conference on computer vision and
  pattern recognition}, pages 2479--2486, 2016.

\bibitem{li2019learning}
Xueting Li, Sifei Liu, Jan Kautz, and Ming-Hsuan Yang.
\newblock Learning linear transformations for fast image and video style
  transfer.
\newblock In {\em Proceedings of the IEEE/CVF Conference on Computer Vision and
  Pattern Recognition}, pages 3809--3817, 2019.

\bibitem{liu2017depth}
Xiao-Chang Liu, Ming-Ming Cheng, Yu-Kun Lai, and Paul~L Rosin.
\newblock Depth-aware neural style transfer.
\newblock In {\em Proceedings of the Symposium on Non-Photorealistic Animation
  and Rendering}, pages 1--10, 2017.

\bibitem{liu2020geometric}
Xiao-Chang Liu, Xuan-Yi Li, Ming-Ming Cheng, and Peter Hall.
\newblock Geometric style transfer.
\newblock {\em arXiv preprint arXiv:2007.05471}, 2020.

\bibitem{luo2016understanding}
Wenjie Luo, Yujia Li, Raquel Urtasun, and Richard Zemel.
\newblock Understanding the effective receptive field in deep convolutional
  neural networks.
\newblock In {\em Proceedings of the 30th International Conference on Neural
  Information Processing Systems}, pages 4905--4913, 2016.

\bibitem{mechrez2018contextual}
Roey Mechrez, Itamar Talmi, and Lihi Zelnik-Manor.
\newblock The contextual loss for image transformation with non-aligned data.
\newblock In {\em Proceedings of the European Conference on Computer Vision
  (ECCV)}, pages 768--783, 2018.

\bibitem{mordvintsev2018differentiable}
Alexander Mordvintsev, Nicola Pezzotti, Ludwig Schubert, and Chris Olah.
\newblock Differentiable image parameterizations.
\newblock {\em Distill}, 3(7):e12, 2018.

\bibitem{newcombe2011kinectfusion}
Richard~A Newcombe, Shahram Izadi, Otmar Hilliges, David Molyneaux, David Kim,
  Andrew~J Davison, Pushmeet Kohi, Jamie Shotton, Steve Hodges, and Andrew
  Fitzgibbon.
\newblock Kinectfusion: Real-time dense surface mapping and tracking.
\newblock In {\em 2011 10th IEEE international symposium on mixed and augmented
  reality}, pages 127--136. IEEE, 2011.

\bibitem{niessner2013real}
Matthias Nie{\ss}ner, Michael Zollh{\"o}fer, Shahram Izadi, and Marc
  Stamminger.
\newblock Real-time 3d reconstruction at scale using voxel hashing.
\newblock {\em ACM Transactions on Graphics (ToG)}, 32(6):1--11, 2013.

\bibitem{oechsle2019texture}
Michael Oechsle, Lars Mescheder, Michael Niemeyer, Thilo Strauss, and Andreas
  Geiger.
\newblock Texture fields: Learning texture representations in function space.
\newblock In {\em Proceedings of the IEEE/CVF International Conference on
  Computer Vision}, pages 4531--4540, 2019.

\bibitem{ruder2016artistic}
Manuel Ruder, Alexey Dosovitskiy, and Thomas Brox.
\newblock Artistic style transfer for videos.
\newblock In {\em German conference on pattern recognition}, pages 26--36.
  Springer, 2016.

\bibitem{ruder2018artistic}
Manuel Ruder, Alexey Dosovitskiy, and Thomas Brox.
\newblock Artistic style transfer for videos and spherical images.
\newblock {\em International Journal of Computer Vision}, 126(11):1199--1219,
  2018.

\bibitem{simonyan2014very}
Karen Simonyan and Andrew Zisserman.
\newblock Very deep convolutional networks for large-scale image recognition.
\newblock {\em arXiv preprint arXiv:1409.1556}, 2014.

\bibitem{Sykora19-EG}
Daniel S\'{y}kora, Ond\v{r}ej Jamri\v{s}ka, Ond\v{r}ej Texler, Jakub Fi\v{s}er,
  Michal Luk\'{a}\v{c}, Jingwan Lu, and Eli Shechtman.
\newblock {StyleBlit}: Fast example-based stylization with local guidance.
\newblock {\em Computer Graphics Forum}, 38(2):83--91, 2019.

\bibitem{tewari2021advances}
Ayush Tewari, O Fried, J Thies, V Sitzmann, S Lombardi, Z Xu, T Simon, M
  Nie{\ss}ner, E Tretschk, L Liu, et~al.
\newblock Advances in neural rendering.
\newblock In {\em ACM SIGGRAPH 2021 Courses}, pages 1--320. 2021.

\bibitem{thies2019deferred}
Justus Thies, Michael Zollh{\"o}fer, and Matthias Nie{\ss}ner.
\newblock Deferred neural rendering: Image synthesis using neural textures.
\newblock {\em ACM Transactions on Graphics (TOG)}, 38(4):1--12, 2019.

\bibitem{ulyanov2016texture}
Dmitry Ulyanov, Vadim Lebedev, Andrea Vedaldi, and Victor~S Lempitsky.
\newblock Texture networks: Feed-forward synthesis of textures and stylized
  images.
\newblock In {\em ICML}, 2016.

\bibitem{waechter2014let}
Michael Waechter, Nils Moehrle, and Michael Goesele.
\newblock Let there be color! large-scale texturing of 3d reconstructions.
\newblock In {\em European conference on computer vision}, pages 836--850.
  Springer, 2014.

\bibitem{wang2020consistent}
Wenjing Wang, Shuai Yang, Jizheng Xu, and Jiaying Liu.
\newblock Consistent video style transfer via relaxation and regularization.
\newblock {\em IEEE Transactions on Image Processing}, 29:9125--9139, 2020.

\bibitem{xia2021real}
Xide Xia, Tianfan Xue, Wei-sheng Lai, Zheng Sun, Abby Chang, Brian Kulis, and
  Jiawen Chen.
\newblock Real-time localized photorealistic video style transfer.
\newblock In {\em Proceedings of the IEEE/CVF Winter Conference on Applications
  of Computer Vision}, pages 1089--1098, 2021.

\bibitem{yin20213dstylenet}
Kangxue Yin, Jun Gao, Maria Shugrina, Sameh Khamis, and Sanja Fidler.
\newblock 3dstylenet: Creating 3d shapes with geometric and texture style
  variations.
\newblock In {\em Proceedings of the IEEE/CVF International Conference on
  Computer Vision}, pages 12456--12465, 2021.

\bibitem{zhou2014color}
Qian-Yi Zhou and Vladlen Koltun.
\newblock Color map optimization for 3d reconstruction with consumer depth
  cameras.
\newblock {\em ACM Transactions on Graphics (TOG)}, 33(4):1--10, 2014.

\bibitem{zhu1997algorithm}
Ciyou Zhu, Richard~H Byrd, Peihuang Lu, and Jorge Nocedal.
\newblock Algorithm 778: L-bfgs-b: Fortran subroutines for large-scale
  bound-constrained optimization.
\newblock {\em ACM Transactions on mathematical software (TOMS)},
  23(4):550--560, 1997.

\end{thebibliography}
}

\appendix
\section{Supplemental}
\label{sec:supplemental}

\subsection{Reprojection Error}\label{subsec:Reprojection Error}
\begin{figure}[ht]
\centering
\setlength\tabcolsep{1pt}
\begin{tabular}{cccc}
\includegraphics[width=0.24\linewidth]{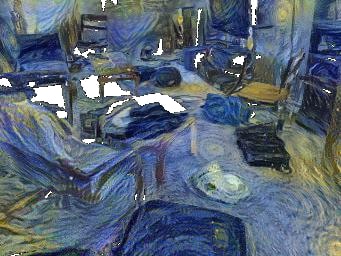} & 
\includegraphics[width=0.24\linewidth]{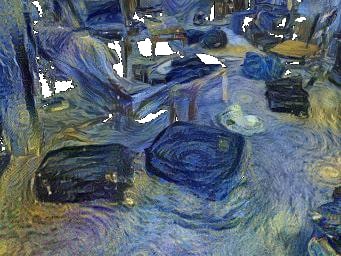} & 
\includegraphics[width=0.24\linewidth]{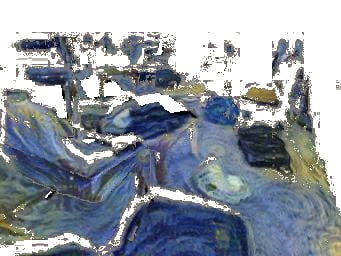} & 
\includegraphics[width=0.24\linewidth]{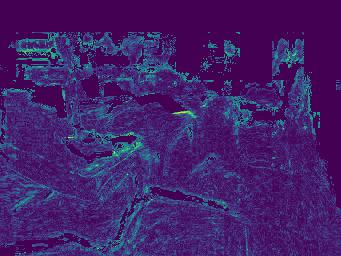} \\
Source Frame & Target Frame & Reprojected & Residual
\end{tabular}
\caption{Sample images used for calculating the reprojection error. We select a source frame and a target frame and warp the target pixels from the target view to the source view. Reprojection error is calculated as the $L_1$ distance of all pixels between source frame and reprojected target frame, that are visible in both views.}
\label{fig:reprojection}
\end{figure}

We calculate reprojection error as the $L_1$ distance between source frame and reprojected target frame.
This allows to quantify the 3D consistency.
While a texture representation is 3D consistent by design, unoptimized texels might still create visible noise artifacts when rendering a trajectory.
We calculate reprojection error on the ScanNet~\cite{dai2017scannet} dataset by using their captured camera trajectories.
For each source frame, we select a target frame that is (a) two frames after the source frame (short-range consistency) or (b) 20 frames after the source frame (long-range consistency).
Using the estimated poses and camera intrinsics, we warp the pixels of the target frame to the source view.
We calculate $L_1$ distance in the normalized image range $[0,1]$ between reprojected target frame and source frame for all pixels that are visible in both views.
Please see~\cref{fig:reprojection} for a visualization of the procedure.

Because the estimated poses are not perfectly accurate, the reprojection error may never sink below a certain threshold that captures this inaccuracy.
Still, it allows to quantify the 3D consistency by measuring the additional inconsistencies caused by unoptimized textures, i.e., the error is higher for unoptimized textures that are less consistent.

\subsection{User Study Setup}\label{subsec:user-study-setup}
\begin{figure*}[ht]
\centering
\setlength\tabcolsep{1pt}
\begin{tabular}{cc}
\includegraphics[width=0.49\linewidth]{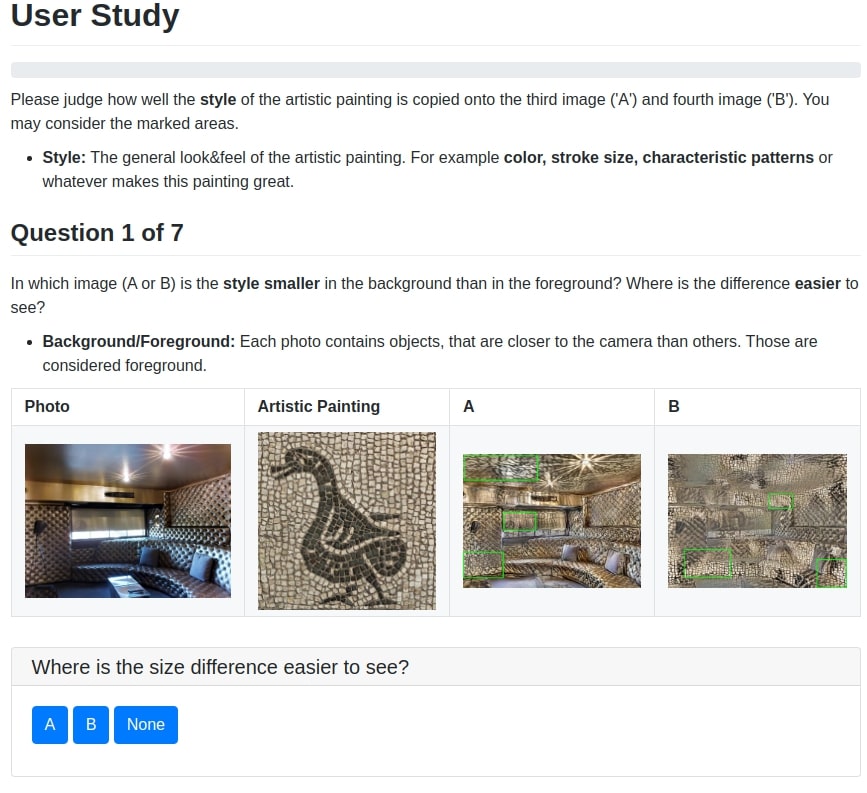} & 
\includegraphics[width=0.49\linewidth]{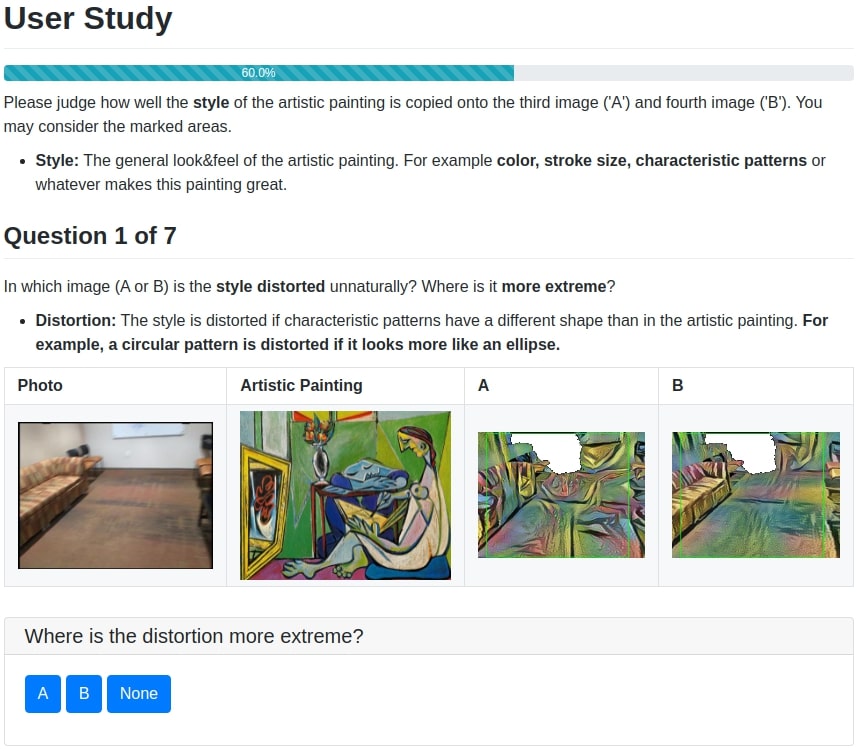} \\
(a) Size Sample & (b) Stretch Sample
\end{tabular}
\caption{Sample images used for the user study. Users judged in which image stylization patterns (a) are smaller in the background and (b) have less visible stretch.}
\label{fig:user-study-setup}
\end{figure*}

We conduct a user study on the effectiveness of our proposed depth- and angle-awareness.
Users compared our method against each baseline separately by preferring one of two images.
They judged in which image stylization patterns (a) have less visible stretch and (b) are smaller in the background.
In total, 20 users each answered 70 questions, comparing against NMR~\cite{kato2018neural}, DIP~\cite{mordvintsev2018differentiable} and ours without angle- and depth-awareness (Only 2D).
We show two sample questions, one for each type, in~\cref{fig:user-study-setup}.
As can be seen, users have the possibility to decide for one of two images or to answer that none of the two is better/worse.
The order of questions and of the ``A'' and ``B'' images is random and different for each user.
Users have the possibility to zoom-in on the images for better judgement.
Additionally, we add rectangles on image regions that might be especially interesting for evaluation of the questions.
Note that users still had to consider the whole image in their answer; the rectangles merely act as additional input.

\subsection{Variation of Depth Levels}\label{subsec:hparam-tuning}
\begin{figure}[!h]
\centering
\setlength\tabcolsep{1pt}
\begin{tabular}{cc}
\includegraphics[width=0.49\linewidth]{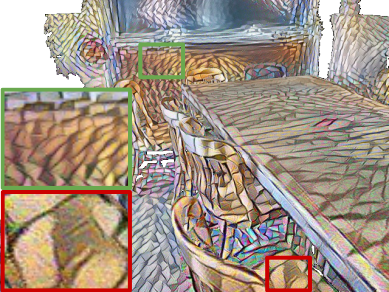} &
\includegraphics[width=0.49\linewidth]{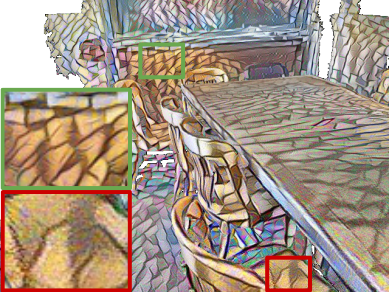} \\
(a) $\theta_l{=}4$ & (b) $\theta_l{=}2$
\end{tabular}
\caption{Variation of the number of depth levels $\theta_l$. Decreasing $\theta_l$ reduces size variation at different depths.}
\label{fig:rebuttal}
\end{figure}

The number of depth levels $\theta_l$ controls the depth variation that can be achieved within rendered poses of a scene.
Setting $\theta_l{=}4$ is sufficient for our datasets, as larger scene extent is rarely captured by many pixels.
We could precompute \textit{uv} maps at larger resolutions to enable depth scaling at even larger depth values.
Small scenes may not require the last layers, in which case they are simply not utilized during optimization.
Adding more layers in-between maps less pixels to one layer, which can yield insufficient Gram matrices and is computationally more expensive.
Decreasing $\theta_l$ reduces size variation at different depths (see \cref{fig:rebuttal}).

\subsection{Circle-Stretch and -Size Metric}\label{subsec:circle-detect}
We describe in more detail the metrics and principles used in the main paper to quantify the effects of our depth and angle awareness.
In order to measure the effects, we stylize a scene with a hand-crafted ``circle'' image (see main paper) and only use the (multi-resolution, part-based) style loss.
After optimization, the red circles are stylized all over the scene and are well-suited to describe the two drawbacks of missing angle- and depth-awareness.
For example, circles become ellipsoidal if a small grazing angle is used for stylization and circles change their radius inconsistently without depth awareness.
We can now measure the degree we alleviate these issues by measuring the \textit{size} and \textit{stretch} of the circles/ellipses.
Naturally, NST creates ellipses of different shapes, but their overall distribution reveals the degree of 3D awareness for the complete scene.

\subsubsection{Segmentation of Ellipses}
\begin{figure*}[ht]
\centering
\setlength\tabcolsep{1pt}
\begin{tabular}{ccc}
\includegraphics[width=0.32\linewidth]{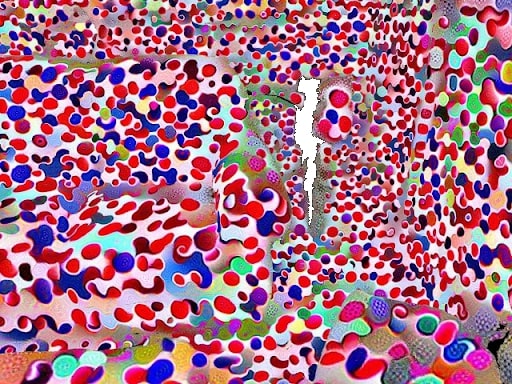} & 
\includegraphics[width=0.32\linewidth]{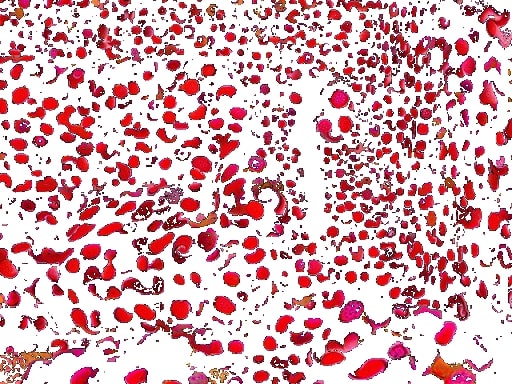} & 
\includegraphics[width=0.32\linewidth]{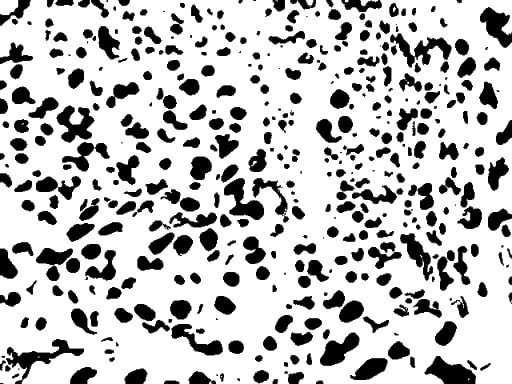} \\
Input & Filtered Red & Denoised Binary Mask \\
\includegraphics[width=0.32\linewidth]{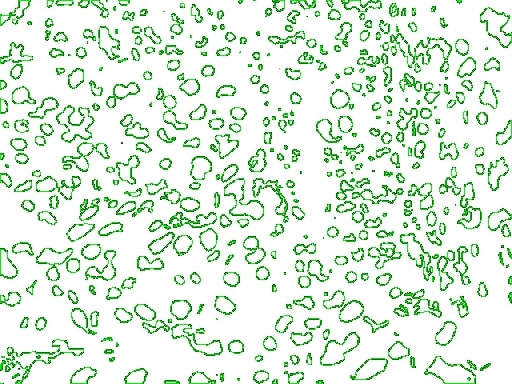} & 
\includegraphics[width=0.32\linewidth]{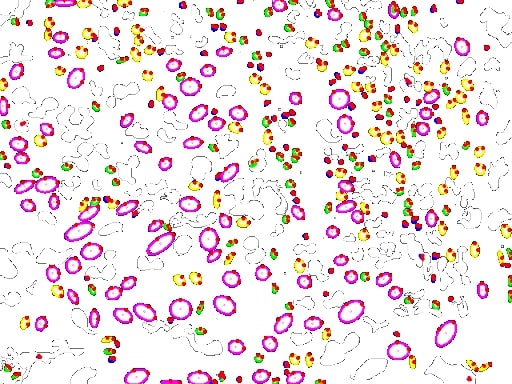} & 
\includegraphics[width=0.32\linewidth]{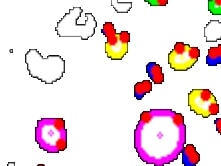} \\
Contour Detection & Detected Ellipses & Ellipses with Axis Points \\
\end{tabular}
\caption{Our pipeline for segmenting ellipses out of a stylized image for quantification of the depth and stretch effects. We filter the image to only contain red colors (Filtered Red) and then transform that to a denoised binary image to remove remaining pixels (Denoised Binary Mask). Afterwards we use contour detection to convert the binary mask into edge maps and finally fit ellipses to all applicable contours (Detected Ellipses). Lastly, we find the horizontal and vertical axis of the ellipses as the distance from the center to corresponding points on the edge of each ellipse.}
\label{fig:circles-detection}
\end{figure*}

First, we automatically segment red ellipses from each image of the stylized scene (see~\cref{fig:circles-detection}).
We first apply an HSV filter and only keep pixels in the ranges $0.6 \leq S, V \leq 1.0$, $0.0 \leq H \leq 0.08$ and $0.88 \leq H \leq 1.0$.
Then we turn the filtered image into a binary mask by thresholding colors above $0.15$ intensity and denoise it with OpenCV's ``fastNLMeansDenoising'' function~\cite{opencv_library}.
Afterwards, we use OpenCV's contour detection to get an edge map.
We filter out all contours with $max_d > 2$, where $max_d$ is the maximum deviation from a convex hull, as measured by ``convexityDefects''~\cite{opencv_library}.
We now fit ellipses to the remaining contours with ``fitEllipse''.
We extract the pixel-radius as
\begin{equation}
r_p = \frac{h_p + v_p}{2}
\end{equation}
for every fitted ellipse and calculate its pixel-stretch as
\begin{equation}
s_p = max(\frac{h_p}{v_p}, \frac{v_p}{h_p})
\end{equation}
where $h_p$ is the pixel-length of the horizontal ellipse radius and $v_p$ the vertical, respectively.
We remove the remaining wrongfully detected ellipses with $r_p < 10$, $r_p > 1000$ and $s_p > 10$ to get a result like in~\cref{fig:circles-detection}.
We use these ellipse characteristics to calculate metrics for depth- and angle-awareness.

\subsubsection{Calculation of Depth Metrics}
We calculate the correlation between per-pixel depth $d_{xy}$ and ellipse radius $r_p$ to quantify the effect of our depth-awareness in the 2D image plane (Corr. 2D).
For each detected ellipse we use the depth value of the pixel corresponding to the ellipse center.
A high negative correlation (e.g., $-0.5$) signals, that ellipse size decreases with increasing depth, whereas a low correlation (e.g., $-0.05$) signals, that ellipse size is independent of changes in depth.
A method that is able to stylize a scene depth-aware would create ellipses with smaller size in the background and thus have a high negative correlation in the 2D image plane.

To quantify the correlation in 3D, we backproject $h_p$ and $v_p$ to world-space using the estimated pose and camera intrinsics and calculate the world-space radius as
\begin{equation}
r_w = \frac{h_w + v_w}{2}
\end{equation}
where $h_w$ and $v_w$ are the backprojected axis lengths.
We then calculate the correlation between $r_w$ and the per-pixel depth $d_{xy}$ (Corr. 3D).
A high negative correlation (e.g., $-0.5$) signals, that ellipse size in world-space still decreases with increasing depth, whereas a low correlation (e.g., $-0.05$) signals, that ellipse size in world-space is independent of changes in depth.
A method that is able to stylize a scene depth-aware would create ellipses with uniformly distributed size in world-space (because the ellipse size should only change when rendering a scene from different poses, due to perspective projection).

Note that the stylized ellipses naturally vary in their sizes (e.g., ellipses can be smaller and larger independent of depth).
Therefore, the correlations will be precise up to a certain threshold.
However, the distribution of all segmented ellipses across the whole scene still allows to quantify the depth-awareness.

\subsubsection{Calculation of Angle Metric}
We backproject the pixel-stretch $s_p$ back to world-space as
\begin{equation}
s_w = max(\frac{h_w}{v_w}, \frac{v_w}{h_w})
\end{equation}
.
Then we calculate the arithmetic mean over all $s_w$ values for all detected ellipses.
A higher mean value means that overall we have more stretch, whereas a lower value signals a more uniform stylization result.
A method that is able to stylize a scene angle-aware would create ellipses with small stretch.

\subsection{Additional Qualitative Results}\label{subsec:baseline-comp-add}
We show additional qualitative results for our method.

\begin{figure*}
\centering
\setlength\tabcolsep{1pt}
\begin{tabular}{cccc}
\includegraphics[width=0.24\linewidth]{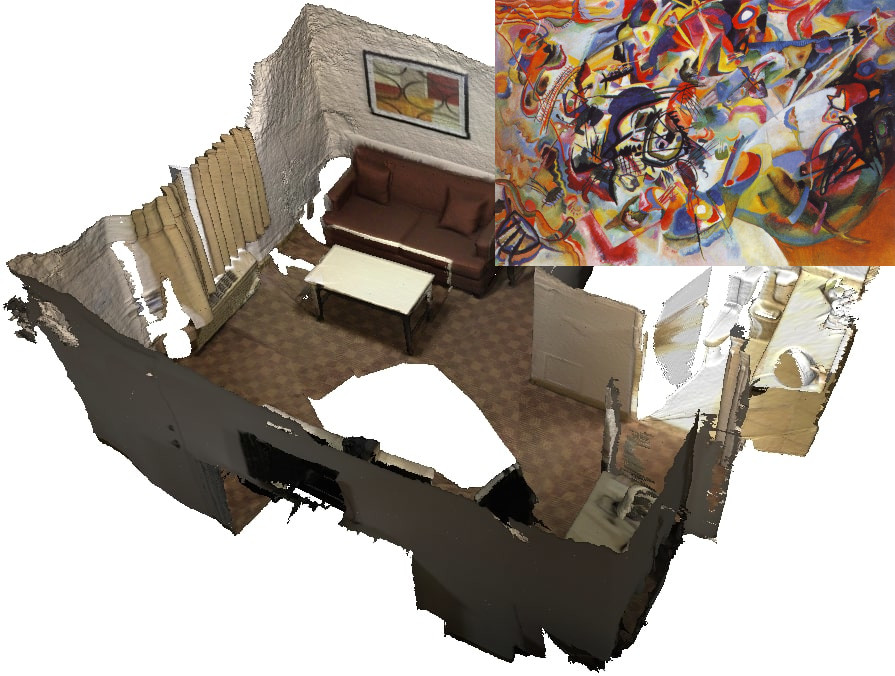} & 
\includegraphics[width=0.24\linewidth]{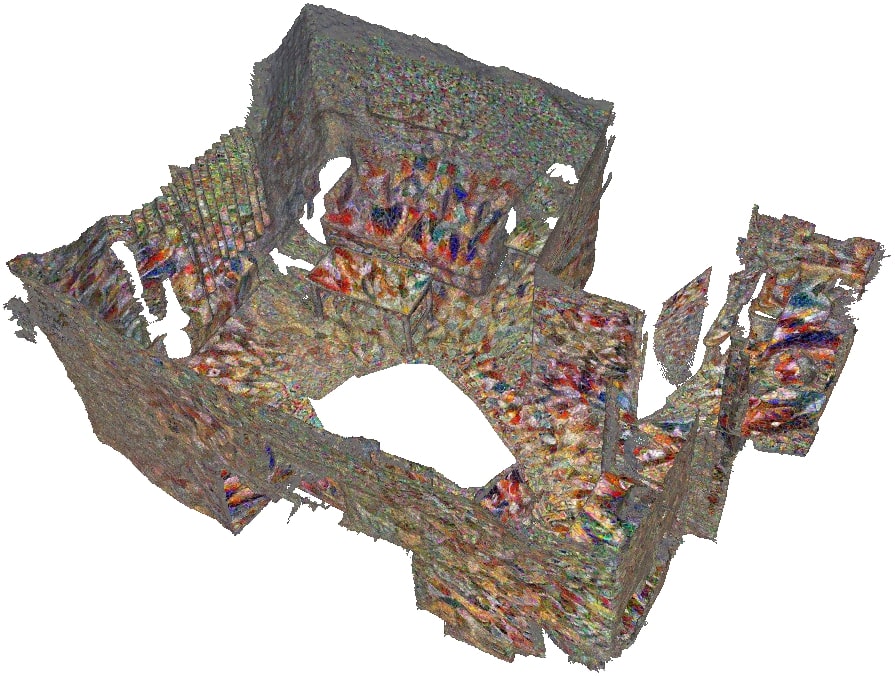} & 
\includegraphics[width=0.24\linewidth]{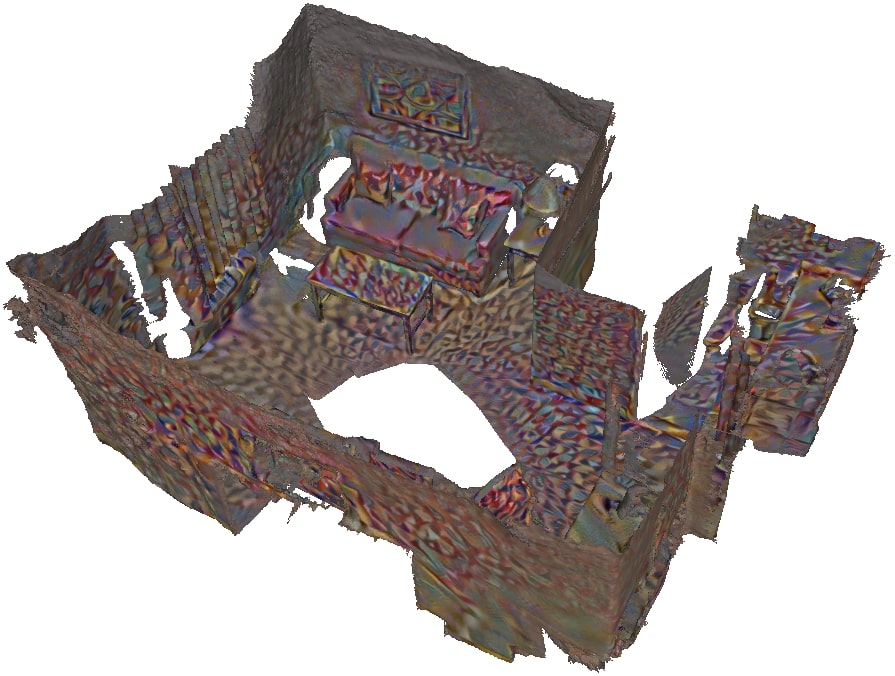} & 
\includegraphics[width=0.24\linewidth]{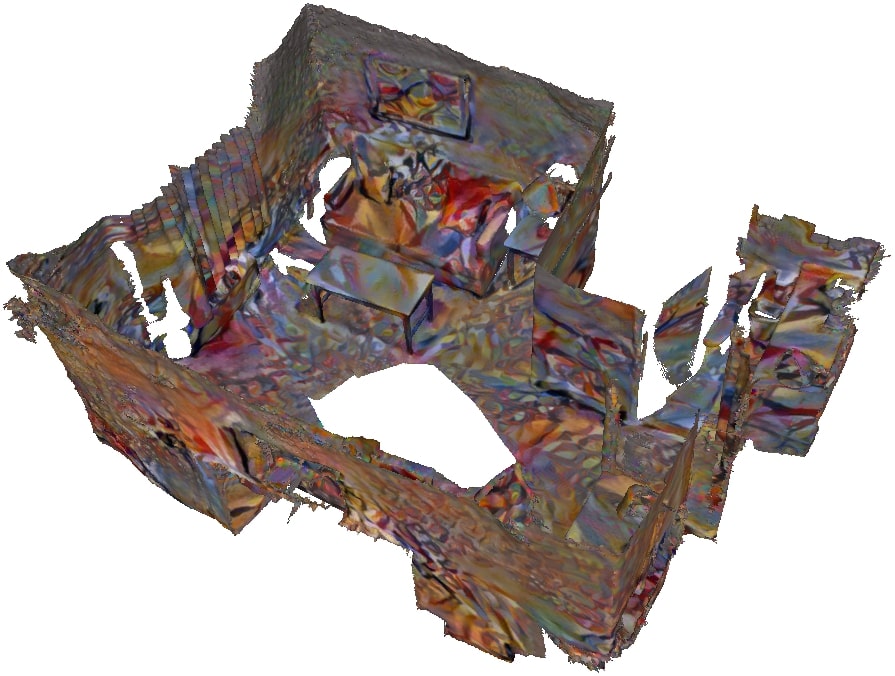} \\
\includegraphics[width=0.24\linewidth]{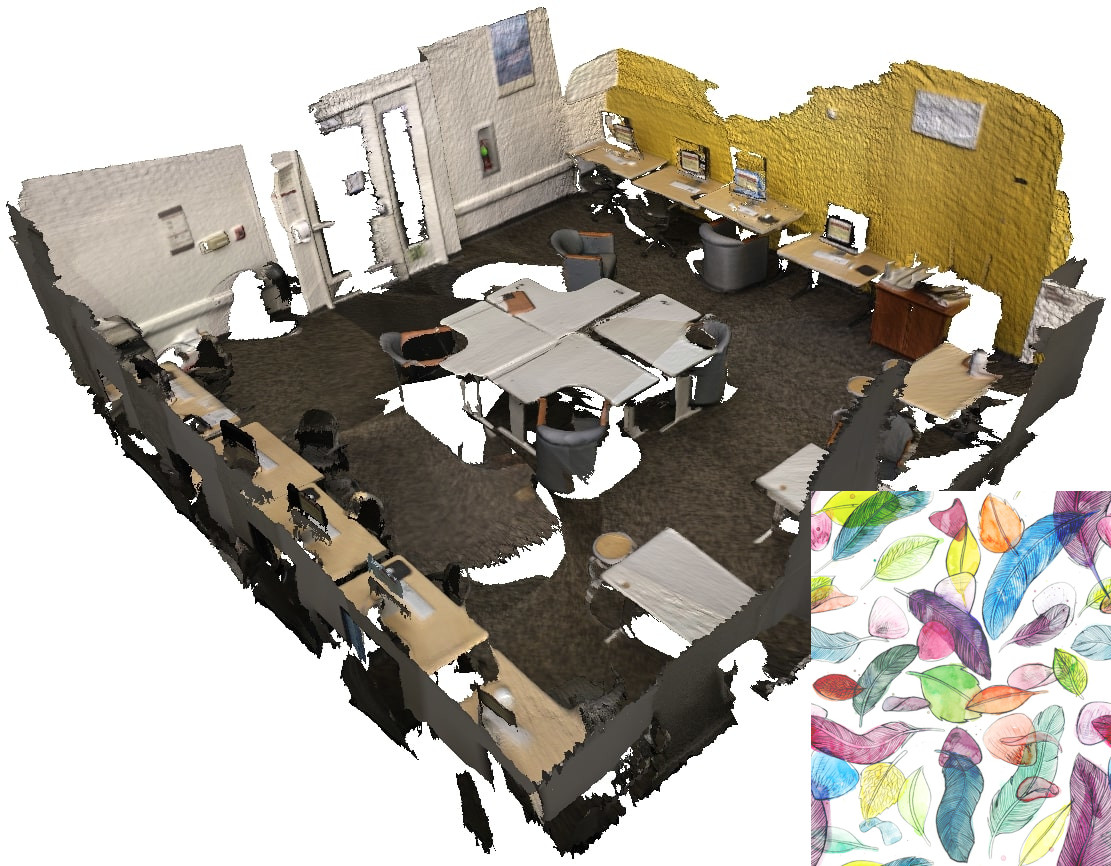} & 
\includegraphics[width=0.24\linewidth]{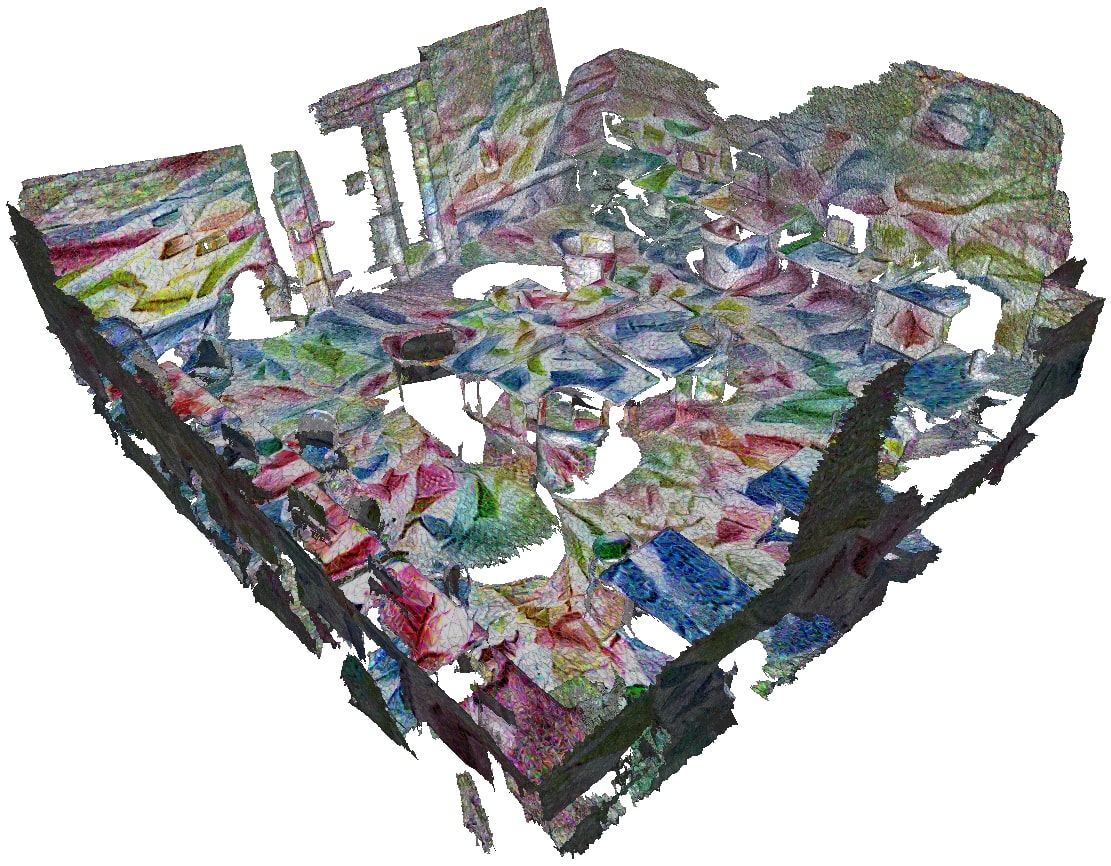} & 
\includegraphics[width=0.24\linewidth]{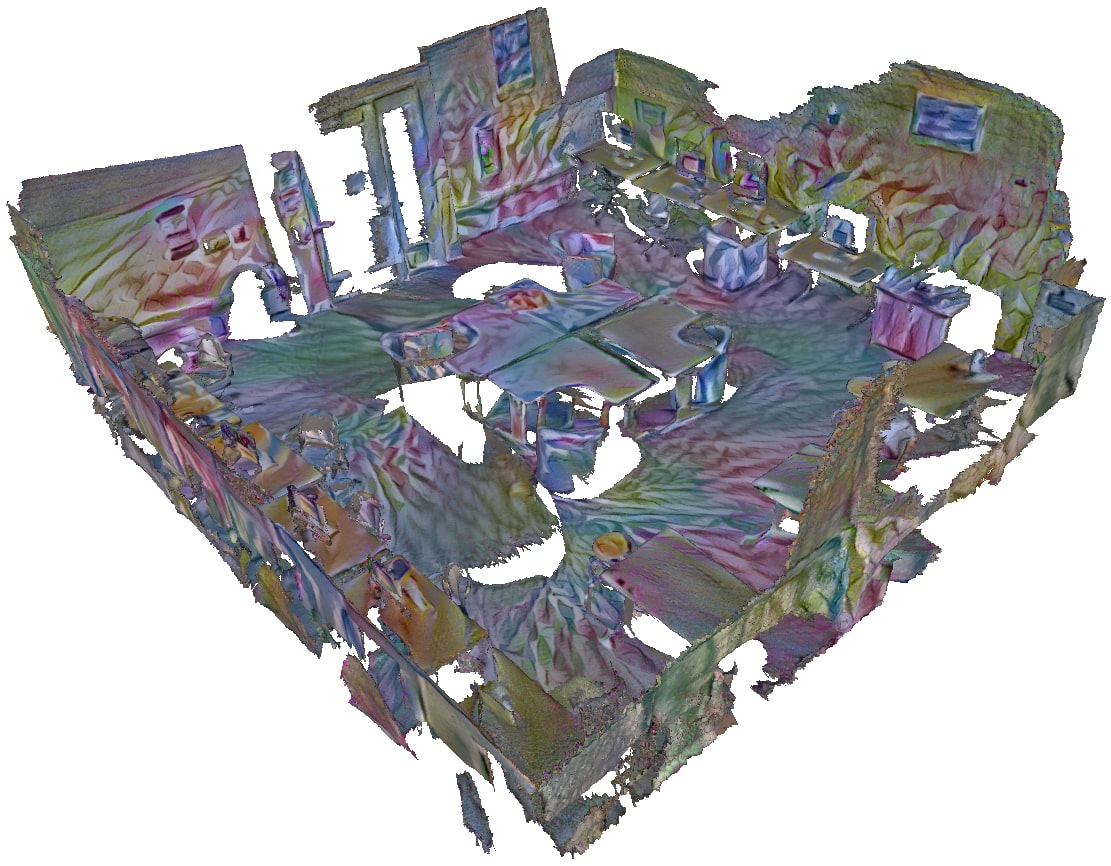} & 
\includegraphics[width=0.24\linewidth]{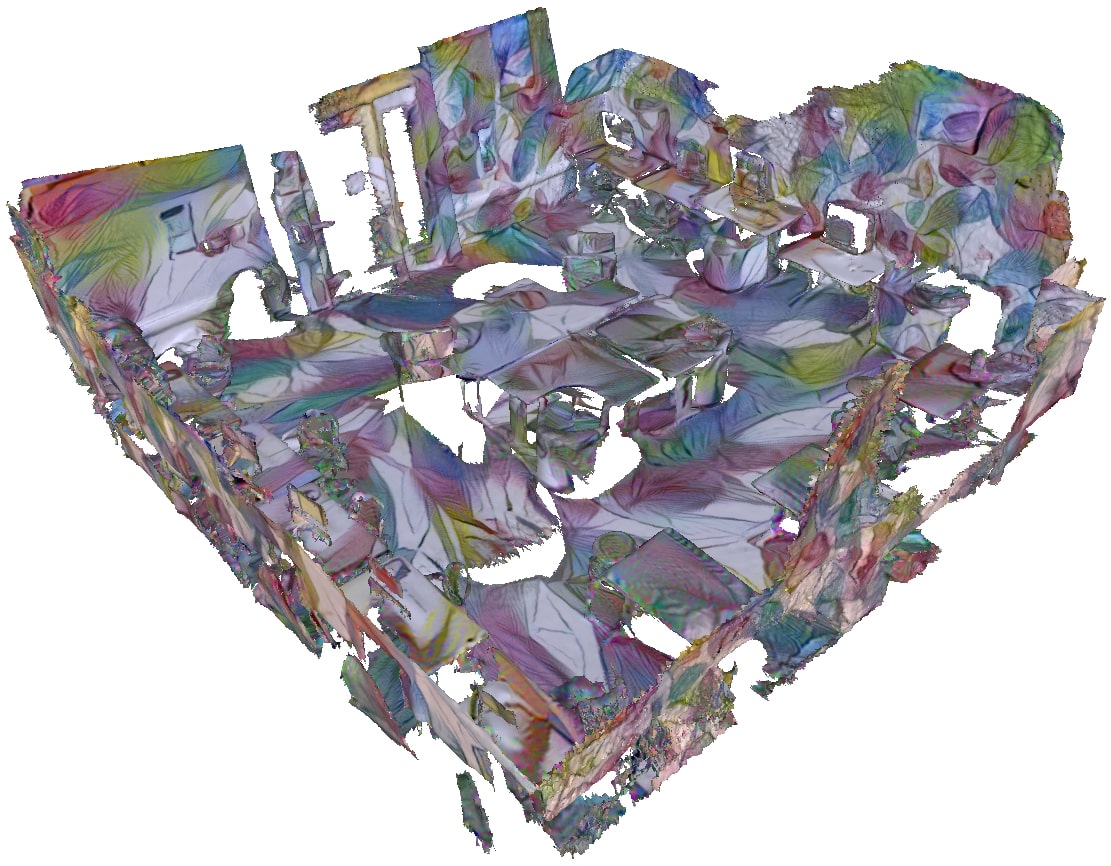} \\
RGB Mesh & NMR~\cite{kato2018neural} & DIP~\cite{mordvintsev2018differentiable} & \textbf{Ours}
\end{tabular}
\caption{Top-down view on stylized meshes in comparison to previous work.}
\label{fig:comp-baseline-3D-add}
\end{figure*}

\begin{figure*}
\centering
\setlength\tabcolsep{1pt}
\begin{tabular}{ccccc}
\includegraphics[height=24mm, width=32mm]{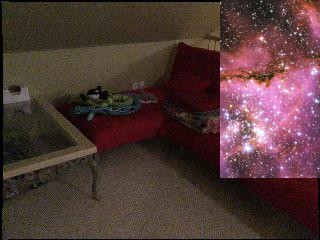} & 
\includegraphics[height=24mm, width=32mm]{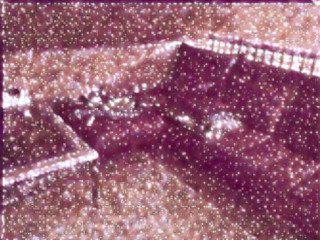} & 
\includegraphics[height=24mm, width=32mm]{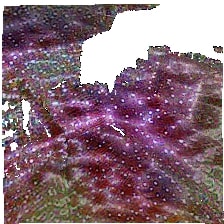} & 
\includegraphics[height=24mm, width=32mm]{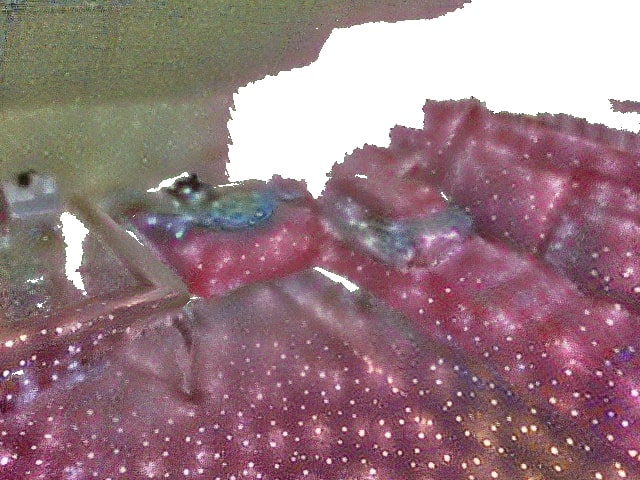} & 
\includegraphics[height=24mm, width=32mm]{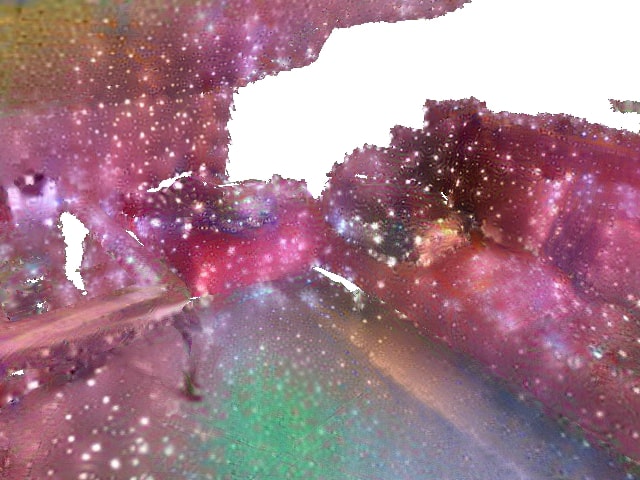} \\
\includegraphics[height=24mm, width=32mm]{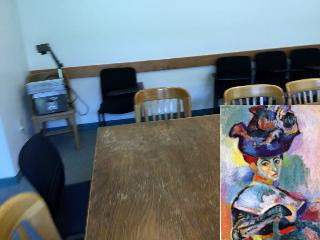} & 
\includegraphics[height=24mm, width=32mm]{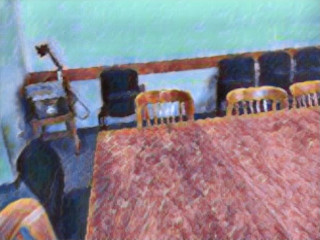} & 
\includegraphics[height=24mm, width=32mm]{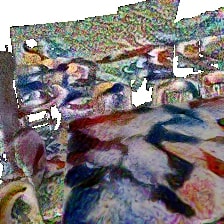} & 
\includegraphics[height=24mm, width=32mm]{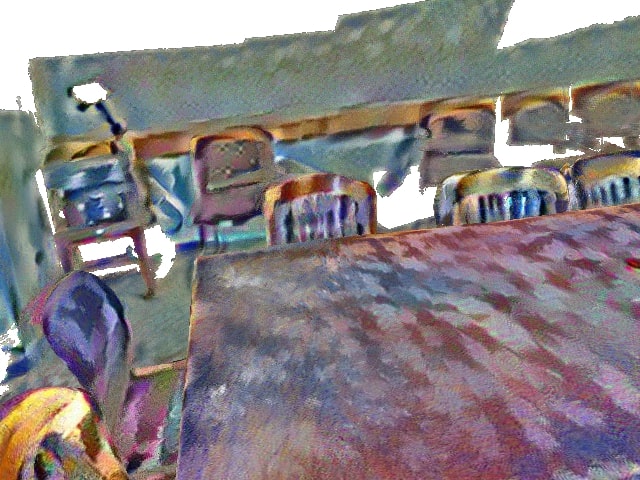} & 
\includegraphics[height=24mm, width=32mm]{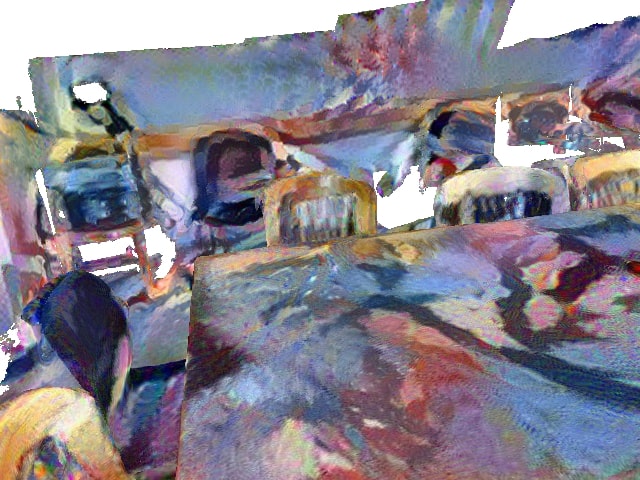} \\
\includegraphics[height=24mm, width=32mm]{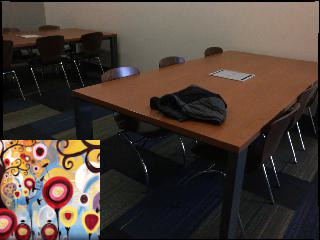} & 
\includegraphics[height=24mm, width=32mm]{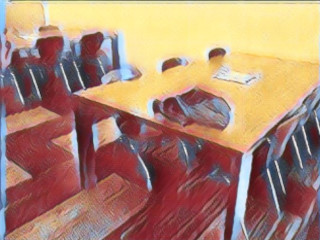} & 
\includegraphics[height=24mm, width=32mm]{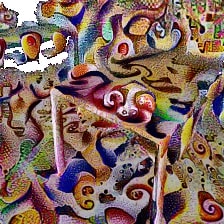} & 
\includegraphics[height=24mm, width=32mm]{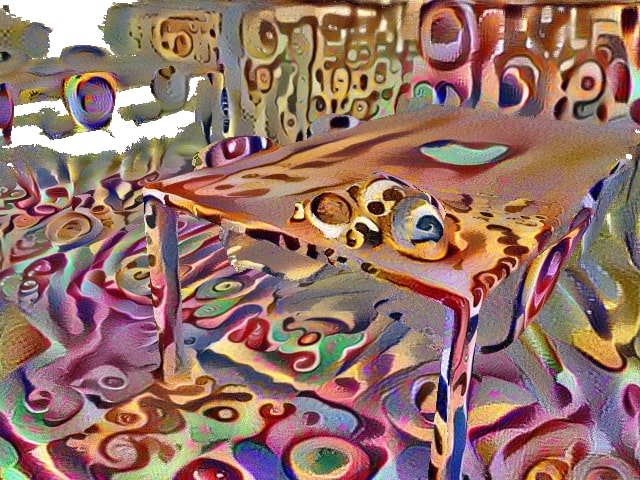} & 
\includegraphics[height=24mm, width=32mm]{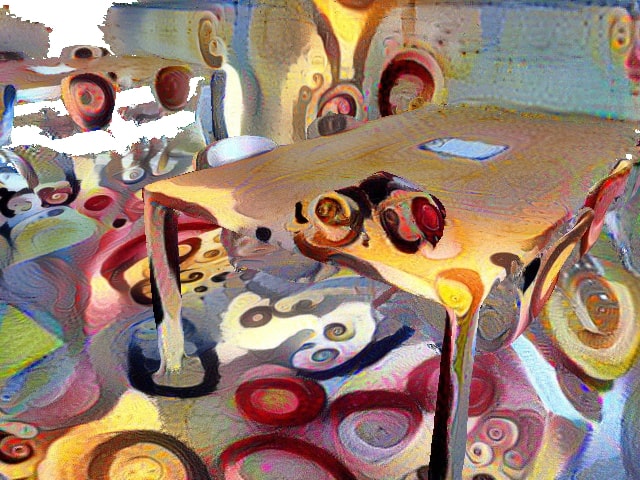} \\
\includegraphics[height=24mm, width=32mm]{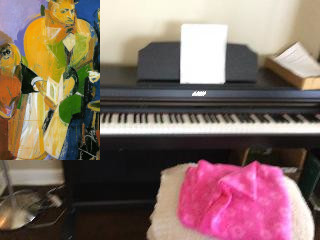} & 
\includegraphics[height=24mm, width=32mm]{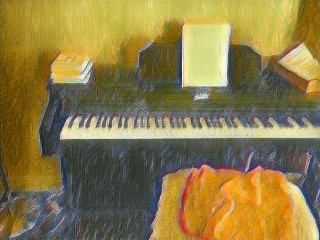} & 
\includegraphics[height=24mm, width=32mm]{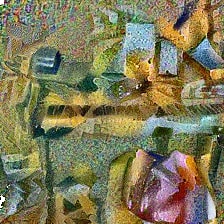} & 
\includegraphics[height=24mm, width=32mm]{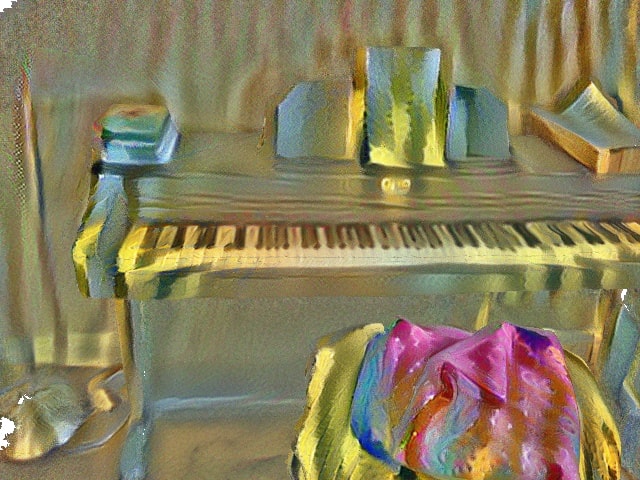} & 
\includegraphics[height=24mm, width=32mm]{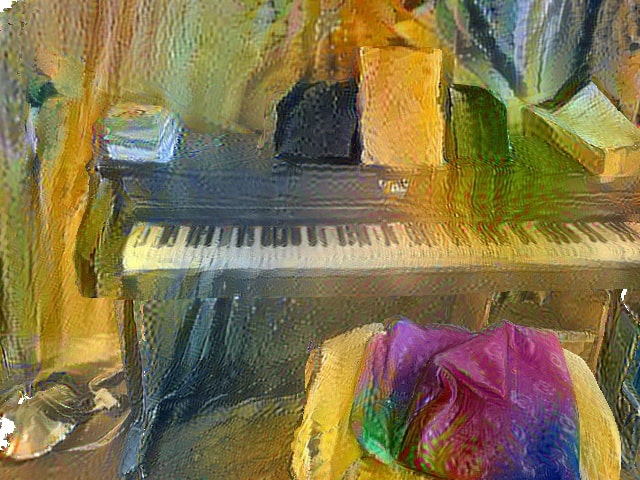} \\
\includegraphics[height=24mm, width=32mm]{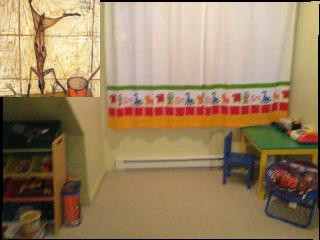} & 
\includegraphics[height=24mm, width=32mm]{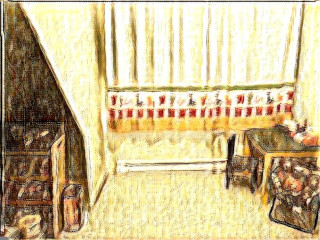} & 
\includegraphics[height=24mm, width=32mm]{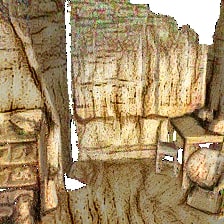} & 
\includegraphics[height=24mm, width=32mm]{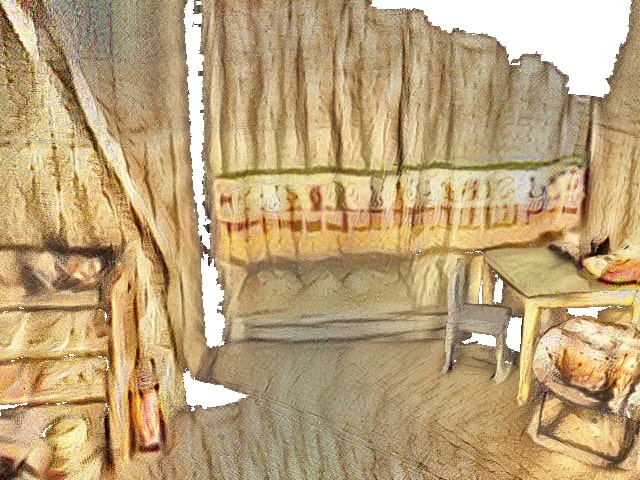} & 
\includegraphics[height=24mm, width=32mm]{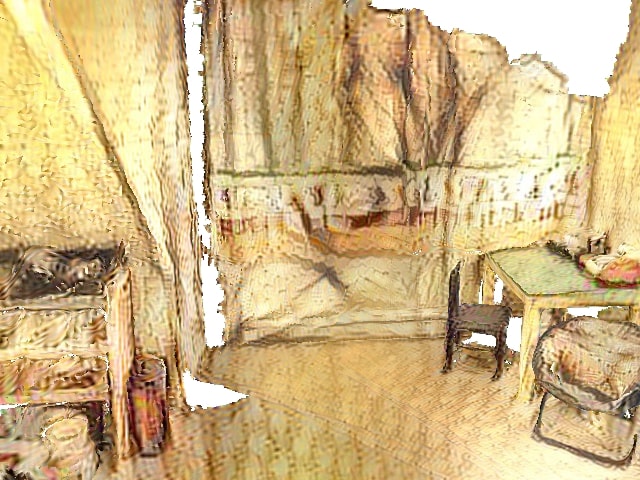} \\
RGB and Style & LSNV~\cite{huang_2021_3d_scene_stylization} & NMR~\cite{kato2018neural} & DIP~\cite{mordvintsev2018differentiable} & \textbf{Ours}
\end{tabular}
\caption{Comparison of stylization results for our method and related work on the ScanNet~\cite{dai2017scannet} dataset. We texture the mesh with each method (point cloud for Huang \etal~\cite{huang_2021_3d_scene_stylization} respectively) and render a single pose that is also captured in the RGB images.}
\label{fig:comp-baseline-scannet-add}
\end{figure*}

Additional comparisons on the ScanNet~\cite{dai2017scannet} dataset can be found in~\cref{fig:comp-baseline-3D-add} and~\cref{fig:comp-baseline-scannet-add}.

\begin{figure*}
\centering
\setlength\tabcolsep{1pt}
\begin{tabular}{cccc}
\includegraphics[width=0.24\linewidth]{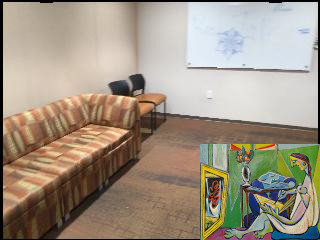} & 
\includegraphics[width=0.24\linewidth]{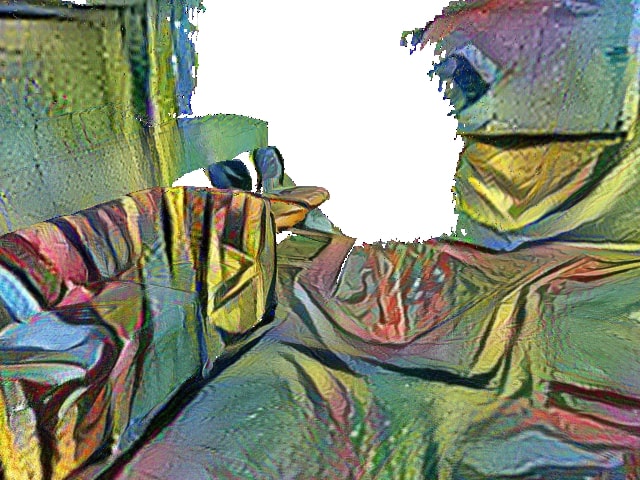} & 
\includegraphics[width=0.24\linewidth]{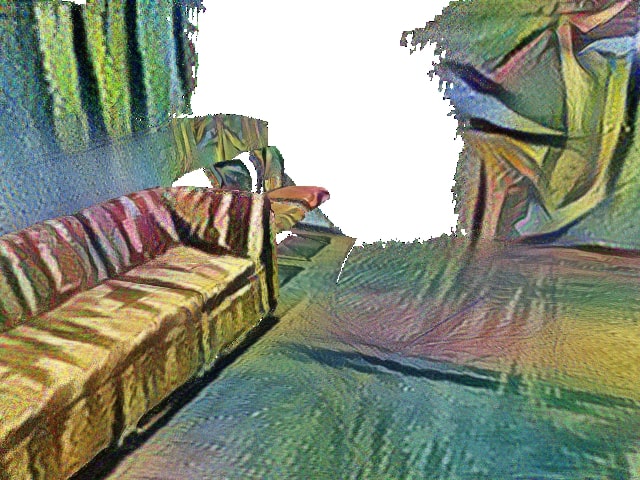} & 
\includegraphics[width=0.24\linewidth]{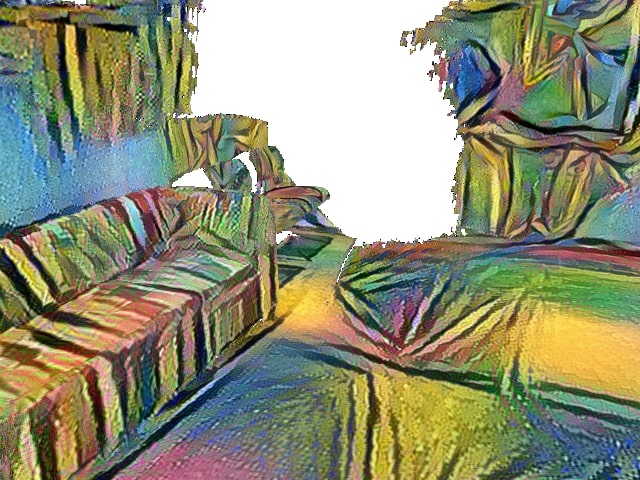} \\
\includegraphics[width=0.24\linewidth]{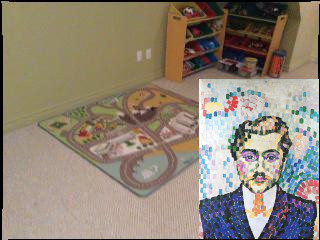} & 
\includegraphics[width=0.24\linewidth]{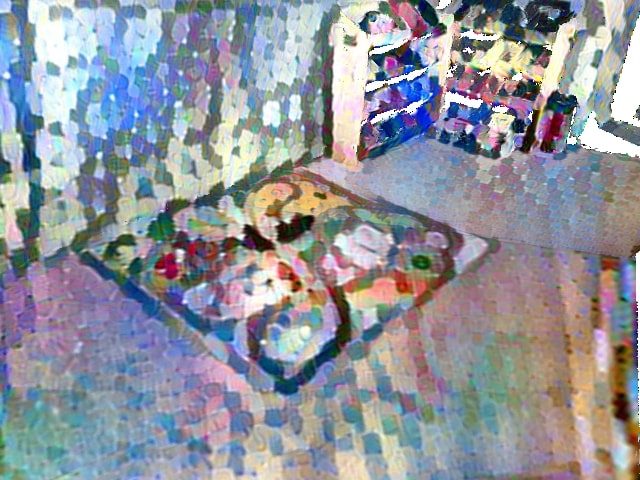} & 
\includegraphics[width=0.24\linewidth]{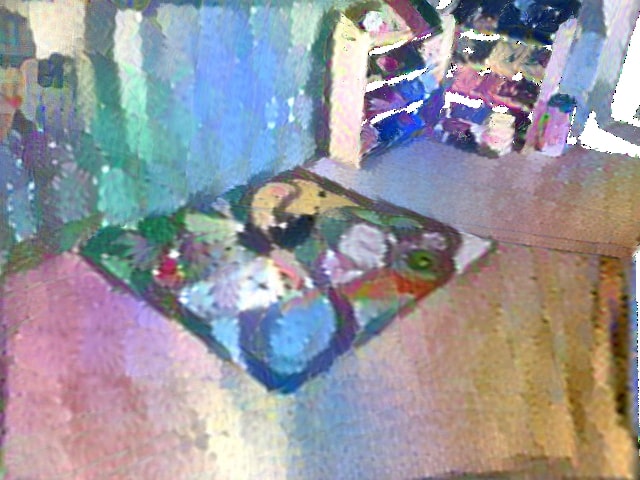} & 
\includegraphics[width=0.24\linewidth]{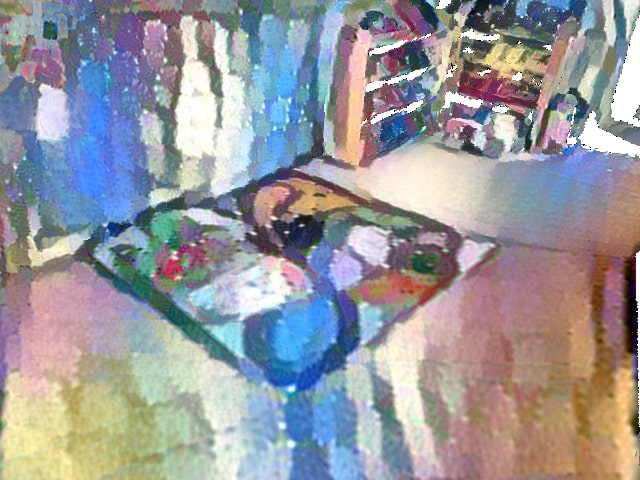} \\
\includegraphics[width=0.24\linewidth]{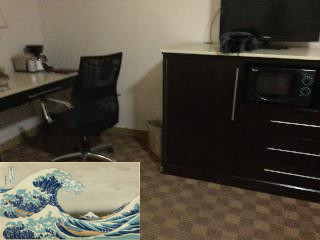} & 
\includegraphics[width=0.24\linewidth]{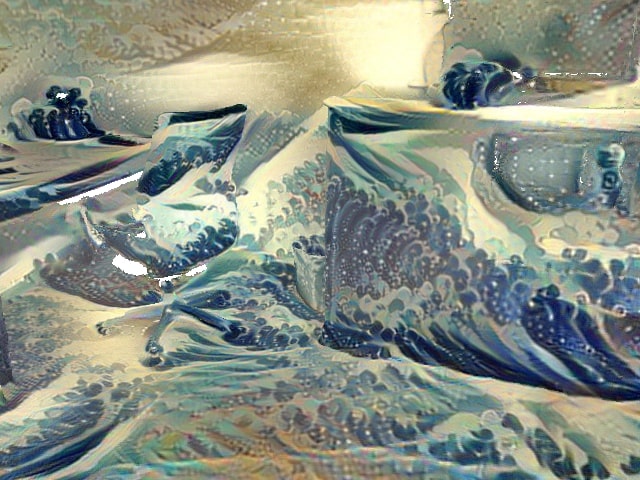} & 
\includegraphics[width=0.24\linewidth]{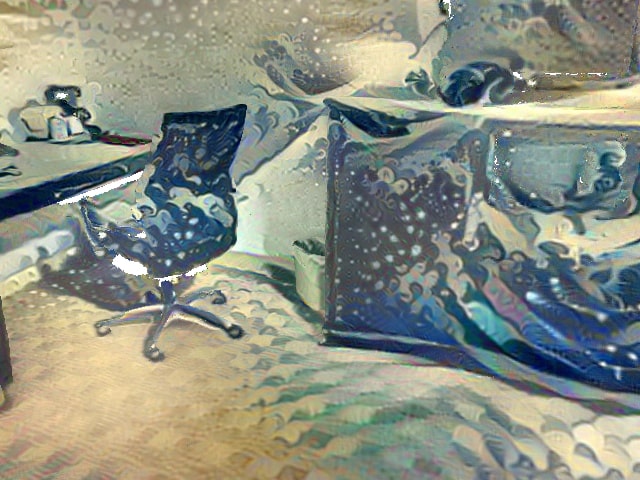} & 
\includegraphics[width=0.24\linewidth]{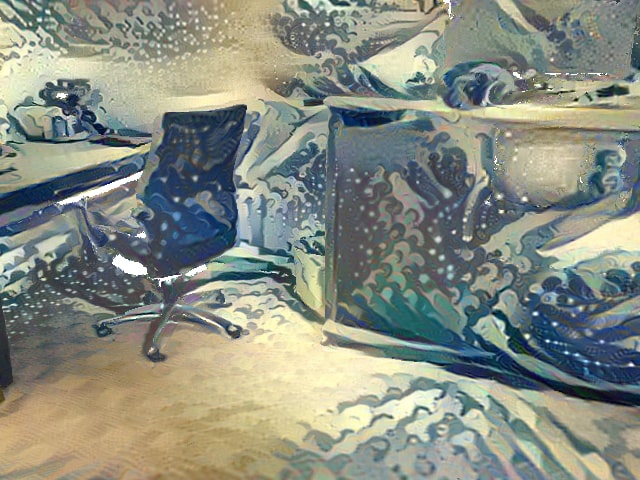} \\
\includegraphics[width=0.24\linewidth]{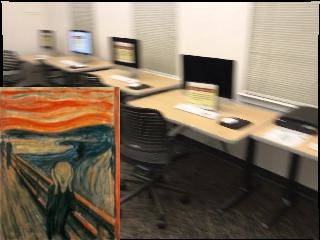} & 
\includegraphics[width=0.24\linewidth]{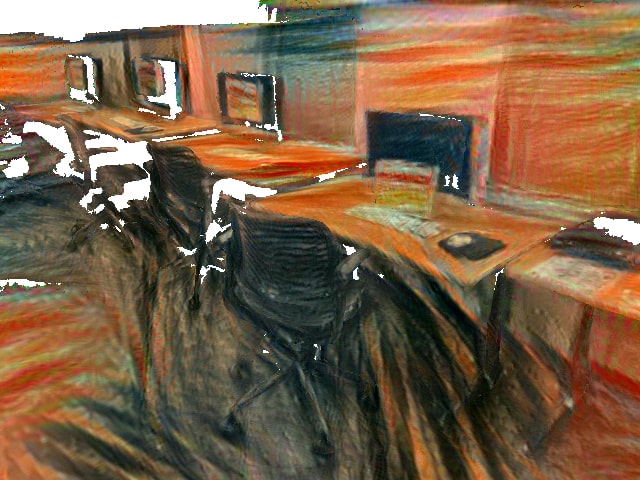} & 
\includegraphics[width=0.24\linewidth]{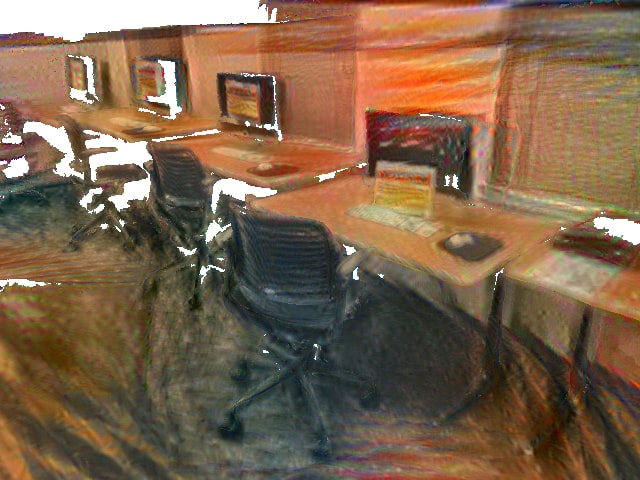} & 
\includegraphics[width=0.24\linewidth]{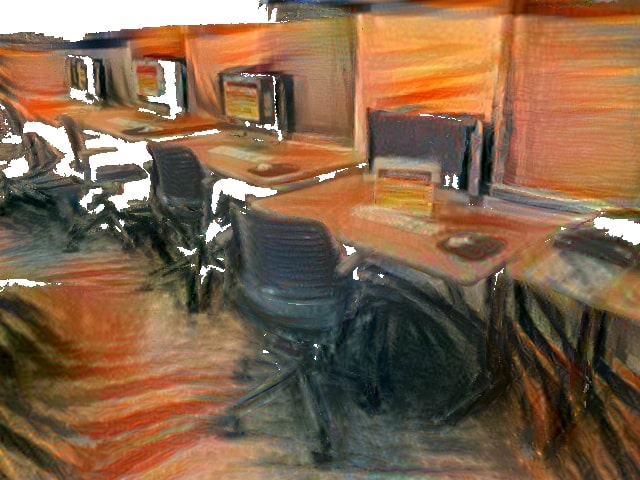} \\
(a) RGB and Style & (b) Only 2D & (c) With Angle & (d) With Angle and Depth
\end{tabular}
\caption{Qualitative ablation study of our method. We compare ours (d) against only using angle (c) and not using angle and depth (b). Using angle better distinguishes surfaces and using depth creates smaller/detailed stylization in the background.}
\label{fig:ablation-add}
\end{figure*}

Additional comparisons for our ablation study can be found in~\cref{fig:ablation-add}.

\subsection{Style Image Assets}\label{subsec:assets}
\begin{figure*}[ht]
\centering
\setlength\tabcolsep{1pt}
\begin{tabular}{p{50mm}p{50mm}p{50mm}}
\includegraphics[height=15mm]{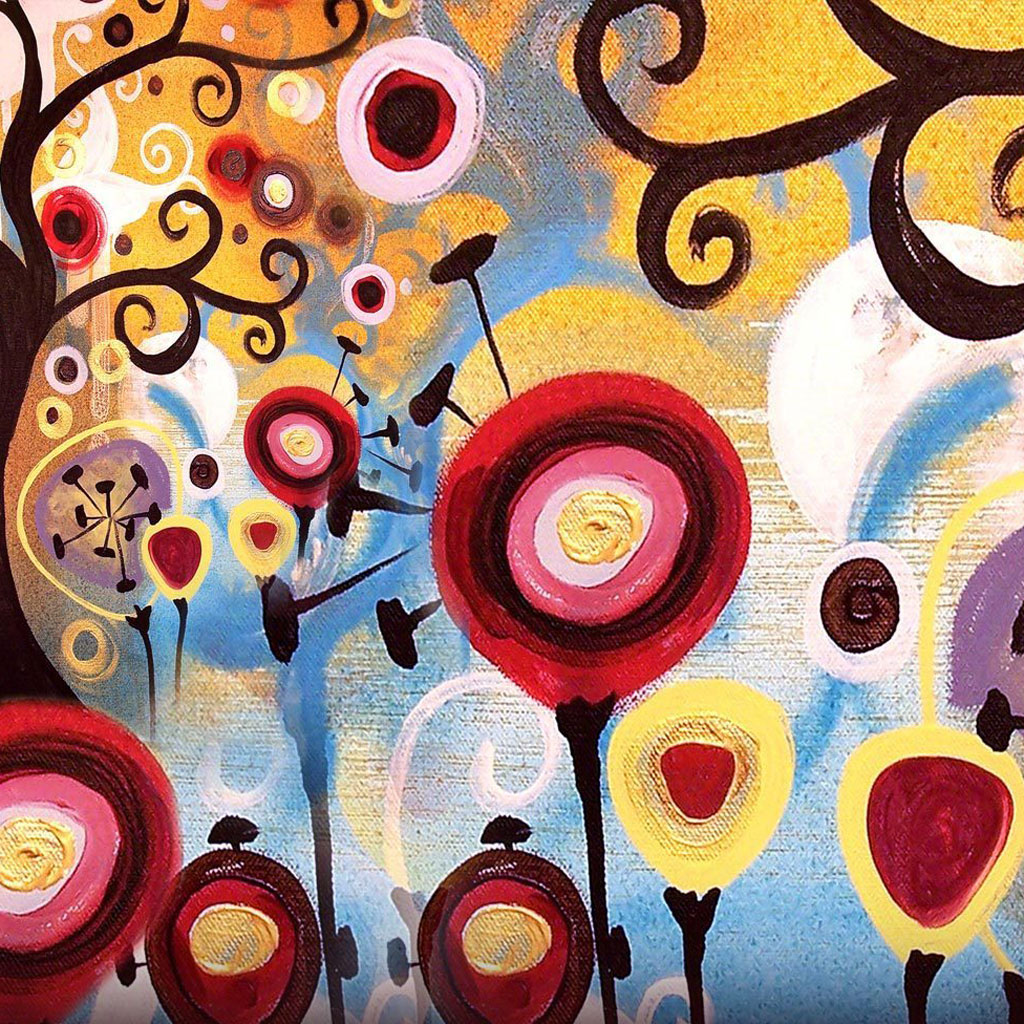} & 
\includegraphics[height=15mm]{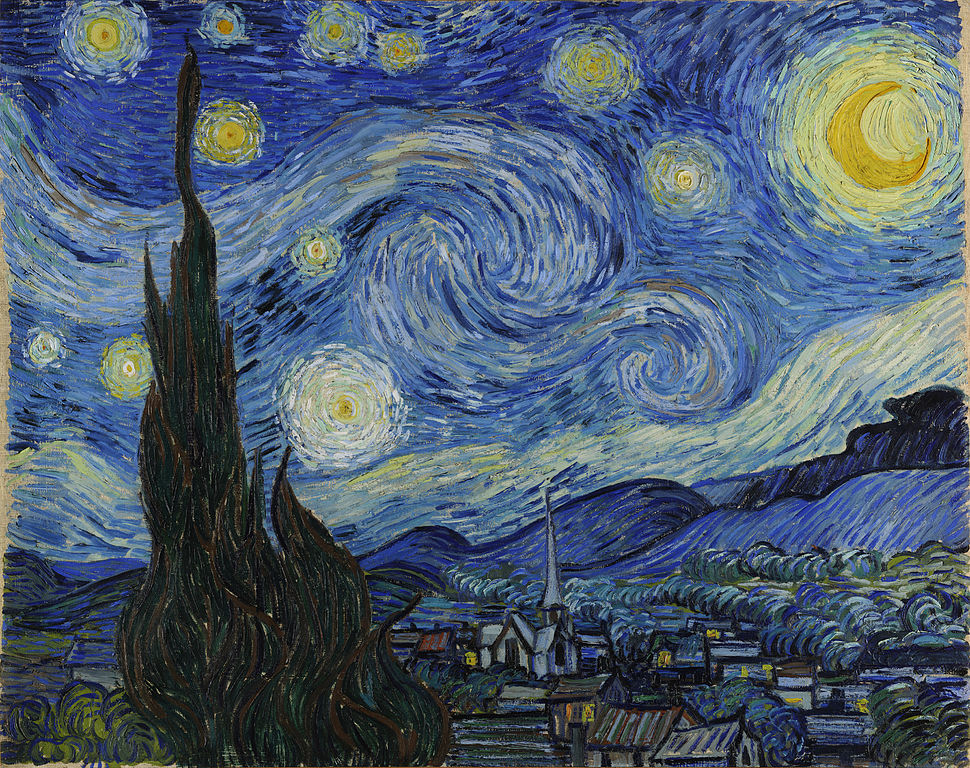} &
\includegraphics[height=15mm]{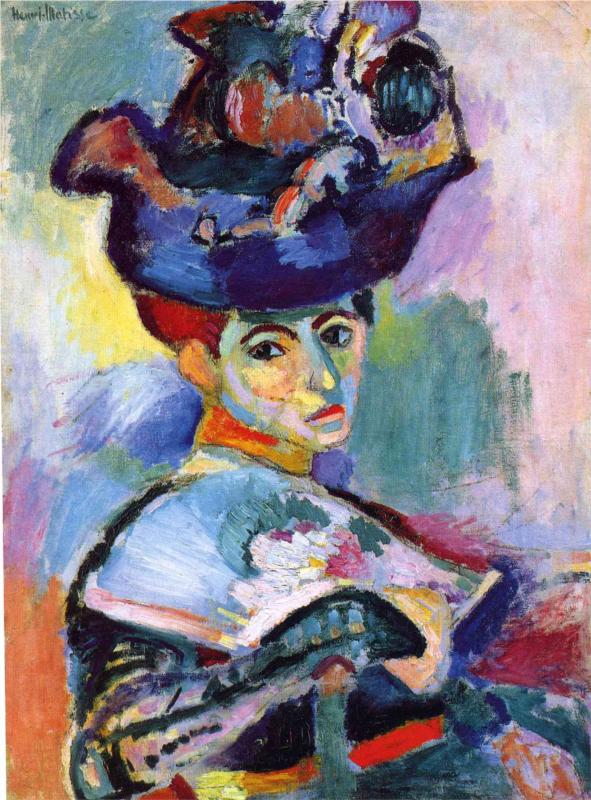} \\
\textit{June Tree},\newline Natasha Wescoat & \textit{The Starry Night}, \newline Vincent van Gogh, 1889 & \textit{Femme au chapeau}, \newline Henri Matisse, 1905 \\

\includegraphics[height=15mm]{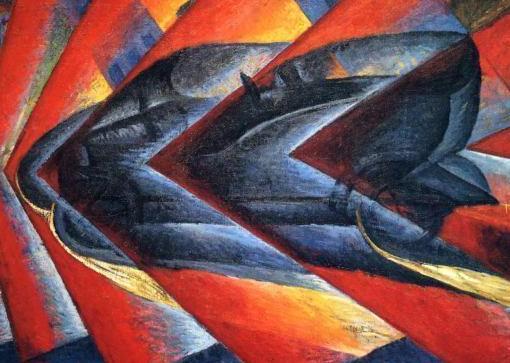} &
\includegraphics[height=15mm]{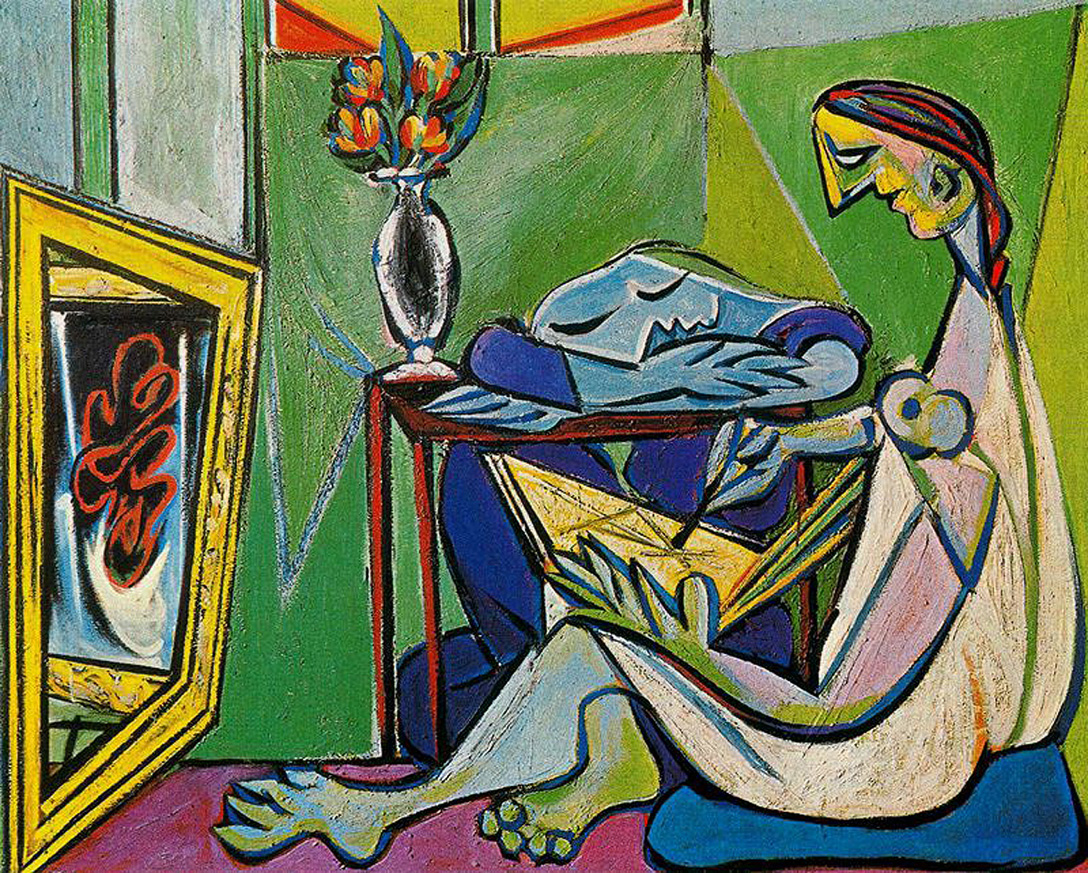} & 
\includegraphics[height=15mm]{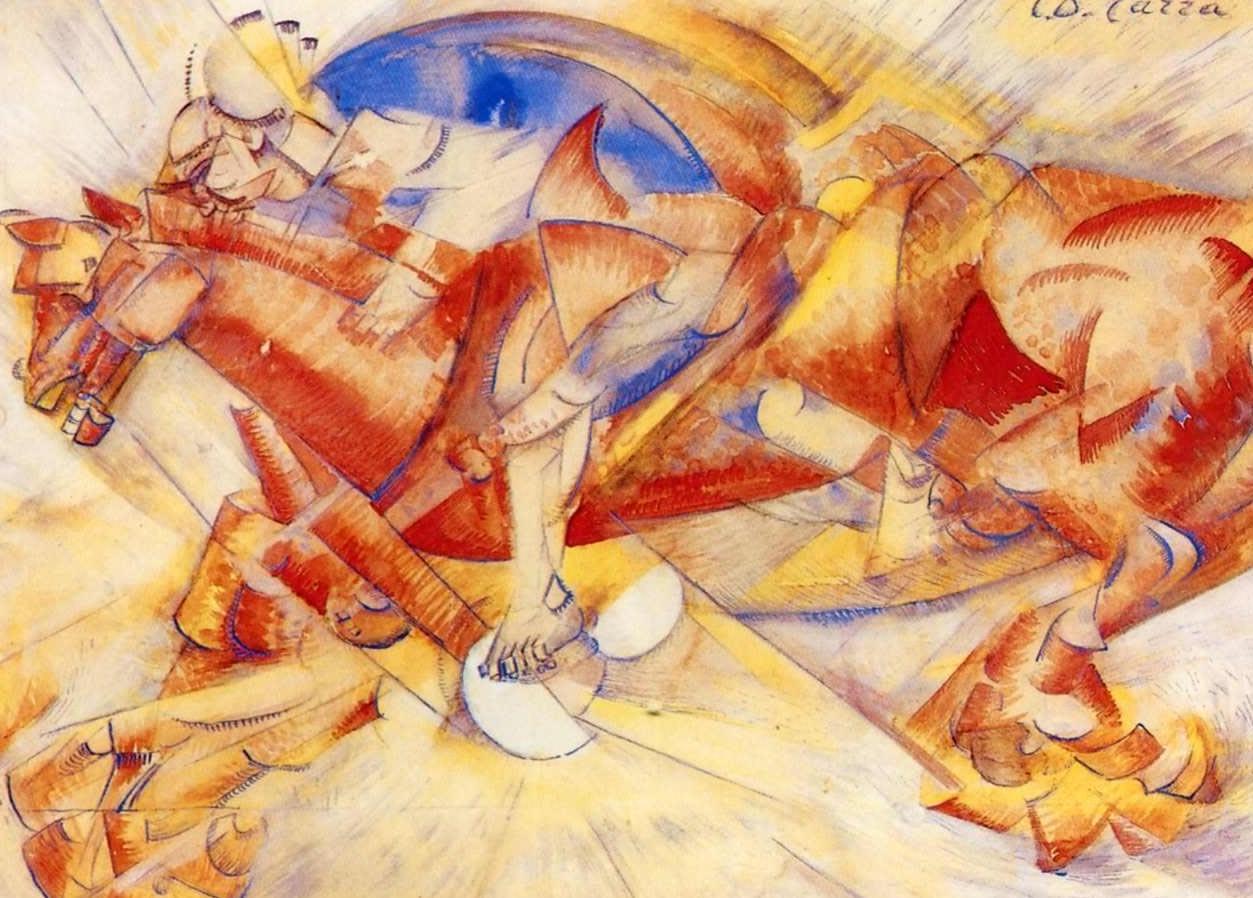} \\
\textit{Dinamismo di un’ automobile}, \newline Luigi Russolo, 1913 & \textit{The Muse}, \newline Pablo Picasso, 1935 & \textit{Il cavaliere rosso}, \newline Carlo Carra, 1913 \\

\includegraphics[height=15mm]{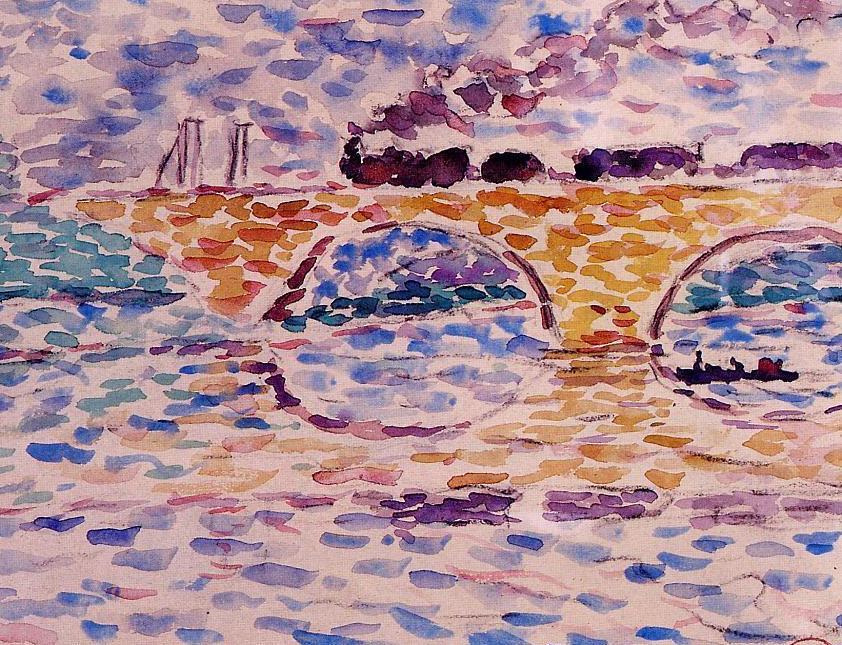} &
\includegraphics[height=15mm]{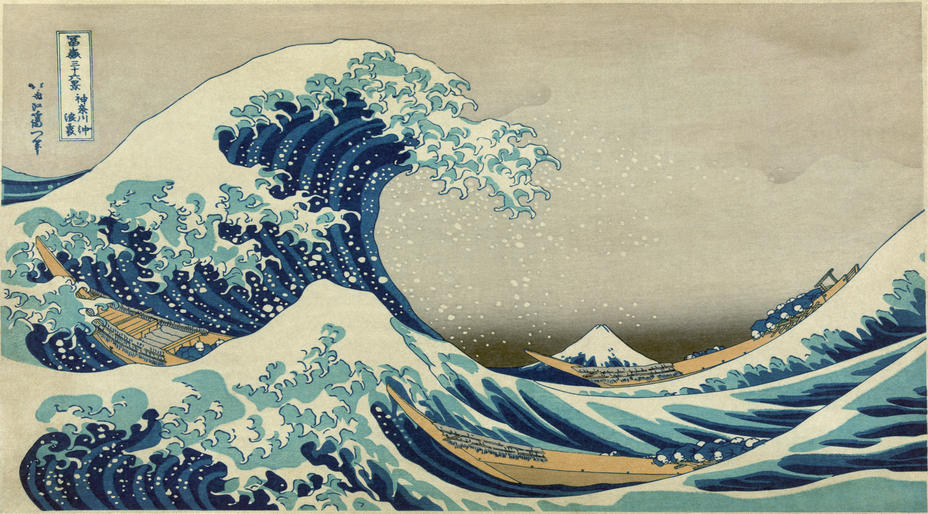} &
\includegraphics[height=15mm]{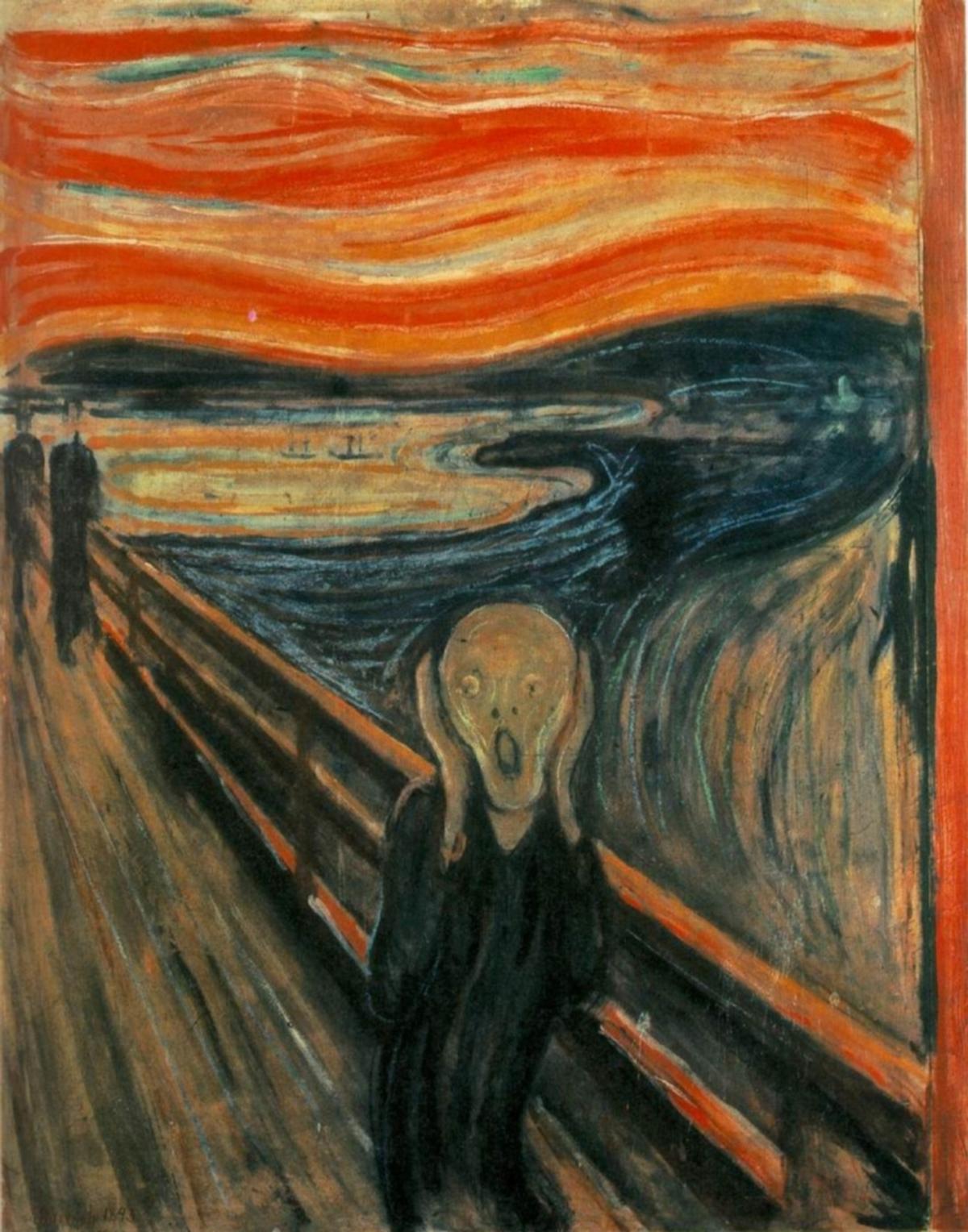} \\
\textit{The Viaduct}, \newline Henri Edmond Cross & \textit{Kanagawa oki nami ura}, \newline Katsushika Hokusai, 1830-1832 & \textit{Skrik}, \newline Edvard Munch, 1893 \\

\includegraphics[height=15mm]{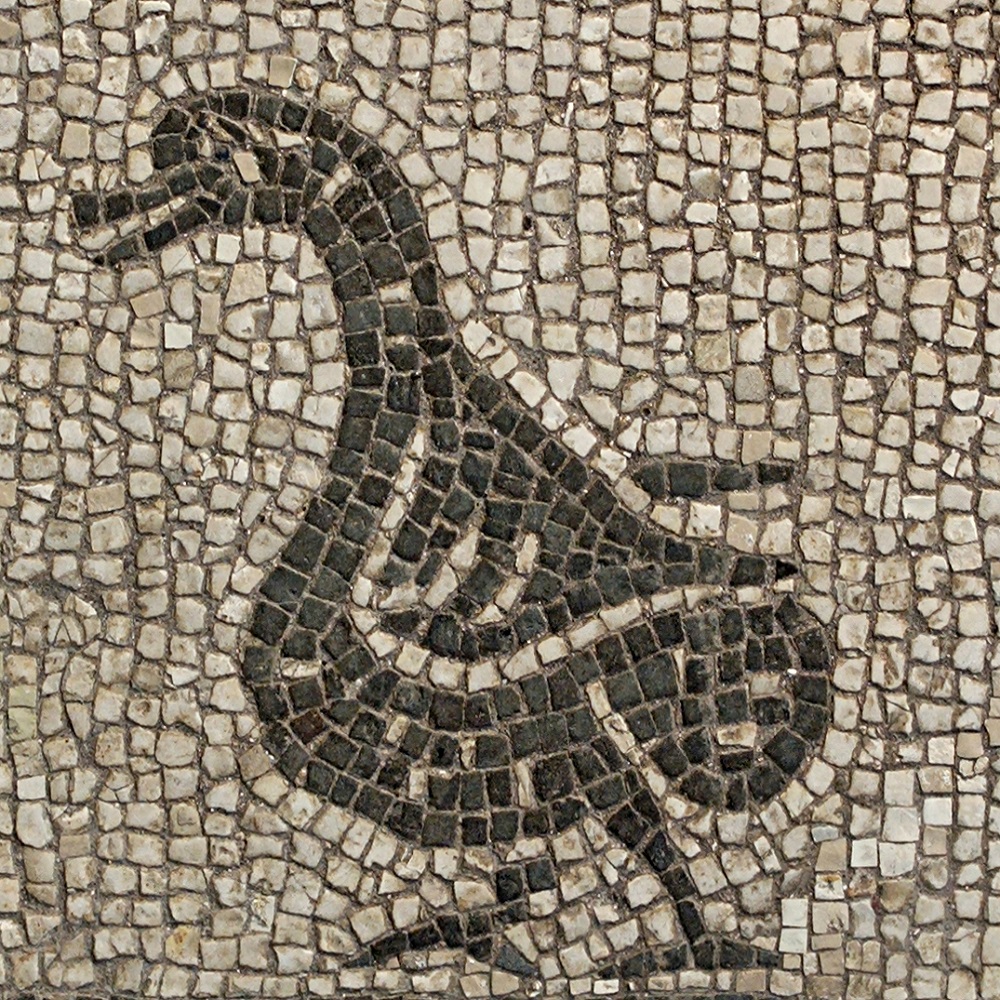} &
\includegraphics[height=15mm]{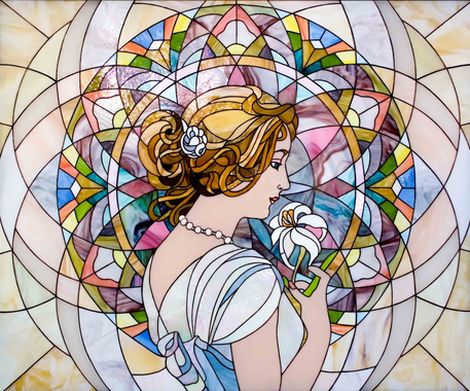} & 
\includegraphics[height=15mm]{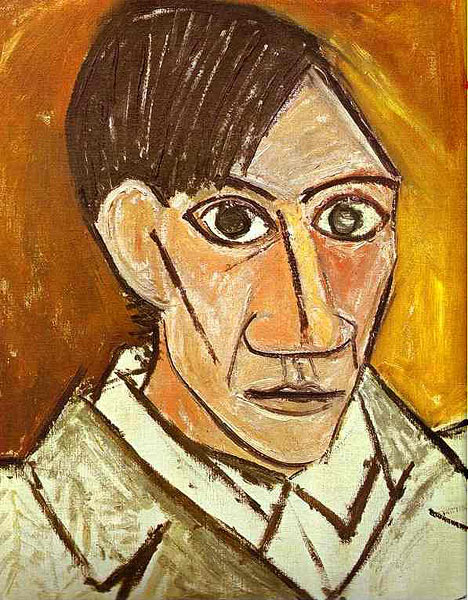} \\
\textit{Mosaic in Opus tessellatum} & \textit{Mosaic (unknown)}, \newline WikiArt.org & \textit{Self-Portrait}, \newline Pablo Picasso, 1907 \\

\includegraphics[height=15mm]{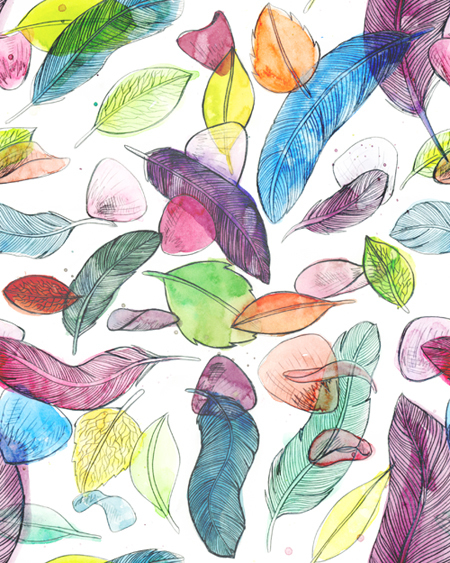} &
\includegraphics[height=15mm]{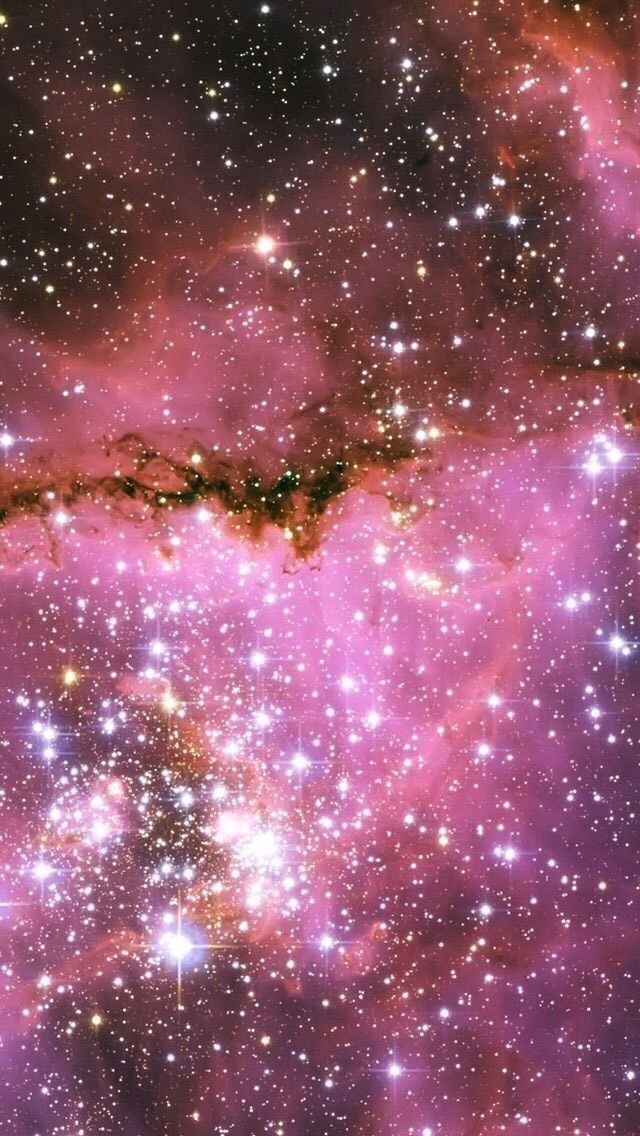} &
\includegraphics[height=15mm]{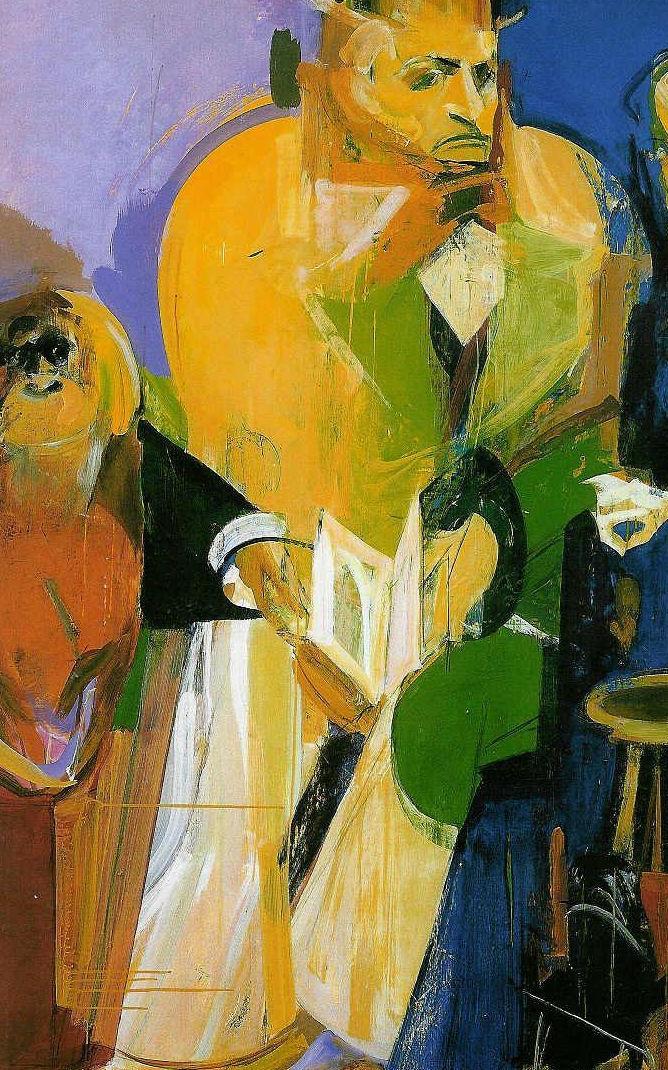} \\
\textit{Feathers Leaves and Petals}, \newline Kathryn Corlett & \textit{Small Magellanic Cloud}, \newline NASA, ESA and A. Nota & \textit{Edgar Poe, Charles Baudelaire, \newline Um Orangotango e o Corvo}, \newline Julio Pomar, 1985 \\

\includegraphics[height=15mm]{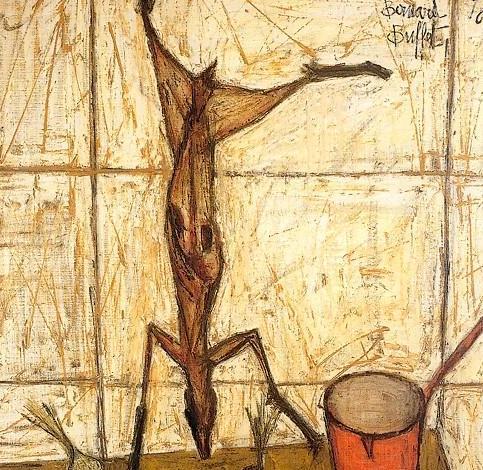} & 
\includegraphics[height=15mm]{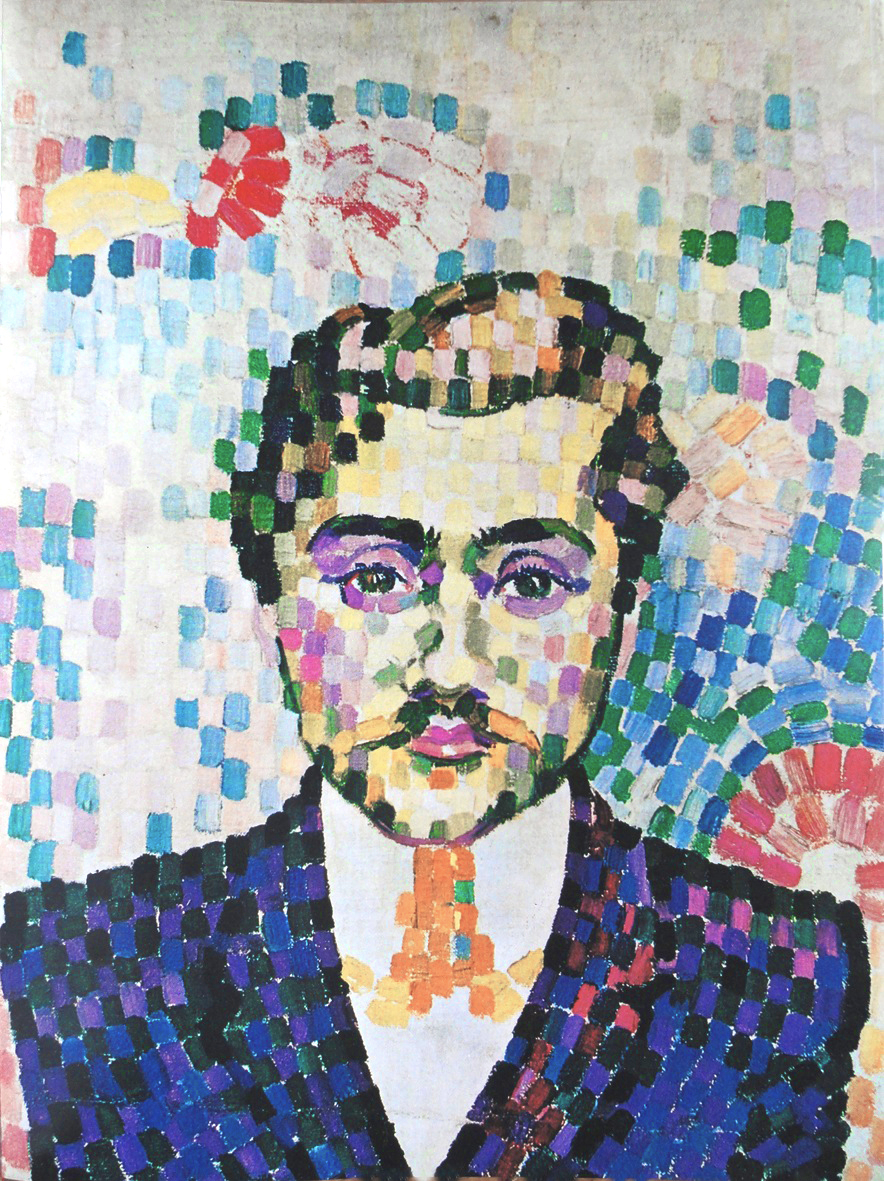} &
\includegraphics[height=15mm]{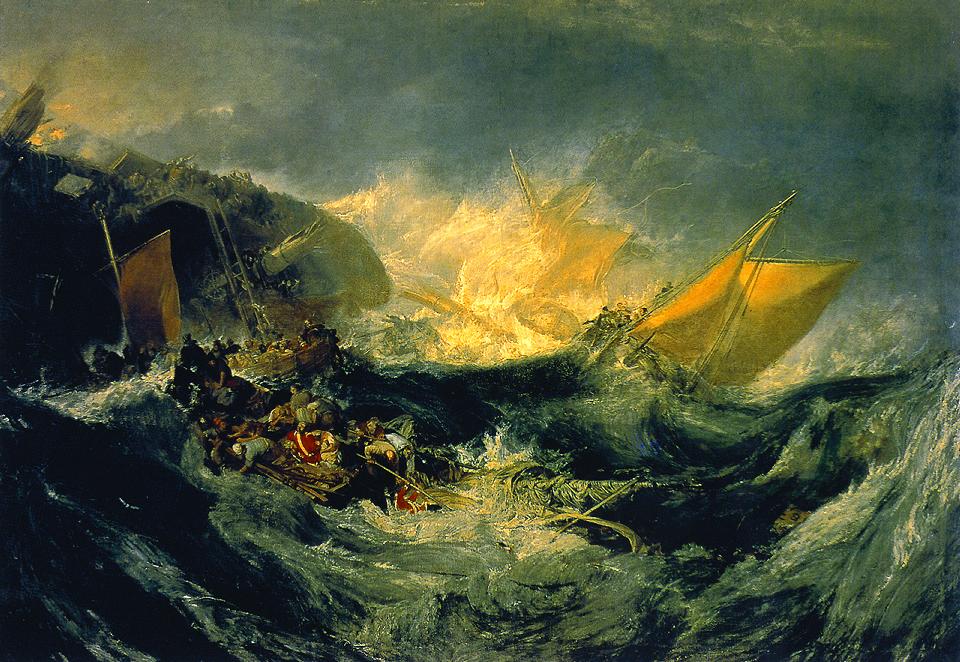} \\
\textit{Lapin et casserole rouge}, \newline Bernard Buffet, 1948 & \textit{L'homme à la tulipe}, \newline Jean Metzinger, 1906 & \textit{The Shipwreck of the Minotaur}, \newline J.M.W. Turner, 1805 \\

\includegraphics[height=15mm]{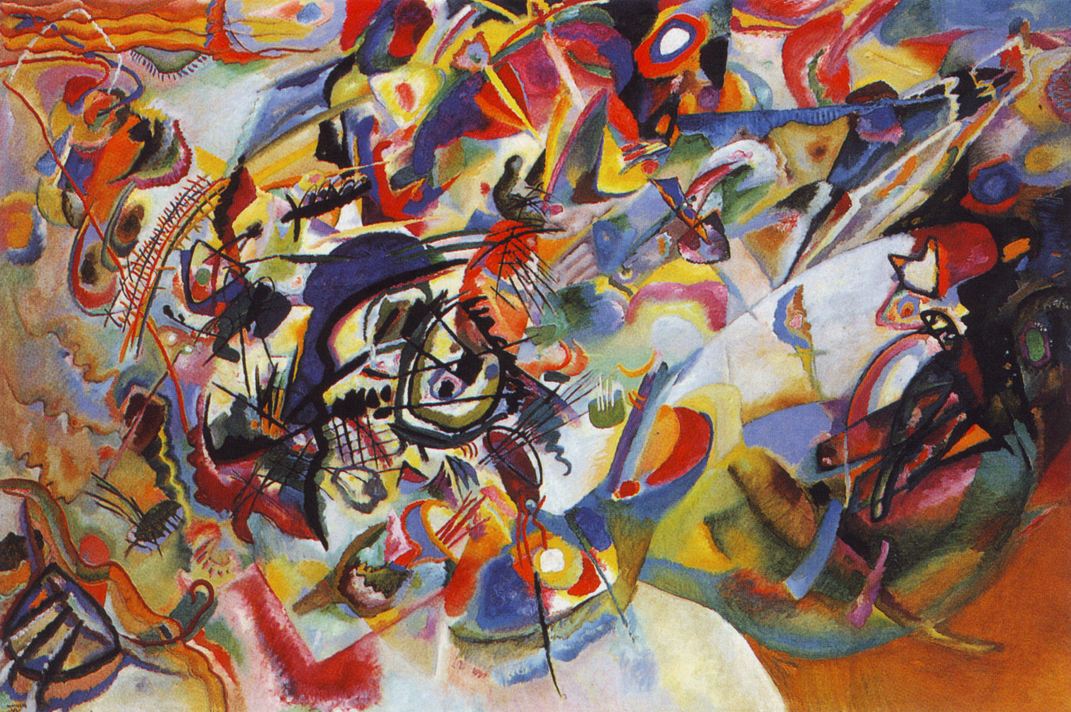} \\
\textit{Sketch 2 for composition VII}, \newline Wassily Kandinsky, 1913 
\end{tabular}
\caption{List of all artistic paintings used throughout the main paper and supplemental material. We list the name of the painting and its author if known.}
\label{fig:image-assets}
\end{figure*}

Throughout the main paper and the supplemental material, we use style images created by artists.
In~\cref{fig:image-assets} we list all images and give credit to their respective creators.

\end{document}